\documentclass{article}

\usepackage[preprint]{neurips_2021}

\usepackage[utf8]{inputenc} 
\usepackage[T1]{fontenc}    
\usepackage[hidelinks]{hyperref}       
\usepackage{url}            
\usepackage{booktabs}       
\usepackage{amsfonts}       
\usepackage{nicefrac}       
\usepackage{microtype}      
\usepackage{xcolor}         
\usepackage[cmex10]{amsmath}

\usepackage{amssymb}
\usepackage[pdftex]{graphicx}
\usepackage{subfigure}
\usepackage{array}

\usepackage{soul}

\DeclareMathOperator{\E}{\mathbb{E}}
\DeclareMathOperator*{\argmax}{argmax}
\newcolumntype{L}{>{\displaystyle}l}
\newcolumntype{M}[1]{>{\centering\arraybackslash$\displaystyle}m{#1}<{$}}
\title{A Unified View of Algorithms for Path Planning Using Probabilistic Inference on Factor Graphs}

\author{%
  Francesco A. N. Palmieri$^1$ \And 
  Krishna R. Pattipati$^2$ \And
  Giovanni Di Gennaro$^1$ \And
  Giovanni Fioretti$^1$ \And
  Francesco Verolla$^1$ \And
  Amedeo Buonanno$^3$ \AND \\[-1em] 
  $^1$Dipartimento di Ingegneria \\ 
   Università degli Studi della Campania ``Luigi Vanvitelli'' \\ 
   Aversa (CE), Italy \\
   \texttt{\{francesco.palmieri, giovanni.digennaro\}@unicampania.it} \\
   \texttt{\{giovanni.fioretti, francesco.verolla\}@studenti.unicampania.it} \\ \AND \\[-1em]
  $^2$Department of Electrical and Computer Engineering \\
   University of Connecticut \\
   Storrs (CT), USA \\
   \texttt{krishna.pattipati@uconn.edu} \\ \AND \\[-1em]
  $^3$ENEA \\
   Department of Energy Technologies and Renewable Energy Sources \\
   Portici (NA), Italy \\ 
   \texttt{amedeo.buonanno@enea.it}
}

\begin{document}

\maketitle

\begin{abstract}
Even if path planning can be solved using standard techniques from dynamic programming and control,  the problem can  also be approached using probabilistic inference. The algorithms that emerge using the latter framework bear some appealing characteristics that qualify the probabilistic approach as a powerful alternative to the more traditional control formulations.  The idea of using estimation on stochastic models to solve control problems is not new and the inference approach considered here falls under the rubric of Active Inference (AI) and Control as Inference (CAI).  In this work, we look at the specific  recursions that arise from various cost functions that, although they may appear similar in scope, bear noticeable differences, at least when applied to typical path planning problems. We start by posing the path planning problem on a probabilistic factor graph, and show how the various algorithms translate into specific message composition rules. We then show how this unified approach, presented both in probability space and in log space, provides a very general framework that includes the Sum-product, the Max-product, Dynamic programming and mixed Reward/Entropy criteria-based algorithms. The framework also expands algorithmic design options for smoother or sharper policy distributions, including generalized Sum/Max-product algorithm, a Smooth Dynamic programming algorithm and modified versions of the Reward/Entropy recursions. We provide
a comprehensive table of recursions and a comparison through simulations, first on a synthetic small grid with a single goal with obstacles, and then on a grid extrapolated from a real-world scene with multiple goals and a semantic map.  
\end{abstract}

\section{Introduction}
Probabilistic methods for modeling  the dynamic behavior of agents navigating in complex environments are becoming increasingly popular in the recent literature. In many applications, the recognition that a stochastic control problem can be solved through probabilistic inference on a generative model,  has sparked new growing interest.
The idea that an agent can use its best inference about the future based on its approximate knowledge of the environment, to condition its present actions, can be a quite powerful paradigm. Dynamic Programming (DP) algorithms for  Markov Decision Processes (MDP) and Partially-Observable Markov Decision Processes (POMDP) \citep{Thrun2006, Bertsekas2019} are based on this concept, suggesting that an agent can act optimally following a best behaviors derived from a value  function back-propagated from the hypothetical future. Analogies between control and estimations can be traced back to the work of Kalman \citep{Todorov2008} and to more recent attempts to see probabilities  and rewards under the same framework \citep{Kappen2012, Levine2018}.  

In path planning, since the original paper by \cite{Attias2003}, there has been a growing body of literature in trying to clarify the connections between the probabilistic methods and more traditional stochastic control strategies  
\citep{Touissaint2006, Toussaint2009, Levine2018}.
Also the terms  Active Inference (AI),   
Control as Inference  (CAI),  have been recently coined \citep{AI2020} with some of these models  based on the so-called free-energy principle \citep{Buckley2017, Parr2018}, on KL-learning \citep{Zenon2018, ParrFrontiers2018, Kaplan2018}, and on Max entropy  \citep{Ziebart2010}. Furthermore, intriguing connections have been drawn, for some of these methods,  to neuroscience and brain theory \citep{Baltieri2017}. Causal reasoning \citep{NairSavareseetal2019} also seems to share some  elements of this goal-directed behavior.

In our present work, we try to go beyond the now well-recognized analogies between probability and control,  because even if some of the resulting algorithms look similar, they may bear marked differences when applied to real world scenarios. 
In Reinforcement Learning (RL) \citep{Sutton2018}, for example,  often we have only partial knowledge of the environment via an approximate  stochastic model and we need to perform  further exploration. But how do we choose  among the various algorithms,  which are typically introduced for known system dynamics, when they are used on temporary knowledge? 
It is often argued that stochastic policy methods may be more appropriate during exploration, because they can easily encompass uncertainties and allow cautious behavior \citep{Levine2018}. However, there are still many open questions on the exploration-exploitation issues and even more fundamentally there is still a lack of general view on this powerful paradigm, even with known dynamics.   

In some of our previous works, we have proposed various techniques for modeling the motion behaviors of pedestrians and ships \citep{CastaldoWirn2014, CosciaFusion2016, CosciaFusion2018, CosciaIEEE2018, CosciaIVC2018}. More recently, however, in experimenting with probability propagation methods for determining the best path \citep{palmieri2020path}, we came to realize  that the probabilistic algorithms may be the most promising approaches  for agile modeling of intelligent agent motion in complex scenes.  

In this work,  we assume that the system's stochastic transition function is known and that both the state and the action spaces are discrete finite sets that can be handled with tabular methods. Extensions to continuous space can be considered with approximations to the value function, but they will not be addressed here.  Also, we do not address learning here, because we believe that a unified view on the various cost functions and recursions with known stochastic system dynamics should be the first step in trying to understand the more challenging RL adaptation rules. More specifically,  focusing on the path planning problem on a discrete grid, and assuming that the system dynamics are stochastic and  known to the agent, we can systematically compare the value functions and the corresponding  optimal policies that result from various recursions. Standard probability message propagation, such as the Sum-product  and the Max-product  algorithms \citep{barberBRML2012}, are compared to DP using a unified view together with other methods based on joint Reward/Entropy maximization \citep{AI2020, Levine2018, Ziebart2009, Millidge2020, Imohiosen2020}. To our knowledge, no comprehensive comparison exists in the literature, and our contribution aims at providing the reader with a ready-to-use suite of algorithms derived withing a unifying framework. 

The basic idea of this work is to show how we can map the stochastic problem to a factor graph where  different methods correspond to different propagation rules through some of the graph's blocks. Information can travel in the system both in the probability space and in the log-space;  this allows us to include and generalize the previously proposed  methods to new suites of algorithms, thereby increasing our options to condition the agent's policy distribution.

In this paper, we use directed Factor Graphs (FG), that assign variables to edges and factors to interconnected blocks. Generally speaking, message propagation in FG  is more easily handled in comparison to propagation in graphs in which the variables are in the nodes \citep{Koller09}.  Further reduction of the burden of defining message composition rules can be achieved using Factor Graphs in normal form (FGn), proposed by \cite{Forney2001} \citep{Loeliger2004}.  A FGn conveniently includes {\em junction nodes} (equality constraint nodes) that split incoming and outgoing messages when variables are shared by multiple factors. We have proposed a small modification to the FGn in our {\em Factor Graph in Reduced normal form (FGrn)} \citep{Palmieri2016} by including {\em shaded blocks} that map single variables to joint spaces. In fact,  in an oriented graph, when a variable has more than one parent, proper forward and backward messages have to go through the parents' joint space (married parents). In a FGrn, the shaded blocks describe this passage and allow a unique definition of message propagation rules  through {\em Single-Input/Single-Output (SISO)} blocks. Computational complexity issues in some FGrn architectures have been addressed in \citep{DiGennaro2021}.  In the standard sum-product algorithm, backward propagation through shaded blocks corresponds to marginalization. We  will show in this paper how this operation can be generalized and how it  plays a crucial role in the different path planning algorithms. As mentioned above, we confine ourselves here to discrete variables. However, factor graphs that propagate continuous distributions are possible and may be devised also for path planning.  Gaussian messages have been introduced in \cite{Loeliger2007a} and have already been used for Kalman filter-based tracking in \citep{CastaldoAero2015} using FGrn. This issue will not be addressed here and will be the subject of future work.

The contributions of this paper can be summarized as follows: 

\noindent
1. The path planning problem is mapped to a Factor Graph in Reduced Normal Form.  

\noindent
2. Various algorithms, such as the Sum-product, the Max-product, DP and Reward/Entropy maximization (the latter related to structural variational inference), are included in the same framework,  both in probability  and in log spaces. They are all derived using different cost functions, but they are all reduced to specific propagation rules through some of the FGrn blocks. 

\noindent
3. The equivalent Q-functions and V-functions in the probability space,  seen as alternatives to the well-known DP formulation,  allows us to write the policy distribution for all the algorithms with a unique expression. 

\noindent
4. Using this general framework, we extend some of the known algorithms to a whole suite of new parametric updates that can control the smoothness in the policy distributions. These proposed  parametric updates can be used to balance exploration and exploitation in reinforcement learning.    

\noindent
5. We provide simulations, first on a small grid with one goal and obstacles, then on a larger grid extracted from a real scene with  multiple goals (exits) and a semantic map. The results   show some marked differences in : (a) the speed of converge to the steady-state value function, where probabilistic methods are clearly favored; (b) how the Max-product algorithm may be preferred for its faster convergence and for the shape and smoothness of its value functions; (c) how various algorithms can be controlled with parametric updates to exhibit different smoothness in their policy distributions.  

We believe that our contribution in this paper may prove useful for further deployment of RL algorithms, especially when the environment is not completely known and exploration and exploitation have to be properly balanced on the basis of partial model knowledge.    

\noindent
{\em Outline of the paper:} In Section \ref{sec:model}, we present the bayesian model and the corresponding factor graph.  In Section \ref{sec:marginalization}, the  Sum-product algorithm is discussed in the framework of FGrn with the message composition rules and the updates. The inferences are   presented both in a parallel and in a progressive version of the message composition. In Section \ref{sec:maxpost}, the maximum a posteriori solution  of the Max-product algorithm is analyzed with our proposed  Sum/Max-product algorithm described  in 
Section \ref{sec:summax}. Dynamic programming is translated into this framework in Section \ref{sec:dp} and our proposals for a generalized SoftDP are in Section  \ref{sec:softdp}.  
The approaches to combined maximum reward and entropy are discussed in Section \ref{sec:maxrewent}, where a cost function that includes  smooth generalizations is also discussed. Simplifications of some of the recursions when the system equations are deterministic,  are discussed in Section \ref{sec:deterministic}. The extension to infinite horizon models and considerations on the steady-state solutions are included in Section \ref{sec:infhstate}. Simulations are conducted on two types of grids in Section \ref{sec:simulations} and conclusions and suggestions for further research are in Section \ref{sec:conclusions}. The analysis of the soft-max functions used in the paper are in Appendix \ref{app:smax}, and the proofs for the reward/entropy methods are in Appendix \ref{app:levgen}.

\begin{figure}[ht]
  \centering
  \includegraphics[width=0.6\linewidth]{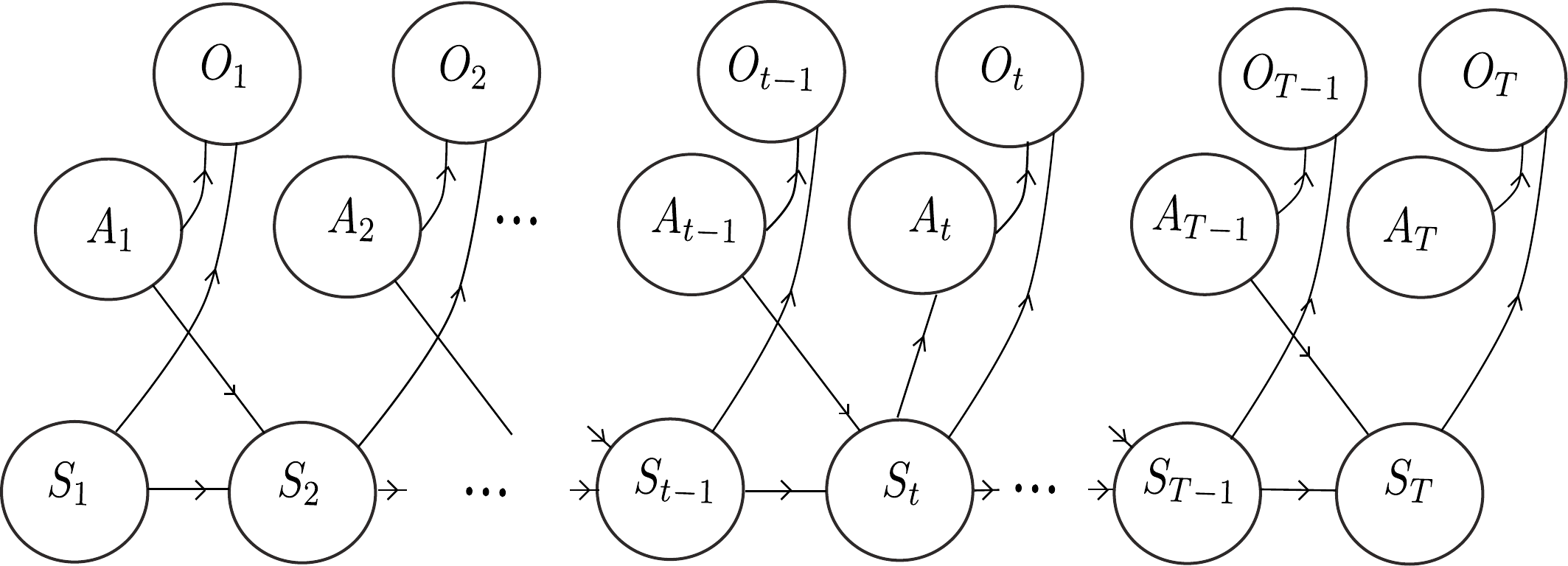} 
  \caption{State-Action Model as a Bayesian graph}
  \label{fig:Bayes}
\end{figure}

\section{The Bayesian Model}
\label{sec:model}

Figure \ref{fig:Bayes} shows the state-action model as a Bayesian graph where $\lbrace S_t \rbrace$ is the {\em state} sequence, 
$\lbrace A_t \rbrace$ is the {\em action} sequence. We assume, without loss of generality, that
both sequences belong to discrete finite sets: $A_t \in {\cal A}$ and $S_t\in {\cal S}$. The reward/outcome sequence $\lbrace O_t \rbrace$ is binary with $O_t \in \lbrace 0, 1 \rbrace$. 

The model evolves over a finite horizon $T$ and the joint probability distribution of the state-action-outcome sequence corresponds to the factorization\footnote{Even if the notation should have capital letters for random variables as subscripts and lower case letters for their values in the functions, as in 
\begin{equation*}
  p_{S_1A_1O_1 \dots S_TA_TO_T}(s_1a_1o_1 \dots s_Ta_To_T); \quad
  p_{S_{t+1}|S_tA_t}(s_{t+1}|s_ta_t); \quad
  p_{O_t|S_tA_t}(o_t|s_ta_t),
\end{equation*}
we  use a compact notation with no subscripts when there is no ambiguity. In some of the messages that  follow we include the subscripts only when necessary.}  
\begin{equation*}
  p(s_1a_1o_1 \dots s_Ta_To_T) = p(o_T|s_Ta_T) p(s_1) p(a_T) \prod_{t=1}^{T-1} p(s_{t+1}|s_ta_t) p(a_t) p(o_t|s_ta_t), 
\end{equation*}
where the function $p(s_{t+1}|s_ta_t)$ describes the {\em system dynamics}, $p(a_t)$ are the action priors and $p(o_t|s_ta_t)$ are the {\em reward/priors} on the state-action pairs. 
More specifically, we assume that 
\begin{equation*}
  P(O_t=1|s_ta_t) \propto c(s_ta_t) \ge 0; \qquad P(O_t=0|s_ta_t) \propto U(s_ta_t),  
\end{equation*}    
where the function $c(s_ta_t)$ acts as a prior distribution on the pair $(s_ta_t)$, only if $O_t=1$. When $O_t=0$, no prior information is available on that state-action pair, and the factor  becomes the uniform distribution $U(s_ta_t)$.\footnote{In our definition, we assume that $c(s_ta_t)$ is normalized to be a valid pdf, even if it is well-known that, in performing inference in a probabilistic graph, normalization is irrelevant.} 

This formulation allows the introduction of a {\em reward} function as
\begin{equation}
  R(s_ta_t)=\log c(s_ta_t) + K,
  \label{eq:rev} 
\end{equation}
where $K$ is an arbitrary positive constant. The value $K$ is really irrelevant because going back to probabilities we have
\begin{equation*}
  c(s_ta_t) \propto e^{R(s_ta_t) - K},
\end{equation*}
with the constant that disappears after normalization. We can set $K$ to a large value if we do not like to handle the negative rewards that we get from the $\log$ function for $K=0$. In the following, without loss of generality, we assume that all our rewards are negative ($K=0$). 

The introduction of the sequence  $\lbrace O_t \rbrace$ has often been proposed for connecting rewards to probabilities \citep{Kappen2013, Levine2018}. We would like to emphasize  that  interpreting the factors $c(s_ta_t)$ as prior information in the probability factorization, may solve, at least for planning problems, the well-known issue of defining an appropriate reward function.  In fact, in a practical problem, we may have available statistics on how often a state is visited and how certain actions may be more likely than others. 

Note that  when a state-action pair has zero probability, for example with forbidden states, or impossible actions,  obstacles, etc., the reward function takes value $-\infty$. This is really not a problem in practice, because we can easily approximate such a value with a large negative number.  

Note that in the model we have included a separate factor $p(a_t)$ for the priors on  $A_t$,  even if such information could be included in $c(s_ta_t)$.  We have preferred, to be consistent with the Bayesian graph of Figure \ref{fig:Bayes}, to keep the two factors separate, one for marginal action priors and one for joint priors (rewards). 
   
Omitting the sequence $\lbrace O_t \rbrace$ in the notation, or equivalently assuming that prior information is always available,  the factorization is more compactly written as 
\begin{equation}
  p(s_1a_1...s_Ta_T)  =
  c(s_Ta_T) p(s_1) p(a_T) \prod_{t=1}^{T-1} p(s_{t+1}|s_ta_t) p(a_t) c(s_ta_t). 
  \label{eq:factmodel}
\end{equation}
This is without loss of generality because, for some values of $t$, the distribution $c(s_ta_t)$ can just be taken to be uniform. 

\begin{figure}[ht]
  \centering
  \includegraphics[width=\linewidth]{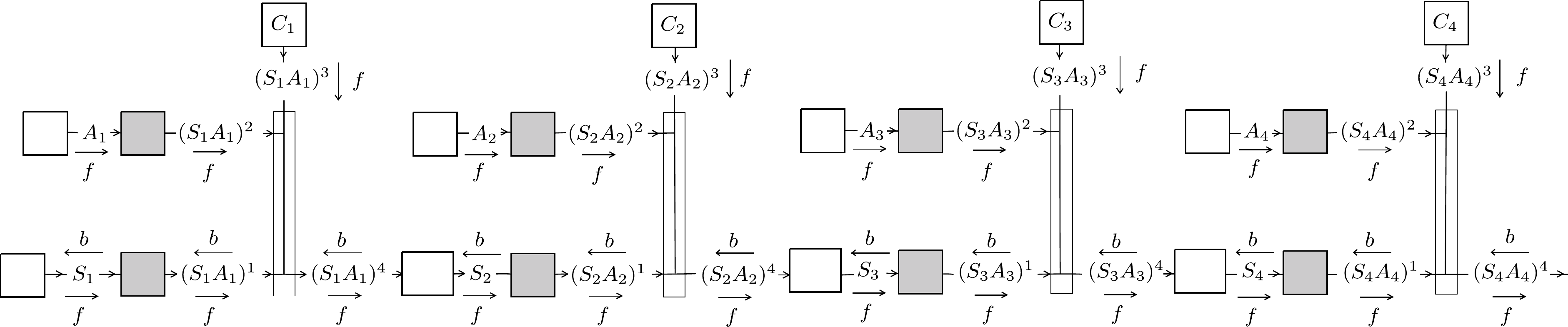} 
  \caption{State-Action Model for $T=4$ as a Factor Graph in Reduced normal form. The one-edge blocks are sources (priors); the two-edge white blocks represent the system dynamics; the shaded blocks map single variables to their joint space; the diverters connect the variables constrained to be equal.}
  \label{fig:FGrn}
\end{figure}

\subsection{The factor graph}

Inference in a graphical model is more easily handled with reference to an equivalent factor graph, where  variables are on the branches and factors are in the blocks. We use here   {\em Factor Graphs in Reduced normal form (FGrn)} \citep{Palmieri2016}. Factor graphs in normal form (FGn) have been introduced by \cite{Forney2001} and discussed by \cite{Loeliger2004},  mostly for coding problems. They allow the probability factorization to be written in terms of a graph in which variables are edges and factors are nodes with {\em diverters} (equality constraints nodes), that multiplex messages to the other factors.  Since, in  a FGn, a factor can still have multiple inputs, we impose a further simplification in the FGrn  by transforming the graph into a configuration that includes {\em shaded blocks} (marginalizers/expanders); this reduces the graph to a composition of only SISO (Single-Input-Single-Output) blocks and diverters. FGrn provide a very detailed display of the message propagation flow and have a simple unique formulation in terms of message propagation rules.
 
Figure \ref{fig:FGrn} shows the factor graph in reduced normal form for the Bayesian model of Figure \ref{fig:Bayes} for $T=4$.  The prior distributions $p(a_t)$ and $c(s_ta_t)$ are in the source nodes and the dynamics $p(s_{t+1}|s_t a_t)$ are in SISO blocks.  The junctions describe equality constraints, and the shaded blocks describe the mapping from single variables' space to a joint space, i.e.  $p(s_ta_t|a_t') = U(s_t) \delta(a_t-a_t')$ and $p(s_ta_t|s_t') = \delta(s_t-s_t') U(a_t)$. Essentially, in a shaded block, the input variable is copied to the output and joined to the other variable that, in that branch, carries no information in the forward direction. Each branch has a direction and a {\em forward} and a {\em backward} message associated to it. Just as in any belief propagation network, all messages are proportional to probability distributions and their composition rules allow the agile derivation of inference algorithms. Note how the replicas $(S_tA_t)^i$, $i=1:4$, of the same variable $(S_tA_t)$ around the diverter block have different names and messages associated with them. 

We will see how the path planning problem has a unique formulation on the factor graph. By changing the propagation rules for some of the blocks, we obtain the optimal solutions for the various problem formulations. 

\subsection{Introducing constraints}

One of the main advantages of studying inference problems on  graphs using messages, is that problem constraints are easily included in the flow. 
Furthermore, we have assumed a finite time segment $t=1:T$, but this model may as well represent a segment $t=t_0+1:t_0+T$ of a longer process, where we have freedom to introduce the initial and final conditions in the forward messages at the beginning, and in the backward messages at the end, respectively.   

For example: 
(a) A known starting state $S_1=\overline{s}_1$ can be included as a forward message 
$f_{S_1}(s_1)=\delta(s_1 - \overline{s}_1)$, where $\delta(x)=1$ if $x=0$, and $0$ else;  
(b) If we have no prior information on $S_1$, we set $f_{S_1}(s_1)=U(s_1 )$; 
(c) Knowledge of the initial action $A_1=\overline{a}_1$ can be included as $f_{A_1}(a_1)=\delta(a_1 - \overline{a}_1)$; 
(d) Knowledge of the final state (only)
$S_T=\overline{s}_T$ is $b_{(S_TA_T)^4}(s_Ta_T)=\delta(s_T - \overline{s}_T) U(a_T)$; 
(e) Knowledge of the state at time $t_0$ may be included as $f_{S_{t_0}}(s_{t_0})=\delta(s_{t_0}-\overline{s}_{t_0})$; 
(f) In a planning problem, a known map $m(s_t)$ can be associated to the factor $c(s_ta_t)$ with 
$f_{(S_tA_t)^3}(s_ta_t) \propto m(s_t) U(a_t)$; (g) In the same  planning problem,  joint map-action information can injected as the message  $f_{(S_tA_t)^3}(s_ta_t) \propto c(s_ta_t)$; if action and map are independent, and the action prior is $p(a_t)$,  $f_{(S_tA_t)^3}(s_ta_t) \propto m(s_t)p(a_t)$, or equivalently $f_{(S_tA_t)^3}(s_ta_t) \propto m(s_t)U(a_t)$ and $f_{A_t}(a_t)=p(a_t)$; etc. 

We denote collectively all the constraints available on the joint model as $K_{1:T}$, with the joint model  written in compact form as $p(s_1a_1 \dots s_Ta_T|K_{1:T})$. 

\subsection{Inference objectives}

Our inference aims at  providing solutions to one or more of  the following problems:
\begin{enumerate}
  \item Find the best {\em state} sequence (S): $s_1^* \rightarrow s_2^* \rightarrow \dots \rightarrow s_T^*$ 
  \item Find the best {\em action} sequence (A): $a_1^* \rightarrow a_2^* \rightarrow \dots \rightarrow a_T^*$
  \item Find the best  {\em joint state-action} sequence (SA) : $(s_1a_1)^* \rightarrow (s_2a_2)^* \rightarrow ... \rightarrow (s_Ta_T)^*$ 
  \item Find the best {\em state-action} sequence (SASA): 
    $s_1^* \rightarrow a_1^* \rightarrow s_2^* \rightarrow a_2^* \rightarrow \dots \rightarrow s_T^* \rightarrow a_T^*$ 
  \item Find the best {\em action-state} sequence (ASAS): 
    $a_1^* \rightarrow s_1^* \rightarrow a_2^* \rightarrow s_2^* \rightarrow \dots \rightarrow a_T^* \rightarrow s_T^*$ 
  \item Find the best {\em policy} distributions (P): $\pi^* (a_1|s_1),  \pi^*(a_2|s_2) \dots \pi^*(a_T|s_T)$ 
\end{enumerate}

We will see in the following how various cost functions determine the message composition rules across the blocks to solve the above problems both in the probability space and in the log-space and according to different optimization criteria.  In the following discussion, we will concentrate mostly on the S sequence, on the SASA sequence and on the policy distribution, since the extensions to A, SA, and ASAS sequences are straightforward.  

\section{Marginalization and the Sum-product}
\label{sec:marginalization}
Standard inference in Bayesian model  consists in marginalizing the total joint probability distribution to obtain  distributions that are proportional to the posteriors on single variables \citep{Koller09}. More specifically, for states only, for actions only and for states and actions jointly, we want to compute the posteriors as
\begin{equation}
  \begin{array}{rL}
    p(s_t|K_{1:T}) & \propto \sum_{\substack{s_j, j\neq t, j=1:T \\ a_j, j=1:T}} p(s_1a_1 \dots s_Ta_T|K_{1:T}), \\
    p(a_t|K_{1:T}) & \propto \sum_{\substack{s_j, j=1:T \\ a_j, j\neq t, j=1:T}} p(s_1a_1 \dots s_Ta_T|K_{1:T}), \\
    p(s_ta_t|K_{1:T}) & \propto \sum_{\substack{s_j,a_j,  j\neq t, j=1:T}} p(s_1a_1 \dots s_Ta_T|K_{1:T}).
  \end{array}
\end{equation}

The {\em policy} distributions are obtained by fixing the state $s_t$ at time $t$, and accounting for the foreseeable future until $T$
\begin{equation}
  \pi^*(a_t|s_t) \triangleq p(a_t|s_t,K_{t:T}) = \frac{p(s_ta_t|K_{t:T})}{p(s_t|K_{t:T})}, \qquad t=1:T. 
\end{equation}
The policy distribution describes at time $t$ how likely it is to take action $a_t$ from state $s_t$, given all the available information (constraints, priors, etc.) about the future ($K_{t:T}$). 

All the above functions can be obtained using forward and backward message propagation using the {\em Sum-product} rule \citep{Koller09, barberBRML2012}. This approach essentially averages over the variables that are eliminated across each SISO block. 
With reference to Figure \ref{fig:FGrn}, just by following the flow,  for some of the backward  messages we have 
\begin{equation}
  \begin{array}{rl}
    b_{(S_tA_t)^4}(s_ta_t) & \propto \sum_{s_{t+1}} p(s_{t+1}|s_ta_t) b_{S_{t+1}}(s_{t+1}), \\
    b_{(S_tA_t)^1}(s_ta_t) & \propto p(a_t) c(s_ta_t) b_{(S_tA_t)^4}(s_ta_t), \\
    b_{(S_tA_t)^2}(s_ta_t) & \propto f_{(S_tA_t)^1}(s_ta_t) c(s_ta_t) b_{(S_tA_t)^4}(s_ta_t), \\
    b_{A_t}(a_t) & = \sum_{s_t} b_{(S_tA_t)^2}(s_ta_t), \\
    b_{S_t}(s_t) & = \sum_{a_t} b_{(S_tA_t)^1}(s_ta_t). 
  \end{array}
  \label{eq:recprob}
\end{equation}
 For some of the forward messages
\begin{equation}
  \begin{array}{rl}
    f_{S_{t+1}}(s_{t+1}) & \propto \sum_{s_ta_t} p(s_{t+1}|s_ta_t) f_{(S_tA_t)^4}(s_ta_t), \\
    f_{(S_tA_t)^1}(s_ta_t) & = f_{S_t}(s_t) U(a_t), \\
    f_{(S_tA_t)^2}(s_ta_t) & = U(s_t)f_{A_t}(a_t), \\
    f_{(S_tA_t)^4}(s_ta_t) & \propto f_{(S_tA_t)^1}(s_ta_t) f_{(S_tA_t)^2}(s_ta_t) c(s_ta_t). \\
  \end{array}
\end{equation}
Note that going backward through a block, the message may not be normalized. Around the diverters, outgoing messages are the product of the incoming ones, and are not normalized. Message composition rules are summarized in Tables \ref{tab:source}, \ref{tab:shadedA}, \ref{tab:shadedS}, \ref{tab:dynamics} and \ref{tab:diverter} for the Sum-product and for the other approaches that will be discusses later.  

Posterior distributions are obtained by taking the product of forward and backward messages
\begin{equation}
  \begin{array}{rl}
    p(s_t|K_{1:T}) & \propto f_{S_t}(s_t) b_{S_t}(s_t), \\
    p(a_t|K_{1:T}) & \propto f_{A_t}(a_t) b_{A_t}(a_t), \\
    p(s_ta_t|K_{1:T}) & \propto f_{(S_tA_t)^i}(s_ta_t) b_{(S_tA_t)^i}(s_ta_t), \qquad \text{for any} \; i=1,2,3,4.
  \end{array}
\end{equation}
For readers not too familiar with probability message propagation, we would like to emphasize that this framework is a rigorous application of Bayes' theorem and marginalization. Also all messages can be normalized to be valid  distributions, even if it is not strictly necessary (it is their shape that matters). However, it is often advised to keep messages normalized for numerical stability. 

The policy distribution at each $t$ is derived as a consequence of the inference obtained  from the probability flow and is 
\begin{equation}
  \pi^*(a_t|s_t) \propto \frac{f_{(S_tA_t)^1}(s_ta_t) b_{(S_tA_t)^1}(s_ta_t)}{f_{S_t}(s_t) b_{S_t}(s_t)} = 
  \frac{f_{S_t}(s_t) U(a_t) b_{(S_tA_t)^1}(s_ta_t)}{f_{S_t}(s_t) b_{S_t}(s_t)} = \frac{b_{(S_tA_t)^1}(s_ta_t)}{b_{S_t}(s_t)},
  \label{eq:policyps}
\end{equation}
where we have used the branch with $i=1$. It is easy to verify that the solution would have an equivalent expression for any other branch. Note also how the policy depends only on the backward messages. The reason for this is that by conditioning on $s_t$, all the information coming from the left side of the graph is blocked. 

\subsection{Max Posterior sequences}
\label{subsec:maxpost}

Optimal sequence values  can be obtained in parallel using maximization on the posteriors 
\begin{equation}
  \begin{array}{rLl}
    s_t^* & = \argmax_{s_t} p(s_t|K_{1:T}) = \argmax_{s_t} f_{S_t}(s_t) b_{S_t}(s_t), & \\
    a_t^* & = \argmax_{a_t} p(a_t|K_{1:T}) = \argmax_{a_t} f_{A_t}(a_t) b_{A_t}(a_t), & \qquad t=1:T \\
    (s_ta_t)^* & = \argmax_{s_ta_t} p(s_ta_t|K_{1:T}) = \argmax_{s_ta_t} f_{(S_tA_t)^1}(s_ta_t) b_{(S_tA_t)^1}(s_ta_t), &
  \end{array}
\end{equation}
The max posterior solutions are taken separately on each variable and,  even if they are often used in the applications (for example in decoding convolutional codes - the algorithm, is named BCJR after its authors \cite{Bahl1974}),  they may provide unsatisfactory sequences for path planning. In fact, the sequences that result from the maximizations are unconstrained in time and may correspond to disconnected paths \citep{palmieri2020path}. 
  
\subsection{Progressive Max Posterior sequences}
\label{subsec:maxppost}
Better solutions for the max posterior approach are obtained progressively in time following a forward procedure.\footnote{On a fixed time horizon, a similar procedure can be derived going backward in time. We prefer to maintain the framework causal and leave it out for brevity.}

For the states-only (S) sequence
\begin{equation}
  \begin{array}{rL}
    s_1^* & = \argmax_{s_1} p(s_1|K_{1:T}) = \argmax_{s_1} f_{S_1}(s_1) b_{S_1}(s_1), \\
    s_2^* & = \argmax_{s_2} p(s_1^*s_2|K_{2:T}) = \argmax_{s_2} f_{S_2}(s_2|s_1^*) b_{S_2}(s_2), \\
    s_3^* & = \argmax_{s_3} p(s_1^*s_2^* s_3|K_{3:T}) = \argmax_{s_3} f_{S_3}(s_3|s_2^*) b_{S_3}(s_3), \\
    \cdots & \\
    s_t^* & = \argmax_{s_t} p(s_1^*...s_{t-1}^*s_t|K_{t:T}) = \argmax_{s_t} f_{S_t}(s_t|s_{t-1}^*) b_{S_{t-1}}(s_{t-1}), \\
    \cdots &
  \end{array}
\end{equation}
where the conditioned forward messages come from a  one-step propagation 
\begin{equation}
  \begin{array}{rL}
    f_{S_t}(s_t|s_{t-1}^*) & = \sum_{a_{t-1}} p(s_t|s_{t-1}^*a_{t-1}) f_{(S_{t-1}A_{t-1})^4}(s_{t-1}^*a_{t-1}) \\
    & = \sum_{a_{t-1}} p(s_t|s_{t-1}^*a_{t-1}) p(a_{t-1}) c(s_{t-1}^*a_{t-1}).
  \end{array}
\end{equation} 
 Note on the graph that knowledge of the state at time $t-1$ "breaks" the forward flow. Only the backward flow drives the inference. 
 
Similarly, for the best State-Action (SASA) sequence, the {\em Progressive Max-posterior} algorithm using the messages on the graph is  
\begin{equation}
  \begin{array}{rL}
    s_1^* & = \argmax_{s_1} p(s_1|K_{1:T}) = \argmax_{s_1} f_{S_1}(s_1) b_{S_1}(s_1), \\
    a_1^* & = \argmax_{a_1} p(s_1^* a_1|K_{1:T}) = \argmax_{a_1} f_{A_1}(a_1) b_{A_1}(a_1|s_1^*), \\
    s_2^* & = \argmax_{s_2} p(s_1^*a_1^*s_2|K_{2:T}) = \argmax_{s_2} f_{S_2}(s_2|s_1^*a_1^*) b_{S_2}(s_2), \\
    a_2^* & = \argmax_{a_2} p(s_1^* a_1^*s_2^*a_2|K_{2:T}) = \argmax_{a_2} f_{A_2}(a_2) b_{A_2}(a_2|s_2^*), \\
    \cdots & \\
    s_t^* & = \argmax_{s_t} p(s_1^*a_1^*...s_{t-1}^*a_{t-1}^*s_t|K_{t:T}) = \argmax_{s_t} f_{S_t}(s_t|s_{t-1}^*a_{t-1}^*) b_{S_t}(s_t), \\
    a_t^* & = \argmax_{a_t} p(s_1^*a_1^*...s_{t-1}^*a_{t-1}^*s_t^* a_t|K_{t:T}) = \argmax_{a_t} f_{A_t}(a_t) b_{A_t}(a_t|s_t^*), \\
    \cdots &
  \end{array}
\end{equation}
where the conditioned forward and backward messages mean that we have considered their values when the conditioning variables on the left side of the graph are fixed. For the conditioned forward we have  
\begin{equation}
f_{S_t}(s_t|s_{t-1}^* a_{t-1}^*)  = p(s_t|s_{t-1}^* a_{t-1}^*). 
\end{equation}  
For the conditioned backward we have  
\begin{equation}
b_{A_t}(a_t|s_t^*) \propto b_{(S_tA_t)^2}(s_t^*a_t) \propto c(s_t^*a_t) b_{(S_tA_t)^4}(s_t^*a_t) f_{(S_tA_t)^1}(s_t^*a_t).
\end{equation} 
that since 
\begin{equation}
b_{(S_tA_t)^1}(s_ta_t) \propto b_{(S_tA_t)^4}(s_ta_t) f_{(S_tA_t)^2}(s_ta_t)c(s_ta_t)= b_{(S_tA_t)^4}(s_ta_t) p(a_t) U(s_t) c(s_ta_t),   
\end{equation}
or
\begin{equation}
b_{(S_tA_t)^4}(s_ta_t) \propto \frac{b_{(S_tA_t)^1}(s_ta_t)}{p(a_t) c(s_ta_t)},   
\end{equation}
can be rewritten as 
\begin{equation}
  b_{A_t}(a_t|s_t^*)  \propto \frac{b_{(S_tA_t)^1}(s_t^*a_t)}{p(a_t)} f_{(S_tA_t)^1}(s_t^*a_t) 
  = \frac{b_{(S_tA_t)^1}(s_t^*a_t)}{p(a_t)} \delta(s_t-s_t^*) U(a_t) = \frac{b_{(S_tA_t)^1}(s_t^*a_t)}{p(a_t)}. 
\end{equation}

Therefore, the  SASA estimation simplifies to 
\begin{equation}
  \begin{array}{rL}
    s_1^* & = \argmax_{s_1} f_{S_1}(s_1) b_{S_1}(s_1), \\
    a_1^* & = \argmax_{a_1} b_{(S_1A_1)^1} (s_1^*a_1), \\
    s_2^* & = \argmax_{s_2} p(s_2|s_1^*a_1^*) b_{S_2}(s_2), \\
    a_2^* & = \argmax_{a_2} b_{(S_2A_2)^1} (s_2^*a_2), \\
    \cdots & \\
    s_t^* & = \argmax_{s_t} p(s_t|s_{t-1}^*a_{t-1}^*) b_{S_t}(s_t), \\
    a_t^* & = \argmax_{a_t} b_{(S_tA_t)^1} (s_t^*a_t), \\
    \cdots &
  \end{array}
\end{equation}
which is obviously the same if we use explicitly the optimal policy distribution (\ref{eq:policyps})
\begin{equation}
  \begin{array}{rL}
    s_1^* & = \argmax_{s_1} p(s_1|K_{1:T}) = \argmax_{s_1} f_{S_1}(s_1) b_{S_1}(s_1), \\
    a_1^* & = \argmax_{a_1} \pi^*(a_1|s_1^*), \\
    s_2^* & = \argmax_{s_2} p(s_2|s_1^*a_1^*) b_{S_2}(s_2), \\
    a_2^* & = \argmax_{a_2} \pi^*(a_2|s_2^*), \\
    \cdots & \\
    s_t^* & = \argmax_{s_t} p(s_t|s_{t-1}^*a_{t-1}^*) b_{S_t}(s_t), \\
    a_t^* & = \argmax_{a_t} \pi^*(a_t|s_t^*), \\
    \cdots &
  \end{array}
  \label{eq:DecMean}
\end{equation}
Note in all cases the crucial role played by the backward flow. We have successfully demonstrated this approach for path planning in our previous work \citep{palmieri2020path}.  In fact, in the progressive max posterior algorithm, the forward flow is not necessary.  Action-only sequences and ASAS sequence can be obtained in a similar fashion and are omitted here for brevity. 

\begin{table}[ht]
  \caption{Summarized backup rules in probability space with $b(s_ta_t) \triangleq b_{(S_tA_t)^1} (s_ta_t)$; $b(s_t) \triangleq b_{S_t} (s_t)$; $ c'(s_ta_t) \triangleq p(a_t) c(s_ta_t)$.}
  \label{tab:prob}
  \centering
  \begin{tabular}{@{}m{1in}M{2.7in}M{1.3in}} \toprule
    & b(s_ta_t)  & b(s_t)  \\ \cmidrule(l){2-3}
    Sum product & 
      c'(s_ta_t) \sum_{s_{t+1}} p(s_{t+1}|s_ta_t) b(s_{t+1}) & 
      \sum_{a_t} b(s_ta_t)  \\
    Max product & 
     c'(s_ta_t) \max_{s_{t+1}} p(s_{t+1}|s_ta_t) b(s_{t+1}) & 
      \max_{a_t} b(s_ta_t)  \\ \addlinespace
    Sum/Max product $(\alpha\ge1)$ & 
      c'(s_ta_t) \sqrt[\alpha]{\sum_{s_{t+1}} p(s_{t+1}|s_ta_t)^\alpha b(s_{t+1})^\alpha} & 
      \sqrt[\alpha]{\sum_{a_t} b(s_ta_t)^\alpha}  \\ \addlinespace
    DP & 
      c'(s_ta_t) e^{\sum_{s_{t+1}} p(s_{t+1}|s_ta_t) \log b(s_{t+1})} & 
      \max_{a_t} b(s_ta_t)  \\ \addlinespace
    Max-Rew/Ent \newline$(\alpha>0)$ & 
      c'(s_ta_t) e^{\sum_{s_{t+1}} p(s_{t+1}|s_ta_t) \log b(s_{t+1})} & 
      \sqrt[\alpha]{\sum_{a_t} b(s_ta_t)^\alpha}  \\ \addlinespace
    SoftDP $(\beta>0)$ & 
      c'(s_ta_t) e^{\sum_{s_{t+1}} p(s_{t+1}|s_ta_t) \log b(s_{t+1})} & 
      e^{\frac{\sum_{a_t} b(s_ta_t)^\beta \log b(s_ta_t)}{\sum_{a'_t} b(s_ta'_t)^\beta}}  \\ \bottomrule
  \end{tabular}
\end{table}

\begin{table}[ht]
  \caption{Summarized backup rules in log space with $Q(s_ta_t) \triangleq Q_{(S_tA_t)^1} (s_ta_t)$; $V(s_t) \triangleq V_{S_t} (s_t)$; $R'(s_ta_t) = \log p(a_t) + R(s_ta_t)$.}
  \label{tab:log}
  \centering
  \begin{tabular}{@{}m{1in}M{2.7in}M{1.3in}} \toprule
    & Q(s_ta_t) & V(s_t)  \\ \cmidrule(l){2-3}
    Sum product & 
     R'(s_ta_t) + \log \sum_{s_{t+1}} e^{\log p(s_{t+1}|s_ta_t) + V(s_{t+1})} &
      \log \sum_{a_t} e^{Q(s_ta_t)}  \\
    Max product & 
      R'(s_ta_t) + \max_{s_{t+1}} \left(\log p(s_{t+1}|s_ta_t) + V(s_{t+1})\right) & 
      \max_{a_t} Q(s_ta_t)  \\ \addlinespace
    Sum/Max product $(\alpha\ge1)$ & 
      R'(s_ta_t) + \frac{1}{\alpha} \log \sum_{s_{t+1}} e^{\alpha\left(\log p(s_{t+1}|s_ta_t) + V(s_{t+1})\right)} & 
      \frac{1}{\alpha} \log \sum_{a_t} e^{\alpha Q(s_ta_t)}  \\ \addlinespace
    DP & 
      R'(s_ta_t) + \sum_{s_{t+1}} p(s_{t+1}|s_ta_t) V(s_{t+1}) & 
      \max_{a_t} Q(s_t,a_t)  \\ \addlinespace
    Max-Rew/Ent \newline$(\alpha>0)$ & 
      R'(s_ta_t) + \sum_{s_{t+1}} p(s_{t+1}|s_ta_t) V(s_{t+1}) & 
      \frac{1}{\alpha} \log \sum_{a_t} e^{\alpha Q(s_ta_t)}  \\ \addlinespace
    SoftDP $(\beta>0)$ & 
      R'(s_ta_t) + \sum_{s_{t+1}} p(s_{t+1}|s_ta_t) V(s_{t+1}) & 
      \frac{\sum_{a_t} Q(s_ta_t) e^{\beta Q(s_ta_t)}}{\sum_{a'_t} e^{\beta Q(s_ta'_t)}}  \\ \bottomrule
  \end{tabular}
\end{table}

\subsection{Sum-product in the log-space}

We have seen above how in the factorized model (\ref{eq:factmodel}),  prior distributions  are related to rewards via the  log transformation in (\ref{eq:rev}). For the comparisons that follow, it is convenient  to consider also some of  the Sum-product recursions in the log-space. We define the following functions
\begin{equation}
  \begin{array}{rl}
    Q_{(S_tA_t)^i}(s_ta_t) & \triangleq \log b_{(S_tA_t)^i}(s_ta_t),  \qquad i=1,2,3,4 \\
    V_{S_t}(s_t) & \triangleq \log b_{S_t}(s_t),
  \end{array}
\end{equation} 
Note that there is a $Q$ function for each message around the diverter. The choice of the notations $Q$ ({\em $Q$-function}) and $V$ ({\em Value-function}) is not casual, as it will be clear in the discussion that will follow on dynamic programming.  There is also a definition of $V$-function for the action variable $A_t$, $V_{A_t}(a_t) \triangleq \log b_{A_t}(a_t)$.
From the definition, is obvious that both the $Q$- and $V$-functions are negative (we have already pointed out above that this is not a limitation).   We concentrate here mostly on the state $S_t$ for which  the backward recursions  in (\ref{eq:recprob}), are written in the log-space as
\begin{equation}
  \begin{array}{rl}
    Q_{(S_tA_t)^4}(s_ta_t) & \propto \log \sum_{s_{t+1}} p(s_{t+1}|s_ta_t) e^{V_{S_{t+1}}(s_{t+1})}, \\
    Q_{(S_tA_t)^1}(s_ta_t) & \propto \log p(a_t) + R(s_ta_t) + Q_{(S_tA_t)^4}(s_ta_t), \\
    V_{S_t}(s_t) & \propto \log \sum_{a_t} e^{Q_{(S_tA_t)^1}(s_ta_t)}.  
  \end{array}
  \label{eq:reclog}
\end{equation}
All messages can be propagated in the  log-space: the product rule around the diverter of Figure \ref{fig:FGrn} becomes a sum and the backward propagation rules across the dynamics block and the shaded block are simply translated. 
For better comparison with the formulations that follow, we re-write the first equation of (\ref{eq:reclog}) as 
\begin{equation}
  Q_{(S_tA_t)^4}(s_ta_t) \propto \log \sum_{s_{t+1}} e^{\log p(s_{t+1}|s_ta_t) +V_{S_{t+1}}(s_{t+1})}.
\end{equation}
The recursions  are summarized for comparison in the first row of Tables \ref{tab:prob} and \ref{tab:log}. 

The same recursions, and some of the definitions in the log-space, have been reported by \cite{Levine2018} that also notes how the transformation $y=\log \sum_{j=1}^Ne^{x_j}$ is a {\em soft-max} ($y \sim \max (x_1,...,x_N)$ when the $x_i$s are large), in contrast to the the hard {\it max} that is used in dynamic programming.   Appendix \ref{app:smax}  summarizes the  properties of the soft-max functions that arise in our analyses.  

The best SASA sequence of equations (\ref{eq:DecMean}) is equivalently written in the log-space as
\begin{equation}
  \begin{array}{rL}
    s_1^* & = \argmax_{s_1} p(s_1|K_{1:T}) = \argmax_{s_1} \log p(s_1) + V_{S_1}(s_1), \\
    a_1^* & = \argmax_{a_1} Q_{(S_1A_1)^1}(a_1,s_1^*), \\
    s_2^* & = \argmax_{s_2} \log p(s_2|s_1^*a_1^*) + V_{S_2}(s_2), \\
    a_2^* & = \argmax_{a_2} Q_{(S_2A_2)^1}(a_1,s_1^*), \\
    \cdots & \\
    s_t^* & = \argmax_{s_t} \log p(s_t|s_{t-1}^*a_{t-1}^*) + V_{S_t}(s_t), \\
    a_t^* & = \argmax_{a_t} Q_{(S_tA_t)^1}(a_1,s_1^*), \\
    \cdots &
  \end{array}
  \label{eq:DecMeanLog}
\end{equation}
The policy distribution (\ref{eq:policyps}) is rewritten as
\begin{equation}
  \pi^*(a_t|s_t) \propto e^{Q_{(S_tA_t)^1}(s_ta_t)- V_{S_t}(s_t)}.
\end{equation}

\begin{figure}[ht]
  \centering
  \includegraphics[width=\linewidth]{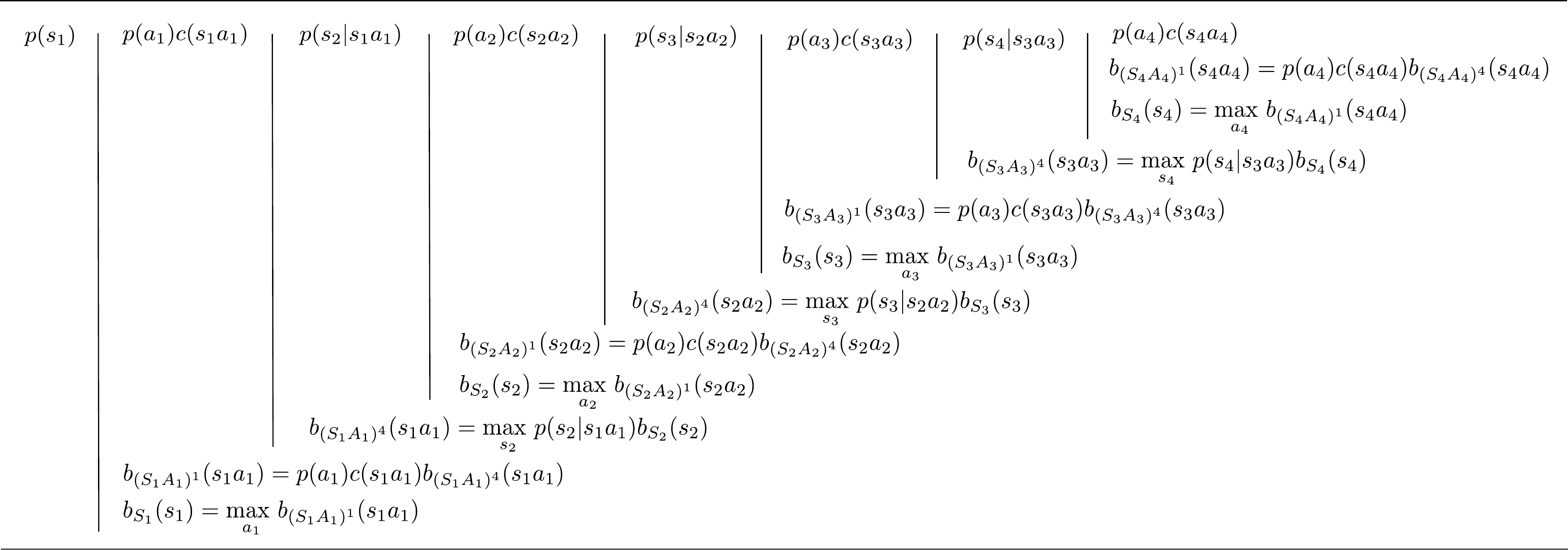} 
  \caption{Backward recursions for the Max-product algorithm for $T=4$. Note the presence of the backward message $b_{(S_4A_4)^4}(s_4a_4)$ at the end of the chain that may carry information from further steps or may represent final constraints.}
  \label{fig:recmax}
\end{figure}

\section{Maximum a posteriori and the Max-product}
\label{sec:maxpost}

The max posterior rules, described in Section \ref{subsec:maxpost} and \ref{subsec:maxppost}, are used extensively for inference in Bayesian networks, even if it is often ignored that they do not necessarily solve the global maximum a posteriori problem 
\begin{equation}
  (s_1^*a_1^* \dots s_T^*a_T^*) = \argmax_{s_1a_1...s_Ta_T} p(s_1a_1 ... s_T a_T|K_{1:T}).
\end{equation}
The Sum-product propagation rules solve {\em marginal} maximum a posteriori problems after  averaging on the eliminated variables, while the global optimization requires a different strategy for obtaining the solution. The {\em Max-product} algorithm \citep{barberBRML2012, Loeliger2007a}, in propagating messages in the graph, instead of computing averages across the blocks, propagates maxima values, provides  the solution. This is often named bi-directional Viterbi algorithm \citep{barberBRML2012}.  The detailed recursions  are  derived explicitely in Figure \ref{fig:recmax} for a model with $T=4$ in the notation of the factor graph of Figure \ref{fig:FGrn}. 
At a generic step $t$, the recursions for some of the backward messages are    
\begin{equation}
  \begin{array}{rL}
    b_{(S_tA_t)^4}(s_ta_t) & = \max_{s_{t+1}} p(s_{t+1}|s_ta_t) b_{S_{t+1}}(s_{t+1}), \\
    b_{(S_tA_t)^1}(s_ta_t) & = p(a_t) c(s_ta_t) b_{(S_tA_t)^4}(s_ta_t), \\
    b_{S_t}(s_t) & = \max_{a_t} b_{(S_tA_t)^1}(s_ta_t). \\
  \end{array}
  \label{eq:recmax}
\end{equation}
Again the crucial role is played by the backward flow that, going through each SISO block, does not undergo a summation, but a max (in Max-product bayesian networks also the forward flow is computed using max rather that sum \citep{barberBRML2012}; we focus here mostly on the backward flow).  In the log-space, the backward recursions for the states are rewritten as 
\begin{equation}
  \begin{array}{rL}
    Q_{(S_tA_t)^4}(s_ta_t) & = \max_{s_{t+1}} \left[ \log p(s_{t+1}|s_ta_t) + V_{S_{t+1}}(s_{t+1}) \right], \\
    Q_{(S_tA_t)^1}(s_ta_t) & = \log p(a_t)+R(s_ta_t)+Q_{(S_tA_t)^4}(s_ta_t), \\
    V_{S_t}(s_t) & = \max_{a_t} Q_{(S_tA_t)^1}(s_ta_t). \\
  \end{array}
\end{equation}
 The best SASA sequence is computed in the forward direction in way similar to the Sum-product,  in both the probability space and in the log-space,  as follows 

At $t=1$, we have 
\begin{equation}
  \begin{array}{rL}
    s_1^* & = \argmax_{s_1} p(s_1) b_{S_1}(s_1) = \argmax_{s_1} p(s_1) e^{V_{S_1}(s_1)} =\argmax_{s_1} \log p(s_1) + V_{S_1}(s_1), \\
    a_1^* & = \argmax_{a_1} b_{(S_1A_1)^1}(s_1^*a_1) = \argmax_{a_1} Q_{(S_1A_1)^1}(s_1^*a_1).
  \end{array}
\end{equation}
At $t=2$
\begin{equation}
  \begin{array}{rLL}
    s_2^* & = \argmax_{s_2} p(s_2|s_1^*a_1^*) b_{S_2}(s_2) & = \argmax_{s_2} p(s_2|s_1^*a_1^*) e^{V_{S_2}(s_2)} \\
    & & = \argmax_{s_2} \log p(s_2|s_1^*a_1^*) + V_{S_2}(s_2), \\
    a_2^* & = \argmax_{a_2} b_{(S_2A_2)^1}(s_2^*a_2) & = \argmax_{a_2} Q_{(S_2A_2)^1}(s_2^*a_2),
  \end{array}
\end{equation}
 ... etc. 

Note how the forward recursions are formally identical to the ones derived for the Sum-product algorithm. 
Also, the policy has the same formal expression  
\begin{equation}
  \pi^*(a_t|s_t) \propto \frac{b_{(S_tA_t)^1}(s_ta_t)}{b_{S_t}(s_t)} = e^{Q_{(S_tA_t)^1}(s_ta_t) - V_{S_t}(s_t)}.
  \label{eq:policy}
\end{equation}
Clearly, the $Q$- and the $V$-functions here have a different meaning. The recursions for the best SASA sequence can be rewritten in terms of policy and they look formally identical to the ones derived for the Sum-product algorithm.   All the other sequences, S, A, SA, ASAS can be computed using the probability flow in the graph following the same formal approach, both in the Max-product and in the Sum-product, simply by changing some of the propagation rules. For brevity, we concentrate here only on some of the messages, but a detailed analysis of other parts of the flow may reveal interesting aspects of the inference. 

Tables \ref{tab:source}, \ref{tab:shadedA}, \ref{tab:shadedS}, \ref{tab:dynamics} and \ref{tab:diverter} summarize the propagation rules across the factor graph for the Sum-product, the Max-product and all the other approaches that will follow.  The main backup recursions are also summarized for comparison in Tables \ref{tab:prob} and \ref{tab:log}  in the log-space. 

We would like to emphasize that propagating information via probability distributions includes all the cases in which there may be deterministic values in the system, i.e., when the distributions are delta functions.  Furthermore,  in the Max-product algorithm, when  multiple equivalent maxima are present, the distributions can carry implicitly multiple peaks. We will see, in some of the simulation examples that will follow, that the Max-product messages provide a complete set of options in the policy distributions, also when more than one best action is available. Obviously, in writing the algorithms,  some attention must be devoted to unnormalized distributions with values that are close zeros, to avoid numerical problems. The problem is usually overcame by normalization and by replacing zeros with very small values.     

\section{The Sum/Max-Product}
\label{sec:summax}

The unifying view provided by the graphical method, both in the Sum-product and in the Max-product approaches, is quite appealing and one wonders whether there may be a general rule that encompasses both. To examine this,  by looking at the recursions for the Sum-product algorithm (Tables \ref{tab:prob} and \ref{tab:log}, first two rows), we immediately observe that the Sum-product, both in the probability and in the log-space, can be seen as a {\em soft version} of the Max-product because of the soft-max functions. Therefore, we propose  a general rule that interpolates between the two solutions using the parametrized soft-max functions discussed in Appendix \ref{app:smax}. We name this generalization the Sum/Max-product algorithm, that in  the log-space gives 
\begin{equation}
  \begin{array}{rL}
    Q_{(S_tA_t)^4}(s_ta_t) & = \frac{1}{\alpha} \log \sum_{s_{t+1}} e^{\alpha \left[ \log p(s_{t+1}|s_ta_t) + V_{S_{t+1}}(s_{t+1}) \right]}, \\
    Q_{(S_tA_t)^1}(s_ta_t) & = \log p(a_t) + R(s_ta_t) + Q_{(S_tA_t)^4}(s_ta_t), \\
    V_{S_t}(s_t) & = \frac{1}{\alpha} \log \sum_{a_t} e^{\alpha Q_{(S_tA_t)^1}(s_ta_t)}, 
  \end{array} 
  \label{eq:1}
\end{equation} 
with $\alpha \ge 1$. In probability space, the updates are immediately translated as
\begin{equation}
  \begin{array}{rL}
    b_{(S_tA_t)^4}(s_ta_t) & \propto \left[ \sum_{s_{t+1}} p(s_{t+1}|s_ta_t)^\alpha  b_{S_{t+1}}(s_{t+1})^\alpha \right]^\frac{1}{\alpha}, \\
    b_{(S_tA_t)^1}(s_ta_t) & \propto p(a_t) c(s_ta_t) b_{(S_tA_t)^4}(s_ta_t), \\
    b_{S_t}(s_t) & \propto \left[ \sum_{a_t} b_{(S_tA_t)^1}(s_ta_t)^{\alpha} \right]^\frac{1}{\alpha}. 
  \end{array}
  \label{eq:2}
\end{equation} 
Note that the function that emerges in the probability space recursions, is also a soft-max. Therefor, both in the log-space and in the probability space, for $\alpha \to \infty$,  the parametric soft-max functions converges to the hard max (see Appendix \ref{app:smax} for details about the soft-max functions). For $\alpha=1$, equations (\ref{eq:1}) and (\ref{eq:2}) become identical to those derived for the Sum-product algorithm.  The Max-product approach usually produces much more defined value functions and policies, in comparison to the Sum-product, as  will be shown in some of the examples that  follow. Interpolating between the two solutions may provide the planning problem with a whole range of new solutions beyond the traditional Sum-product and  Max-product approaches. The Sum/Max-product updates are added as the third row in Tables \ref{tab:prob} and \ref{tab:log} and the detailed block propagation rules are included in  Tables \ref{tab:source}, \ref{tab:shadedA}, \ref{tab:shadedS}, \ref{tab:dynamics} and \ref{tab:diverter}. The policy is formally the same as in the Sum-product and the Max-product. Evidently, the messages, both in the probability space and in the log-space, carry different information.  

\subsection{What function is being optimized?}

The generalization of the Sum/Max-product has been derived as a straightforward interpolation between the Sum-product and the Max-product and such a function can span the whole range of solutions between 
the maximization of the marginals of the Sum-product algorithm to the maximization of the global posterior of the Max-product. What is then, for each value of the parameter $\alpha$, the function that the algorithm optimizes ? 
 
\begin{figure}[ht]
  \centering
  \includegraphics[width=\linewidth]{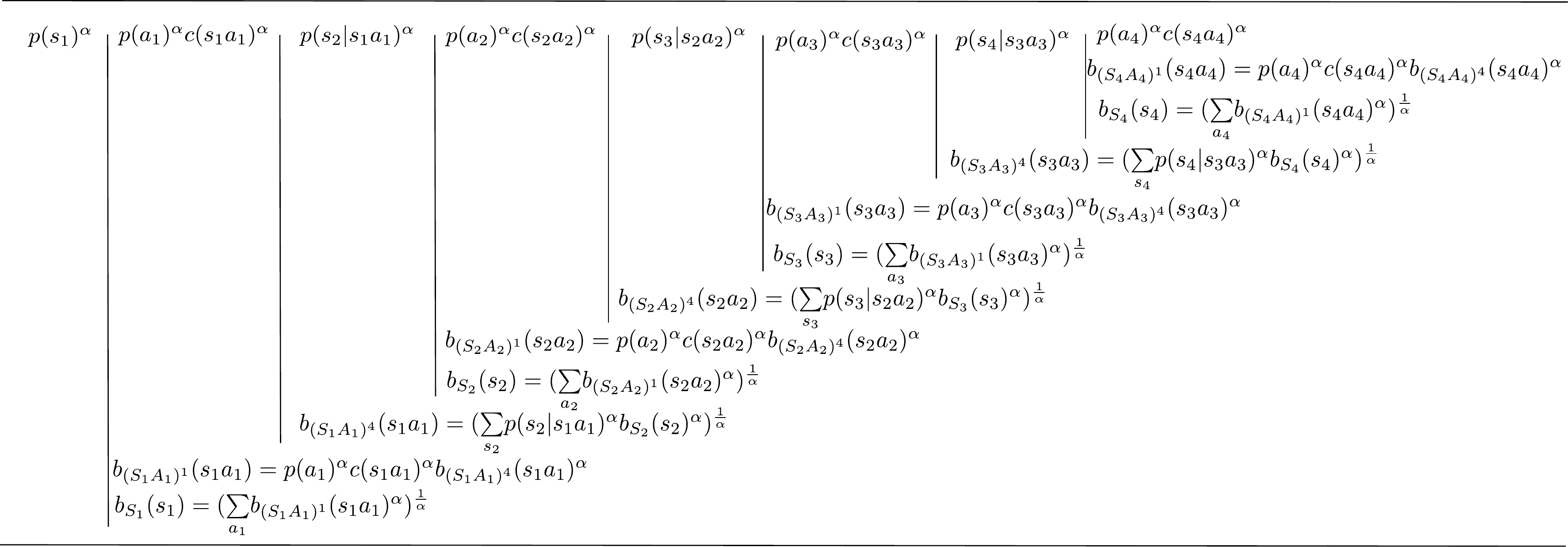} 
  \caption{Backward recursions for the Sum/Max-product algorithm for $T=4$. Note how the recursions can be seen as the Sum-product algorithm applied to the factorization where all the factors are raised to a power $\alpha$.}
  \label{fig:recsummax}
\end{figure}

In the lower part of Figure \ref{fig:recsummax}, we have reported  the recursions of the Sum/Max-product algorithm in the probability space for $T=4$. It is easily seen, by looking at the top of the same figure, that they match the recursions of the Sum-product algorithm as applied to the factorization
\begin{equation}
  p(s_1a_1 \dots s_Ta_T)^\alpha =
  c(s_Ta_T)^\alpha p(s_1)^\alpha p(a_T)^\alpha \prod_{t=1}^{T-1} p(s_{t+1}|s_ta_t)^\alpha  p(a_t)^\alpha c(s_ta_t)^\alpha,  
  \label{eq:factalpha}
\end{equation}
Obviously the power of a distribution is not a normalized distribution, but this is not a problem as we mentioned before, because normalization is just a scale that is irrelevant for the inference. 
 
Therefore, in analogy to the Sum-product algorithm,  the  Sum/Max-product algorithm provides the posteriors
\begin{equation}
  \begin{array}{rL}
    p(s_t|K_{1:T}) & \propto \sum_{\substack{s_j, j\neq t, j=1:T \\ a_j, j=1:T}} p(s_1a_1 \dots s_Ta_T|K_{1:T})^\alpha, \\
    p(a_t|K_{1:T}) & \propto \sum_{\substack{s_j, j=1:T \\ a_j, j\neq t, j=1:T}} p(s_1a_1 \dots s_Ta_T|K_{1:T})^\alpha, \\
    p(s_ta_t|K_{1:T}) & \propto \sum_{\substack{s_j,a_j,  j\neq t, j=1:T}} p(s_1a_1 \dots s_Ta_T|K_{1:T})^\alpha.
  \end{array}
  \label{eq:sumaxpost}
\end{equation}
To better explain the generalization, we recall that raising a probability distribution to a power greater than one, has the effect of sharpening the distribution around its maximum (or maxima, if multiple maxima are present). Therefore, raising the whole joint density to a large power has the effect of concentrating it on the global maximum a posteriori solution of the Max-product algorithm.  Note that the maxima on the posteriors can be computed in parallel, or progressively in sequence. The specific discussion is  omitted here for brevity, but follows the same strategy used for the Sum-product algorithm.   

\begin{figure}[ht]
  \centering
  \includegraphics[width=\linewidth]{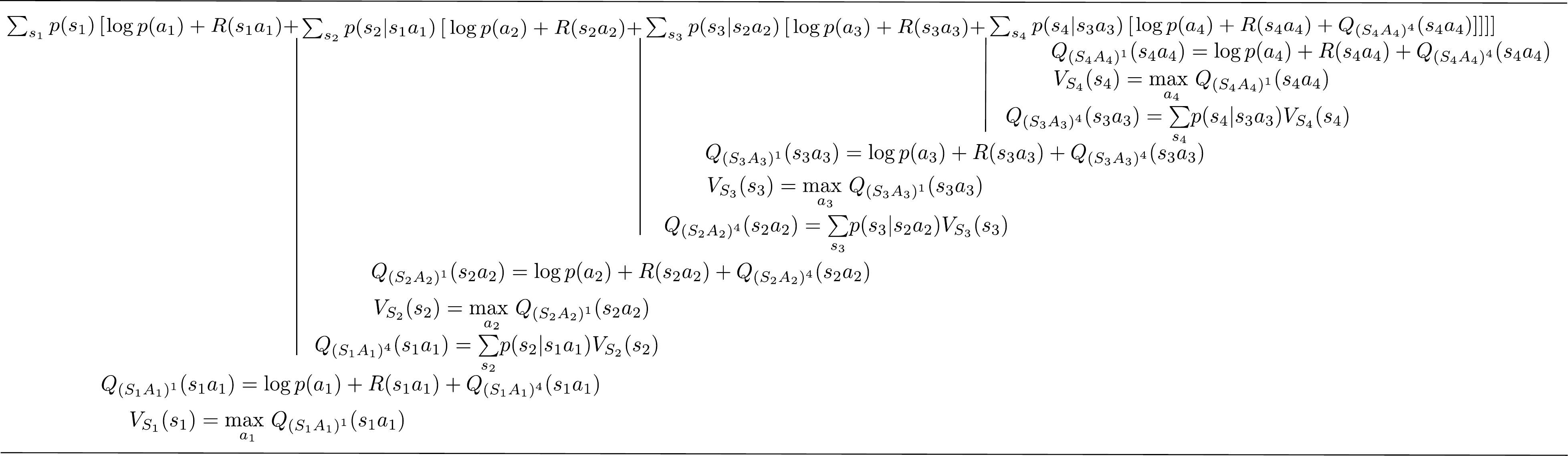} 
  \caption{Backward recursions for the Dynamic programming  algorithm for $T=4$. Note the presence of the backward message $Q_{(S_4A_4)^4}(s_4a_4)$ that may carry information from  time steps beyond $T$, or may represent final constraints.}
  \label{fig:recdp}
\end{figure}

\section{Dynamic programming on the factor graph}
\label{sec:dp}

The standard approach to dynamic programming  is based on the maximization of the expected sum of rewards \citep{Bertsekas2019, Sutton2018}. In previous sections, we have included rewards in factorization (\ref{eq:factmodel}), but we have formulated the optimization problem as the maximization of posterior probabilities, or marginals, which only implicitly involve the rewards. Obviously, one wonders whether the two approaches can be seen under a unified framework - after all Bellman backups resemble backward message combinations.  We  show here that it is possible to map DP directly into the factor graph formulation if we consider rewards and their expectations as contributing to probability messages, but in the log-space. We can see how  in the probability space, the DP messages can travel on the factor graph, just as in the Sum/Max-product algorithm, but with different definitions of the propagation rules through the building blocks.     
  
The dynamic programming algorithm \citep{Bertsekas2019} is derived as the solution to the following problem 
\begin{equation}
  \begin{array}{rL}
    (a_1^* \dots a_T^*) & = \argmax_{a_1...a_T} \E_{\sim p(s_1a_1 ... s_Ta_T)} \left[ \sum_{t=1}^T (R(s_ta_t) + \log p(a_t) ) \right] \\
    & = \argmax_{a_1...a_T} \sum_{s_1,...,s_T} p(s_1a_1 ... s_Ta_T) \left[ \sum_{t=1}^T (R(s_ta_t) + \log p(a_t)) \right] \\
    & = \argmax_{a_1...a_T} \sum_{s_1,...,s_T} p(s_1) \prod_{t=1}^{T-1} p(s_{t+1}|s_ta_t) \left[ \sum_{t=1}^T (R(s_ta_t) + \log p(a_t)) \right],
  \end{array}
\end{equation}
where $p(s_1a_1 ... s_Ta_T)$ does not include the rewards and the priors on $a_t$  appears in the log in the summation. 
This is slightly different than the sum of pure rewards.  The reason for this modification has been  to obtain the same formal recursions that we have derived for the Sum-product and for the Max-product algorithms. In any case, this is not a crucial problem because $\log p(a_t)$ could be incorporated into $R(s_ta_t)$ and $p(a_t)$  can be assumed to be uniform.  

We have reported in Figure \ref{fig:recdp}, the DP recursions in the notations of the factor graph of Figure \ref{fig:FGrn} for $T=4$. Note that, in comparison to analogous backups in the probability space for the Sum-product (or Max-product) algorithm, the rewards appear as additive terms and there is a mix of max and sums. Formally 
\begin{equation}
  \begin{array}{rL}
    Q_{(S_tA_t)^1}(s_ta_t) & = \log p(a_t) +R(s_ta_t) + \sum_{s_{t+1}} p(s_{t+1}|s_ta_t) V_{S_{t+1}}(s_{t+1}), \\
    V_{S_t}(s_t) & = \max_{a_t} Q_{(S_tA_t)^1}(s_ta_t). \\
  \end{array}
\end{equation}
Translating the recursions in the probability space, we have 
\begin{equation}
  \begin{array}{rL}
    b_{(S_tA_t)^1}(s_ta_t) & = p(a_t) c(s_ta_t) e^{\sum_{s_{t+1}} p(s_{t+1}|s_ta_t) \log b_{S_{t+1}}(s_{t+1})}, \\
    b_{S_t}(s_t) & = \max_{a_t} b_{(S_tA_t)^1}(s_ta_t). \\
  \end{array}
\end{equation}
The crucial difference between DP and the Sum-product algorithm is in the fact that averages and maxima are taken in the log-space on the value function. Conversely in the Sum-product, they are taken in the probability space on the backward distributions. Therefore, DP can be formulated in terms of probability messages traveling on the same factor graph of the Sum-product algorithm, but with different combination rules.     
The DP recursions are reported in Tables \ref{tab:prob} and \ref{tab:log} for comparison, and the specific rules for the messages through the blocks are in Tables  \ref{tab:source}, \ref{tab:shadedA}, \ref{tab:shadedS}, \ref{tab:dynamics} and \ref{tab:diverter}.  

The best SASA sequence, written both in the log-space and in the probability space, is immediately derived from the graph: 

At $t=1$ we have 
\begin{equation}
  \begin{array}{rL}
    s_1^* & = \argmax_{s_1} p(s_1) V_{S_1}(s_1) = \argmax_{s_1} p(s_1) \log b_{S_1}(s_1), \\
    a_1^* & = \argmax_{a_1} Q_{(S_1A_1)^1}(s_1^*a_1) = \argmax_{a_1} b_{(S_1A_1)^1}(s_1^*a_1). 
  \end{array}
\end{equation}
At $t=2$
\begin{equation}
  \begin{array}{rL}
    s_2^* & = \argmax_{s_2} p(s_2|s_1^*a_1^*) V_{S_2}(s_2) = \argmax_{s_2} p(s_2|s_1^*a_1^*) \log b_{S_2}(s_2), \\
    a_2^* & = \argmax_{a_2} Q_{(S_2A_2)^1}(s_2^*a_2) = \argmax_{a_2} b_{(S_2A_2)^1}(s_2^*a_2) ,  
  \end{array}
\end{equation}
 ... etc. 

The unique formulation on the factor graph allows the derivation of all other inferences, such as A, S, AS -sequences, also for DP,   simply using the specific propagation rules on the graph. The policy distribution has the same formal expression as in  
(\ref{eq:policy}).

\section{SoftDP}
\label{sec:softdp}

The presence of the max operator in the DP algorithm, suggests that, similarly to the Sum/Max-product approach, we could replace the max operator with a generic soft-max function to provide a different interpolation between a more entropic solution and the optimal DP algorithm. Using a soft-max function, we propose the following {\em SoftDP} updates
\begin{equation}
  \begin{array}{rL}
    Q_{(S_tA_t)^1}(s_ta_t) & = \log p(a_t) + R(s_ta_t) + \sum_{s_{t+1}} p(s_{t+1}|s_ta_t) V_{S_{t+1}}(s_{t+1}), \\
    V_{S_t}(s_t) & = \frac{\sum_{a_t} e^{\beta Q_{(S_tA_t)^1}(s_ta_t)} Q_{(S_tA_t)^1}(s_ta_t)}{\sum_{a_t'} e^{\beta Q_{(S_tA_t)^1}(s_ta_t')} } .
  \end{array}
  \label{eq:recsoftDP}
\end{equation}
The parameter $\beta$ can be used to control the sharpness of the soft-max function. If $\beta$ is a large potisive number, the soft-max tends to the maximum. When $\beta$ is a small positive number, the soft-max function tends to return the mean. The soft-max function used here is popular in the neural networks literature. Details about its behavior are in Appendix \ref{app:smax}.
We have not investigated the existence of a function that these recursions  optimize for a finite value of $\beta$, as in the case of the Sum/Max-product algorithm. We leave it to further analyses. However, we observe that lowering the value of $\beta$ shifts the policy distribution  towards a smoother, i.e., more entropic, configuration. We show this effect in the simulations in a later section. 
 
In the probability space, the  recursion for the backward message $b_{(S_tA_t)^1}(s_ta_t)$ is  the same as in DP, while the update for $b_{S_t}(s_t)$ becomes
\begin{equation}
  b_{S_t}(s_t) \propto \exp \left[ \frac{\sum_{a_t} \log b_{(S_tA_t)^1}(s_ta_t) b_{S_t}(s_t)^\beta}{\sum_{a_t'}  b_{(S_tA_t)^1}(s_ta_t')^\beta} \right].
\end{equation}
All recursions are included in Tables \ref{tab:prob} and \ref{tab:log}. Also the propagation rule through the blocks are in Tables \ref{tab:source}, \ref{tab:shadedA}, \ref{tab:shadedS}, \ref{tab:dynamics} and \ref{tab:diverter}.

\section{Maximum expected reward and entropy}
\label{sec:maxrewent}

In all the previous approaches to optimal control, we have derived the solutions as optimal inferences on the factorized model of Figure \ref{fig:Bayes}, in the probability space for the Sum- and Max-product algorithms, or as the best action sequence in the log-space for DP. The policy distribution is then written {\em as a consequence} of the optimization algorithm on that graph.  

A different formulation can be adopted if we formally add to the Bayesian graph "policy" branches $\pi(a_t|s_t)$ that go from each state $S_t$ to each action $A_t$ and pose the problem  as the functional optimization problem of finding  the best $\pi(a_t|s_t)$, given the evidence $K_{1:T}$.   The question is: how do we formalized the total reward function? 

\cite{Levine2018}, in his review, suggests that ”less confident” behaviors with respect to the standard probabilistic inference (the Sum-product) could be obtained if we modify the 
function to optimize. In fact, he maintains that the recursions for the Sum-product approach derived above, may be too optimistic within
the context of RL. The idea is to add an extra term to the rewards to account also  for policy entropy. Levine shows that the
modification can also be related to structural variational inference \citep{Levine2018}.  Entropy maximization is also a common criterion in practical uses of  RL  \citep{Ziebart2009} and  stochastic control \citep{Ziebart2010}.
\cite{Levine2018} proposes the following formulation
\begin{equation} 
  \begin{array}{L}
    \lbrace \pi^*(a_1|s_1) \dots \pi^*(a_T|s_T) \rbrace = \\
    \qquad\quad \argmax_{\pi(a_1|s_1) \dots \pi(a_T|s_T)} \E_{\sim \hat p(s_1a_1 \dots s_Ta_T)} 
      \left[ \sum_{t=1}^T \left( R(s_ta_t) + \log p(a_t) - \log \pi(a_t|s_t) \right) \right]
  \end{array}
  \label{eq:costLevine}
\end{equation}
where
\begin{equation*}
  \hat p(s_1a_1 \dots s_Ta_T) = p(s_1) \pi(a_T|s_T) \prod_{t=1}^{T-1} p(s_{t+1}|s_ta_t) \pi(a_t|s_t).   
\end{equation*}
Note that here the policy distributions are included in the factorization. The extra term $\log \pi(a_t|s_t)$ will gives rise to entropy maximization This will be better explained in the generalization that follows and in Appendix \ref{app:levgen}.  
The backup recursions for the optimal policy distributions \citep{Levine2018}, in the factor graph notations, are  
\begin{equation}
  \begin{array}{rL}
    Q_{(S_tA_t)^1}(s_ta_t) & = \log p(a_t) +R(s_ta_t) + \sum_{s_{t+1}} p(s_{t+1}|s_ta_t) V_{S_{t+1}}(s_{t+1}), \\
    V_{S_t}(s_t) & = \log \sum_{a_t} e^{Q_{(S_tA_t)^1}(s_ta_t)}. \\
  \end{array}
  \label{eq:reclevine}
\end{equation}
The optimal policy distributions are also shown to have the usual formal expression 
\begin{equation}
  \pi^*(a_t|s_t) \propto e^{\left( Q_{(S_tA_t)^1}(s_ta_t) - V(s_t) \right)}.
\end{equation}    
In our effort to provide more general approaches to the policy search, we have generalized the soft-max function to include an extra parameter $\alpha$, with the recursions 
\begin{equation}
  \begin{array}{rL}
    Q_{(S_tA_t)^1}(s_ta_t) & =  \log p(a_t) + R(s_ta_t) + \sum_{s_{t+1}} p(s_{t+1}|s_ta_t) V_{S_{t+1}}(s_{t+1}), \\
    V_{S_t}(s_t) & = \frac{1}{\alpha} \log \sum_{a_t} e^{\alpha Q_{(S_tA_t)^1}(s_ta_t)}. \\
  \end{array}
  \label{eq:reclevinegen}
\end{equation}
The function used in the value function update is the same one used for the Sum/Max product algorithm and is such that for $\alpha \to \infty$ gives the maximum and therefore the DP solution. 

We have worked the recursion backward and we show (see Appendix \ref{app:levgen} for the proof) that the above recursions solve the following optimization problem
\begin{equation} 
  \begin{array}{L}
    \lbrace \pi^*(a_1|s_1) \dots \pi^*(a_T|s_T) \rbrace = \\
    \qquad\quad \argmax_{\pi(a_1|s_1) \dots \pi(a_T|s_T)} \E_{\sim \hat p(s_1a_1 \dots s_Ta_T)}
      \left[ \sum_{t=1}^T \left( R(s_ta_t) + \log p(a_t) - \frac{1}{\alpha} \log \pi_\alpha(a_t|s_t) \right) \right] 
  \end{array}
  \label{eq:costgen}
\end{equation}
where
\begin{equation}
  \begin{array}{rLl}
    \hat p(s_1a_1 \dots s_Ta_T) & = p(s_1) \pi_\alpha(a_T|s_T) \prod_{t=1}^{T-1} p(s_{t+1}|s_ta_t) \pi_\alpha(a_t|s_t), & \\
     \pi_\alpha(a_t|s_t) & = \frac{\pi(a_t|s_t)^\alpha}{\sum_{a_t'} \pi(a_t'|s_t)^\alpha}, & t=1:T.
  \end{array} 
\end{equation}
The above expression generalizes (\ref{eq:costLevine}) by including a parameter $\alpha>0$. We show in Appendix \ref{app:levgen} that, when $\alpha$ is  large, the extra term becomes progressively irrelevant, and the distributions $\pi_\alpha(a_t|s_t)$ become more concentrated on the max value of the $Q$-function resulting in a hard DP solution. Furthermore,  when $\alpha <1 $, more weight is given to the extra term, the distributions  $\pi_\alpha(a_t|s_t)$  becomes smoother and we have more entropic policy distributions. We demonstrate this effect in some of the simulations that follow.  

In Appendix \ref{app:levgen}, we discuss explicitly this formulation for $T=4$ deriving the recursions (\ref{eq:reclevinegen}) showing how the extra term gives rise to (iterative) simultaneous reward and entropy  maximization. It is pointed out in our discussion that the criterion does not simply add an entropy term to the rewards because the policy distribution affects also the reward as it appears in the factorization used in the expectation.

The recursions (\ref{eq:reclevinegen}) can be translated in the probability space as
\begin{equation}
  \begin{array}{rL} 
    b_{(S_tA_t)^1} (s_ta_t) & = p(a_t) c(s_ta_t) e^{\sum_{s_{t+1}} p(s_{t+1}|s_ta_t) \log b_{S_{t+1}}(s_{t+1})} \\ 
    b_{S_t}(s_t) & = \left[ \sum_{a_t} b_{(S_tA_t)^1} (s_ta_t)^\alpha \right]^\frac{1}{\alpha} 
  \end{array}
\end{equation}
The recursions, both in the probability and the log space, are included in Tables \ref{tab:prob} and \ref{tab:log}. Also the propagation rule through the blocks are reported in Tables \ref{tab:source}, \ref{tab:shadedA}, \ref{tab:shadedS}, \ref{tab:dynamics} and \ref{tab:diverter}.

\section{Deterministic systems}
\label{sec:deterministic}

The approach to optimal control in this paper is based on the assumption that the system description is stochastic. This is quite useful in the applications when we do not have exact knowledge of the system dynamics and the the probability distribution $p(s_{t+1}|s_ta_t)$ can be  our best estimate. There are cases however, in which the system response is deterministic, i.e., given $s_ta_t$, we have exact knowledge of $s_{t+1}$ through a deterministic function $s_{t+1}=g(s_ta_t)$. In the stochastic framework, this translates into  a transition probability function that is delta function 
\begin{equation} 
  p(s_{t+1}|s_ta_t)=\delta \left( s_{t+1} - g(s_ta_t) \right). 
\end{equation}
Also, if no prior on the actions is available, $p(a_t)=U(a_t)$. The updates in these cases do not change, but  some of them in the various methods may coincide, because the summations (expectations) in the updates disappear and the prior on $A_t$ is irrelevant. More specifically, by looking at Tables \ref{tab:prob} and \ref{tab:log}, the updates for the $Q$-functions, and their probability-space counterparts, have the same (Bellman's) recursions
\begin{equation}
  \begin{array}{rL}
    Q_{(S_tA_t)^1}(s_ta_t) & = R(s_ta_t)+V_{S_{t+1}}(g(s_ta_t)), \\
    b_{(S_tA_t)^1}(s_ta_t) & = c(s_ta_t)b_{S_{t+1}}(g(s_ta_t)).
  \end{array}
\end{equation}
However, there  are differences in the $V$-function updates. For the Sum-product and the Max-Rew/Ent ($\alpha=1$) we have 
\begin{equation}
  \begin{array}{rL}
    V_{S_t}(s_t) & = \log \underset{a_t}{\sum}  e^{Q_{(S_tA_t)^1}(s_ta_t)}, \\
    b_{S_t}(s_t) & = \sum_{a_t} b_{(S_tA_t)^1}(s_ta_t). \\
  \end{array}
\end{equation}
For the Max-product and DP, we have  
\begin{equation}
  \begin{array}{rL}
    V_{S_t}(s_t) & = \max_{a_t} Q_{(S_tA_t)^1}(s_ta_t), \\
    b_{S_t}(s_t) & = \max_{a_t} b_{(S_tA_t)^1}(s_ta_t). \\
  \end{array}
\end{equation}
For the others we have the parametrized soft-max function with various values of $\beta$ and $\alpha$. 

By direct comparison, we can conclude that, when the system is deterministic: {\em DP} and {\em Max-product} coincide; {\em Mean-product} and {\em Max-Rew/Ent ($\alpha=1$)} coincide (also recognized in \cite{Levine2018}). The remaining cases are interpolations of the others.  We have verified in our limited simulations that this is indeed the case and that the solutions in the various groups, even in this deterministic case, are different. 

\section{Infinite horizon case and the steady-state}
\label{sec:infhstate}

We have presented the model in Figure \ref{fig:Bayes} and the various algorithm with reference to a finite horizon scenario.  However, all the analyses easily extends to an infinite-horizon framework simply by adding a discount factor $0< \gamma \le 1$ to the optimized functions and then to the updates. For example, the standard DP updates, in both spaces become
\begin{equation}
  \begin{array}{rL}
    Q_{(S_tA_t)^1}(s_ta_t) & = \log p(a_t) + R(s_ta_t) + \gamma \sum_{s_{t+1}} p(s_{t+1}|s_ta_t) V_{S_{t+1}}(s_{t+1}), \\
    b_{(S_tA_t)^1}(s_ta_t) & = p(a_t) c(s_ta_t) e^{\gamma \sum_{s_{t+1}} p(s_{t+1}|s_ta_t) \log b_{S_{t+1}}(s_{t+1})}.
  \end{array}
\end{equation}  
Also, for the Sum-product, we have immediately 
\begin{equation}
  Q_{(S_tA_t)^1}(s_ta_t) = \log p(a_t) + R(s_ta_t) + \gamma \log \sum_{s_{t+1}} e^{\log p(s_{t+1}|s_ta_t) + V_{S_{t+1}}(s_{t+1})}
\end{equation}
In general, also if $\gamma=1$,  the backward recursions can be run to verify that a steady-state configuration for the $Q$, the $V$-function and the policy $\pi^*$ can be found. This is clearly the solution to a  Bellmann equation applied to the different updates. The analysis of the mathematical conditions for convergence are beyond the scope of this paper. However, generally speaking, if all the states are reachable, a stable configuration should exist. We have verified experimentally that all the methods do converge (also for $\gamma=1$), but they exhibit  marked differences in  the number of iterations necessary to reach the steady state equilibrium. The Max-product algorithm  shows the fastest convergence with the Sum-product following in the list. DP and the other methods seem to show a much slower convergence speed.  We show this effect in the simulations that follow. 

\begin{figure}[ht]
  \centering
  \subfigure{\includegraphics[width=0.45\linewidth]{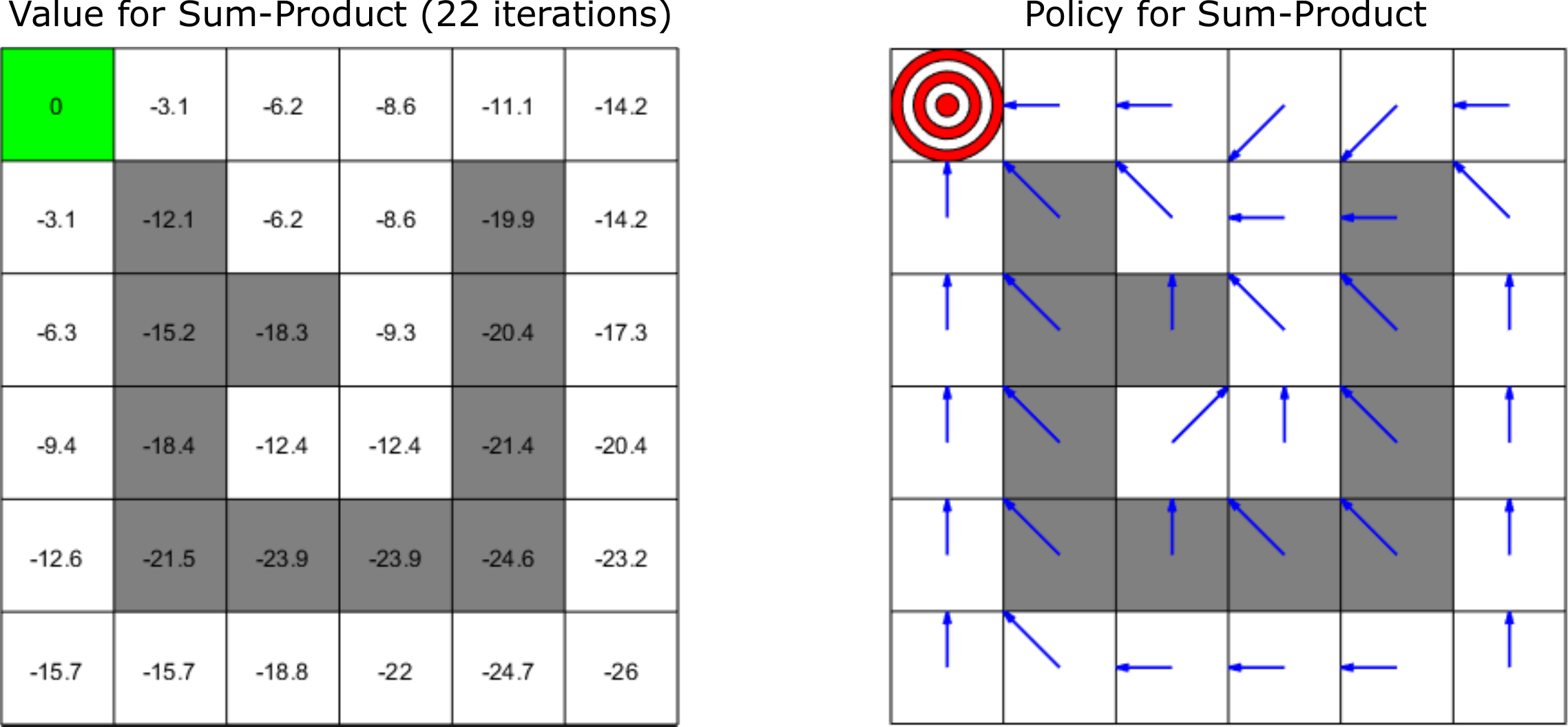}} \hfill
  \subfigure{\includegraphics[width=0.45\linewidth]{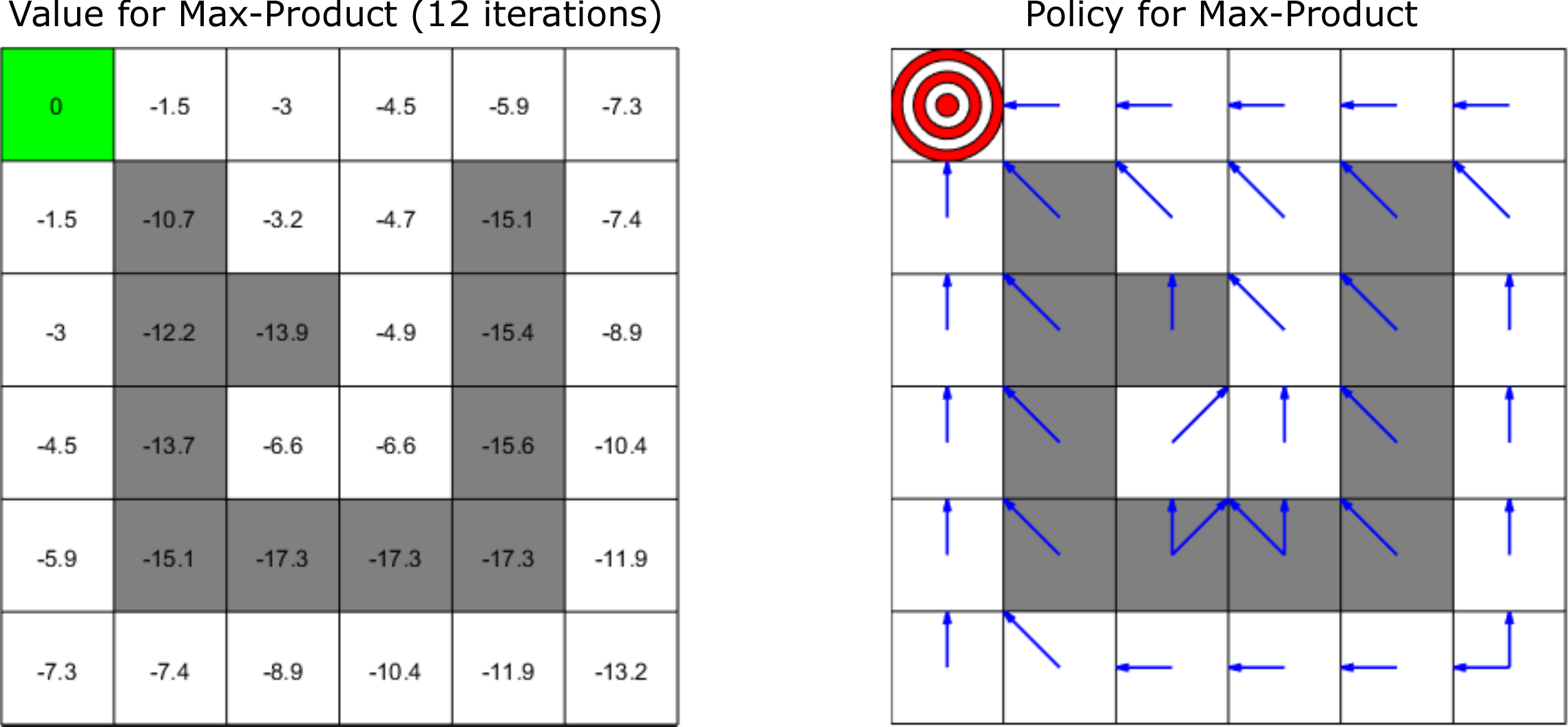}} \hfill
  \subfigure{\includegraphics[width=0.45\linewidth]{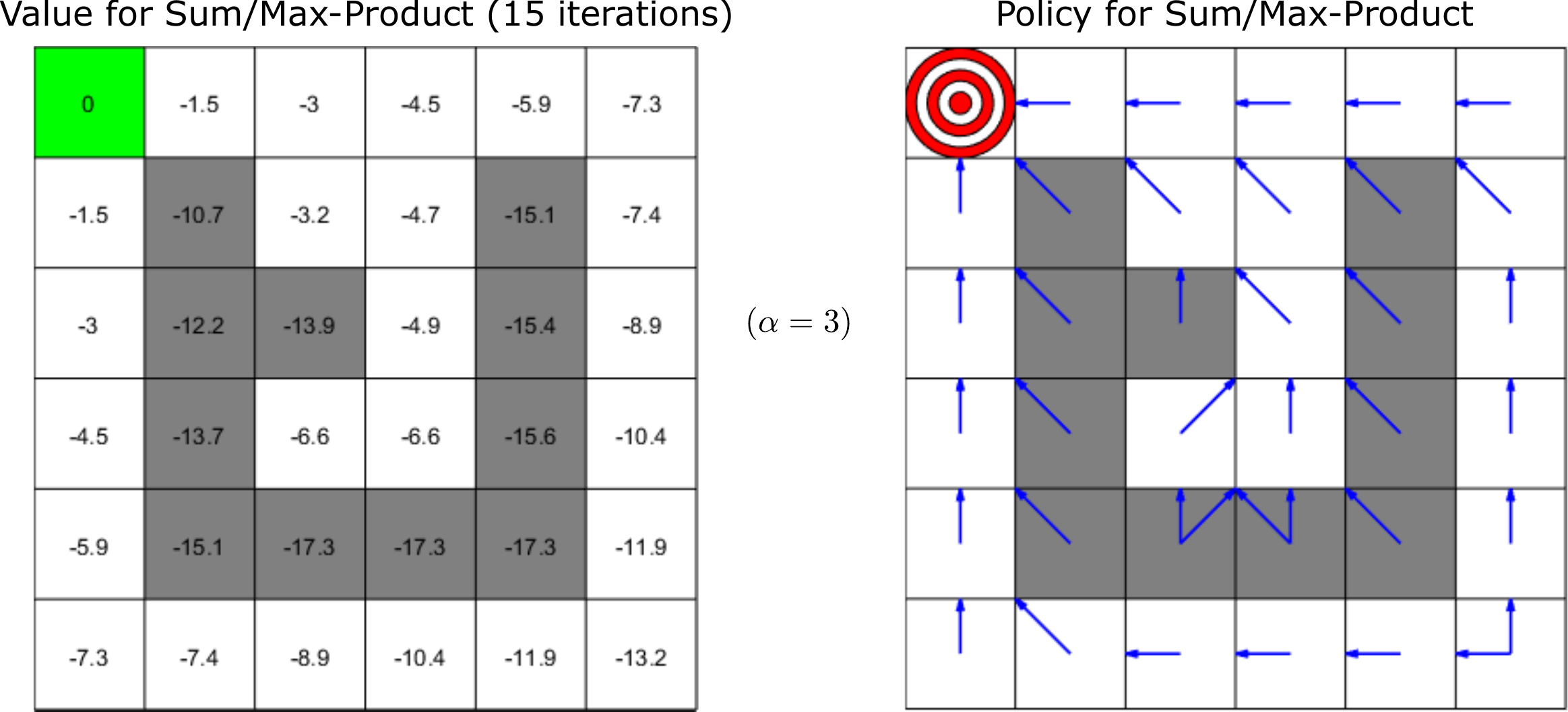}} \hfill
  \subfigure{\includegraphics[width=0.45\linewidth]{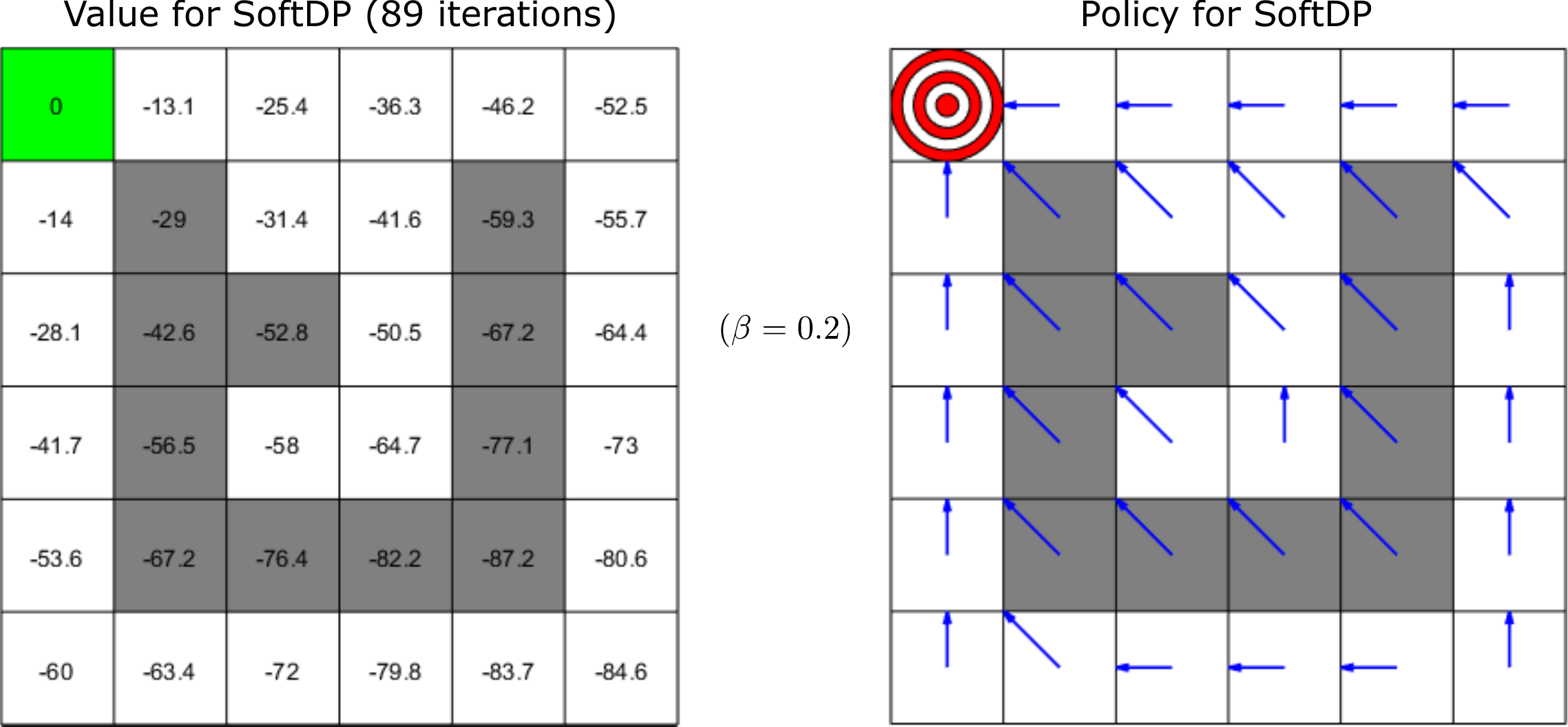}} \hfill
  \subfigure{\includegraphics[width=0.45\linewidth]{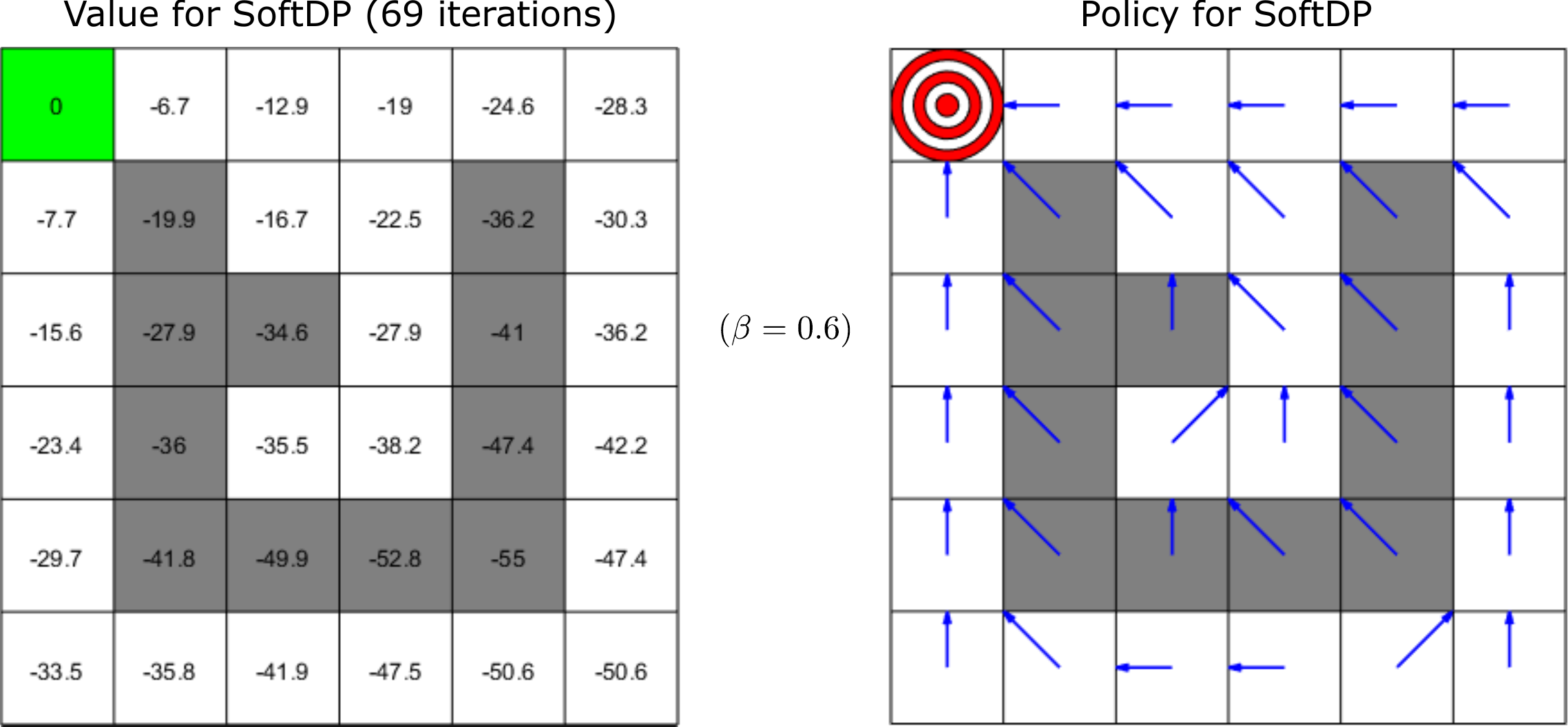}} \hfill
  \subfigure{\includegraphics[width=0.45\linewidth]{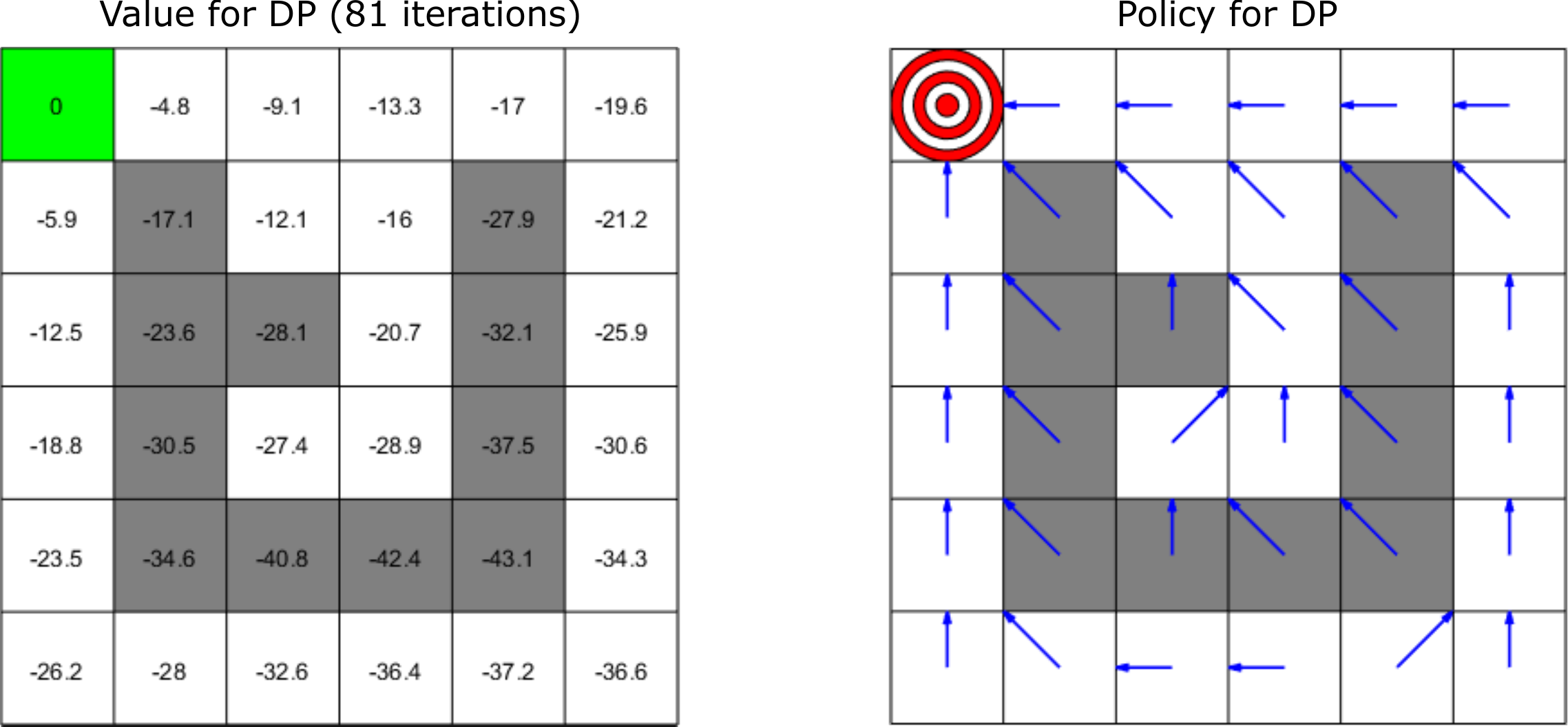}} \hfill
  \subfigure{\includegraphics[width=0.45\linewidth]{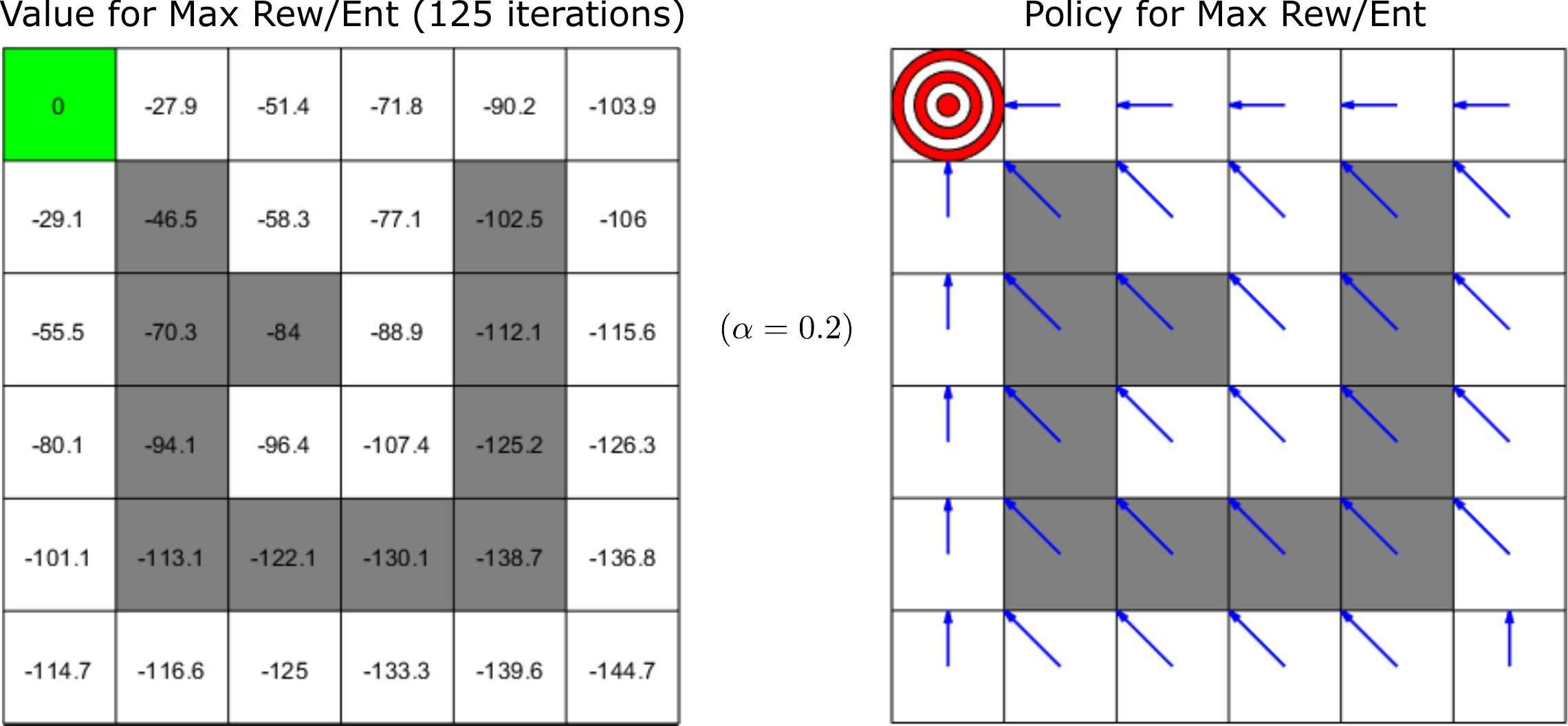}} \hfill
  \subfigure{\includegraphics[width=0.45\linewidth]{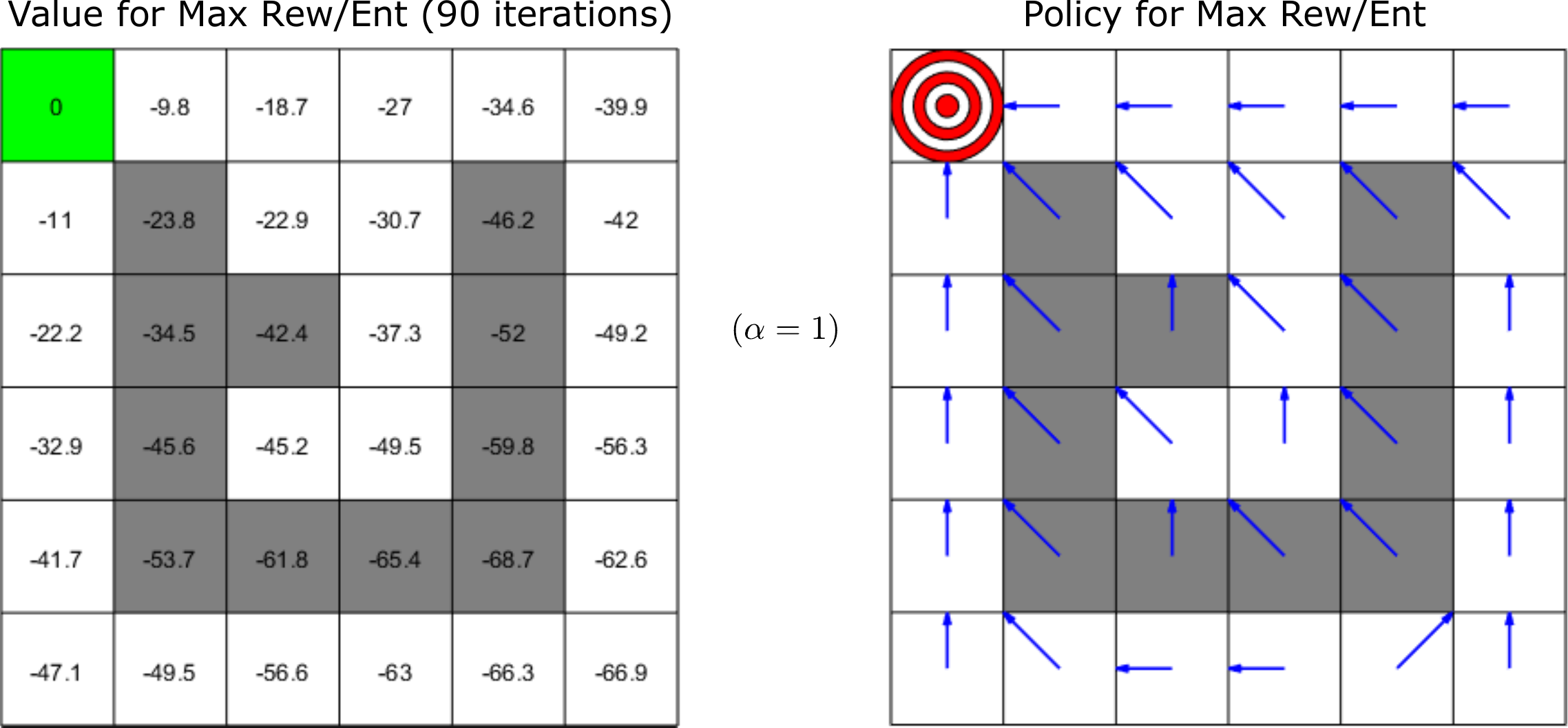}} \hfill
  \subfigure{\includegraphics[width=0.45\linewidth]{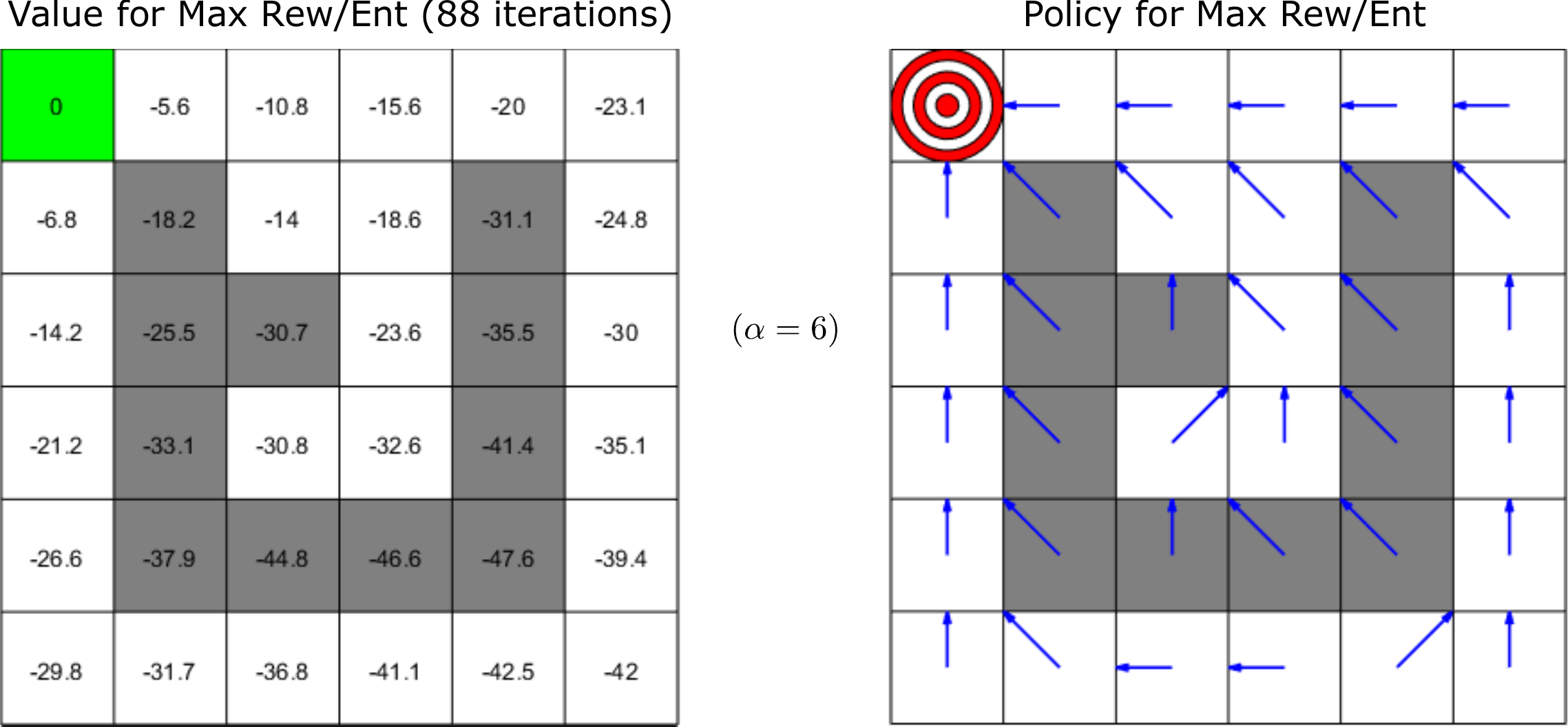}} \hfill
  \subfigure{\includegraphics[width=0.45\linewidth]{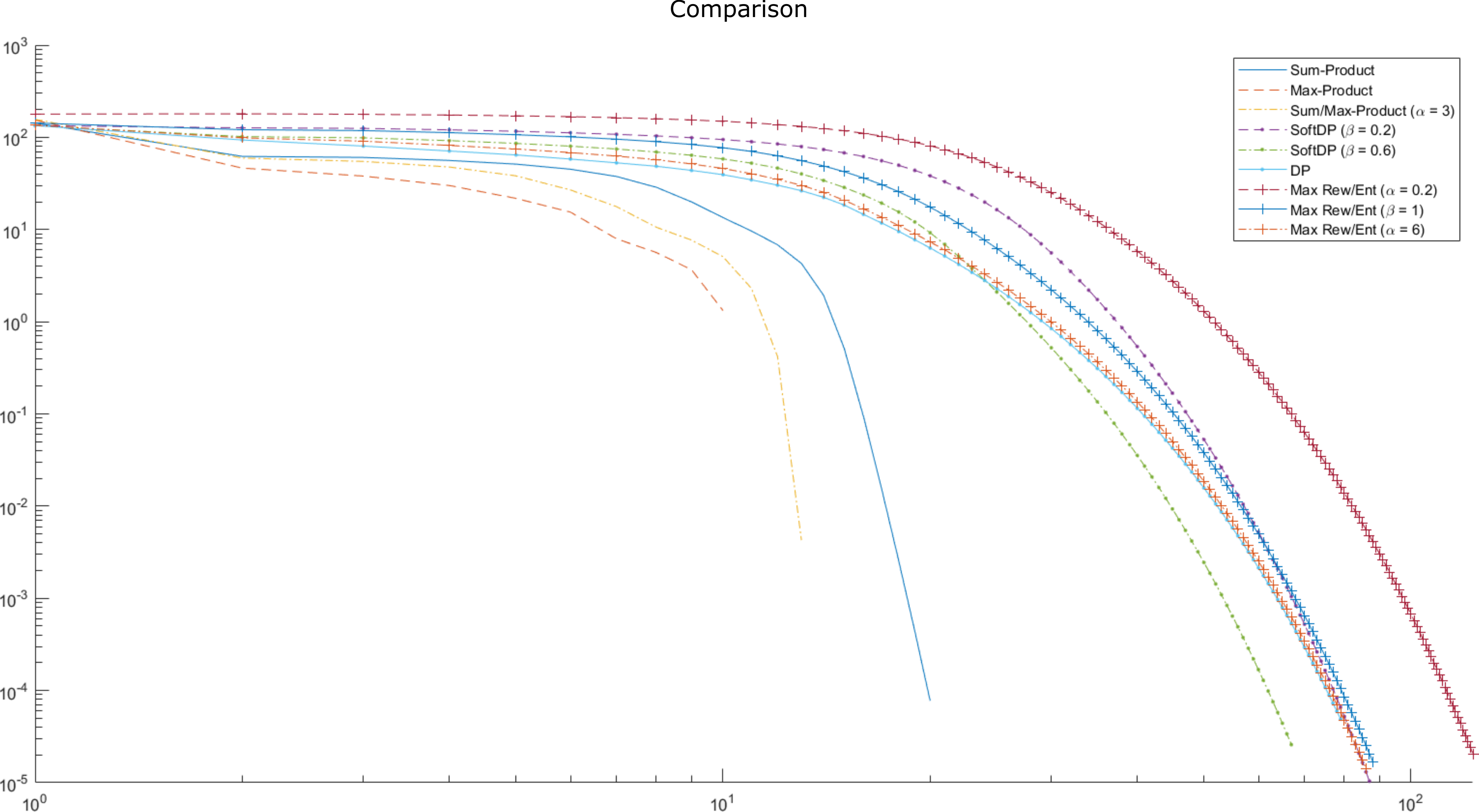}} 
  \caption{Visualization of the max policy direction for the various algorithms. At the top of each figure are reported also the number of iterations necessary to reach a steady-state value function. Reported on the left columns are also the numerical final values for the value function. The lowest right plot shows the value function increments as the iterations progress towards steady-state.}
 \label{fig:Maze}
\end{figure}

\begin{figure}[ht]
  \centering
  \subfigure{\includegraphics[width=0.28\linewidth]{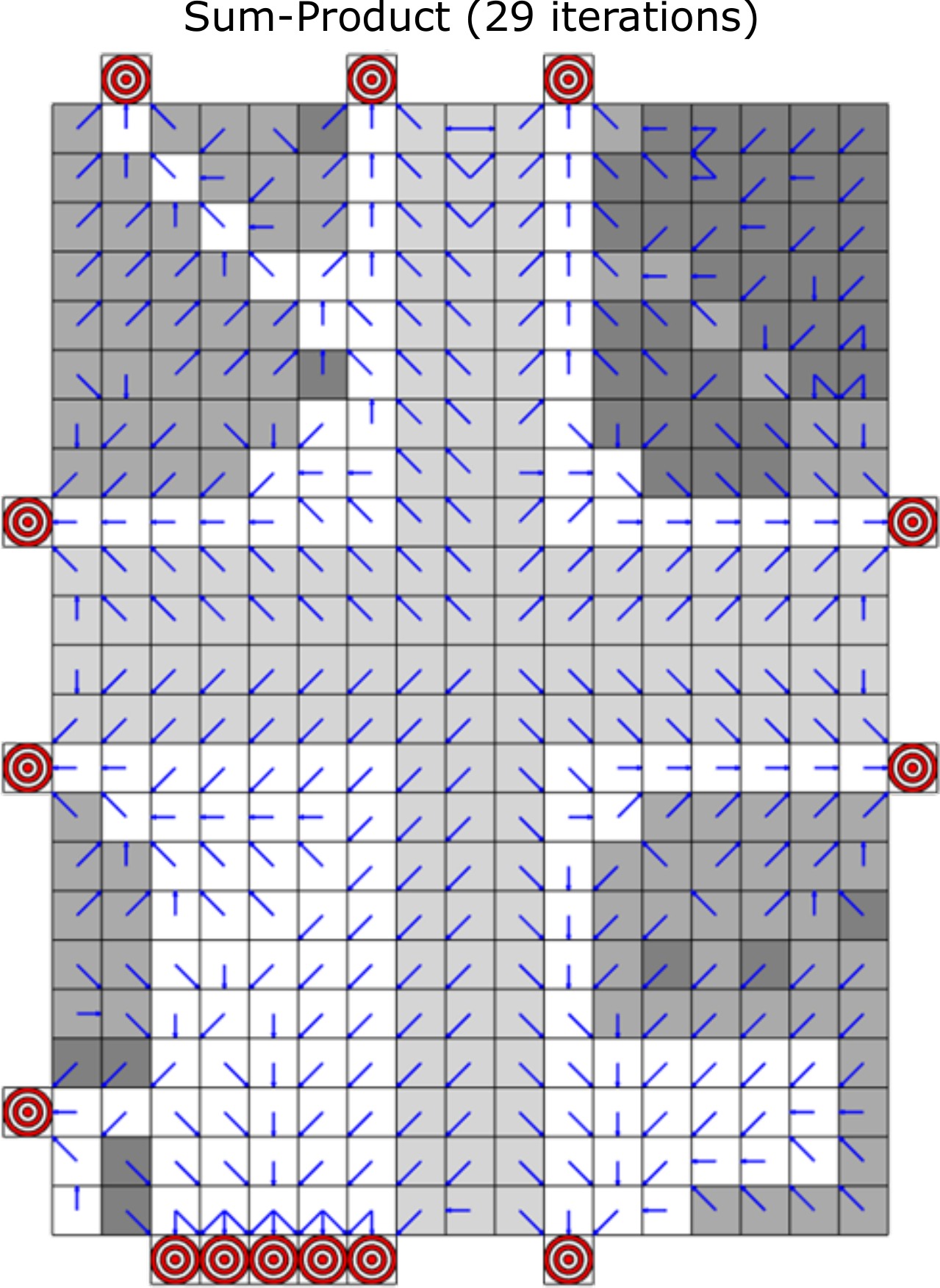}}
  \subfigure{\includegraphics[width=0.28\linewidth]{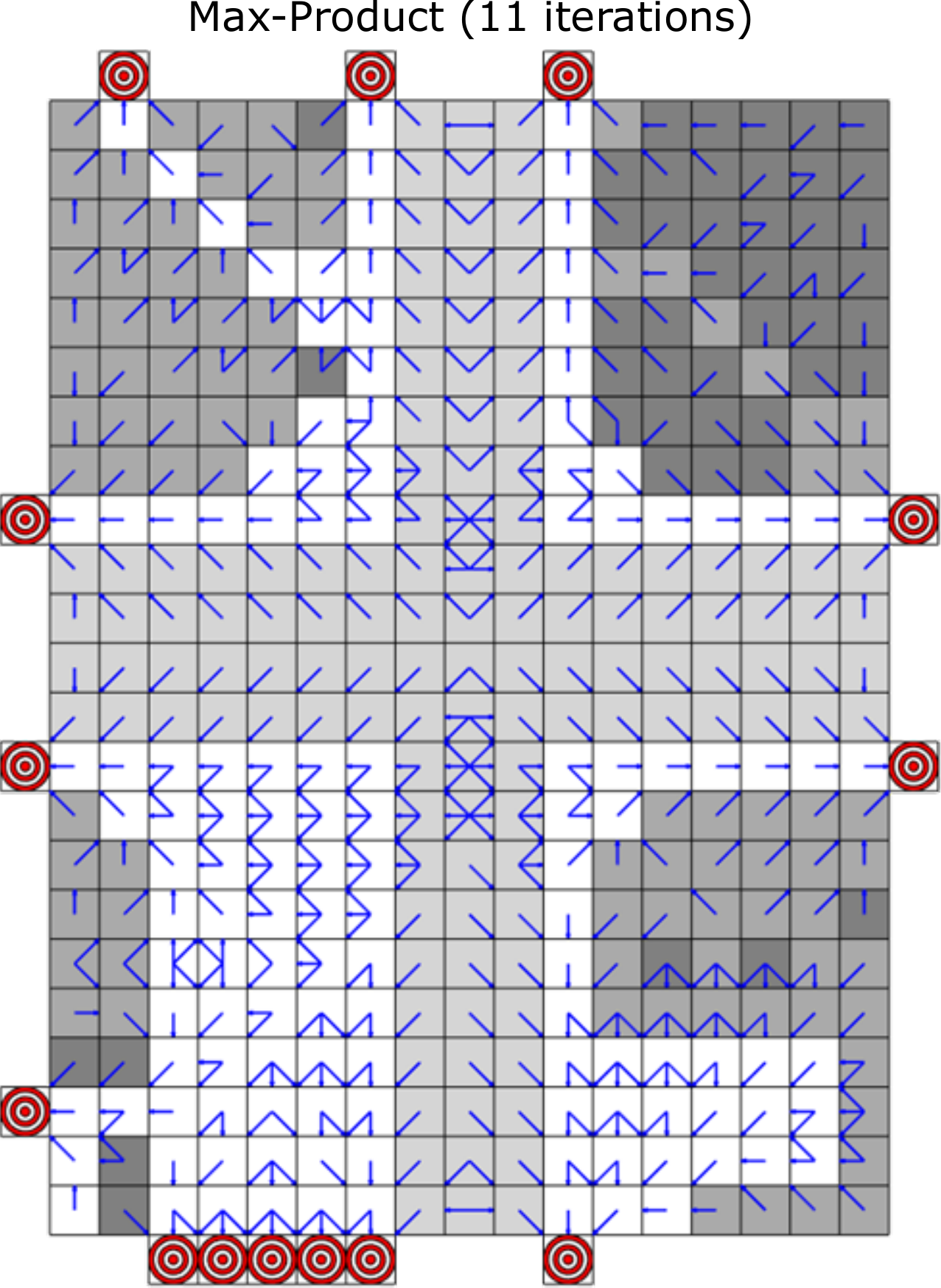}}
  \subfigure{\includegraphics[width=0.28\linewidth]{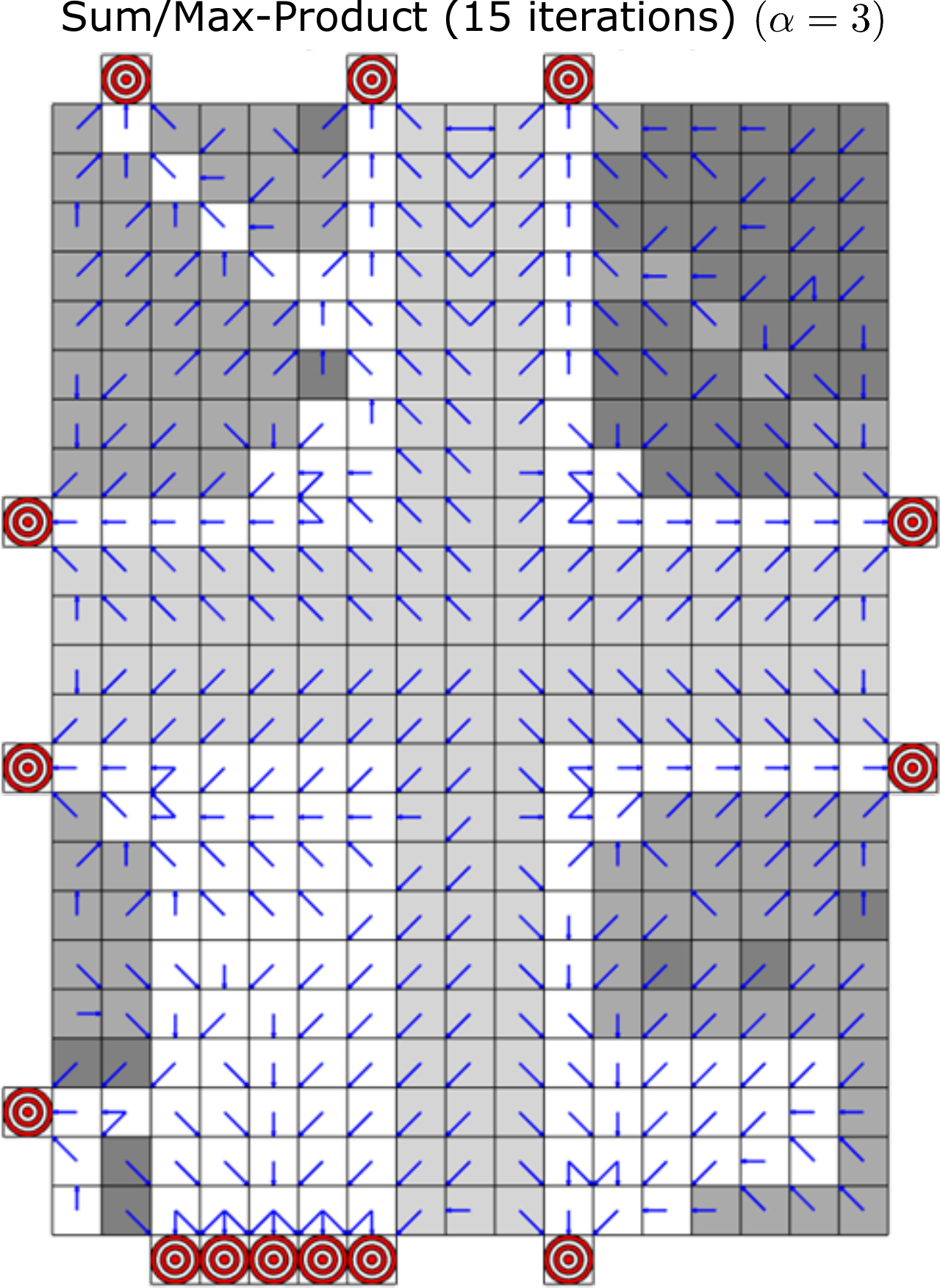}}
  \subfigure{\includegraphics[width=0.28\linewidth]{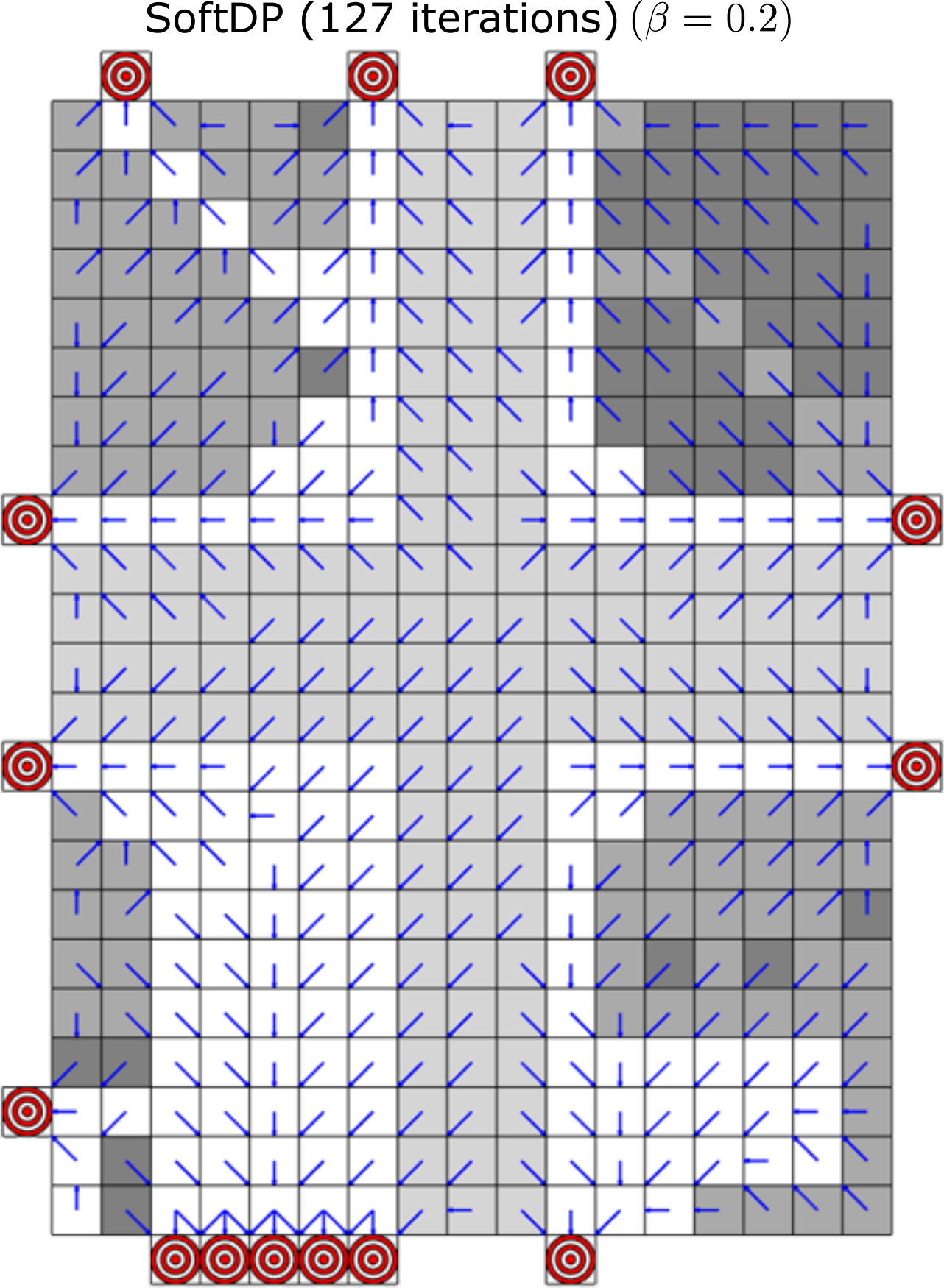}}
  \subfigure{\includegraphics[width=0.28\linewidth]{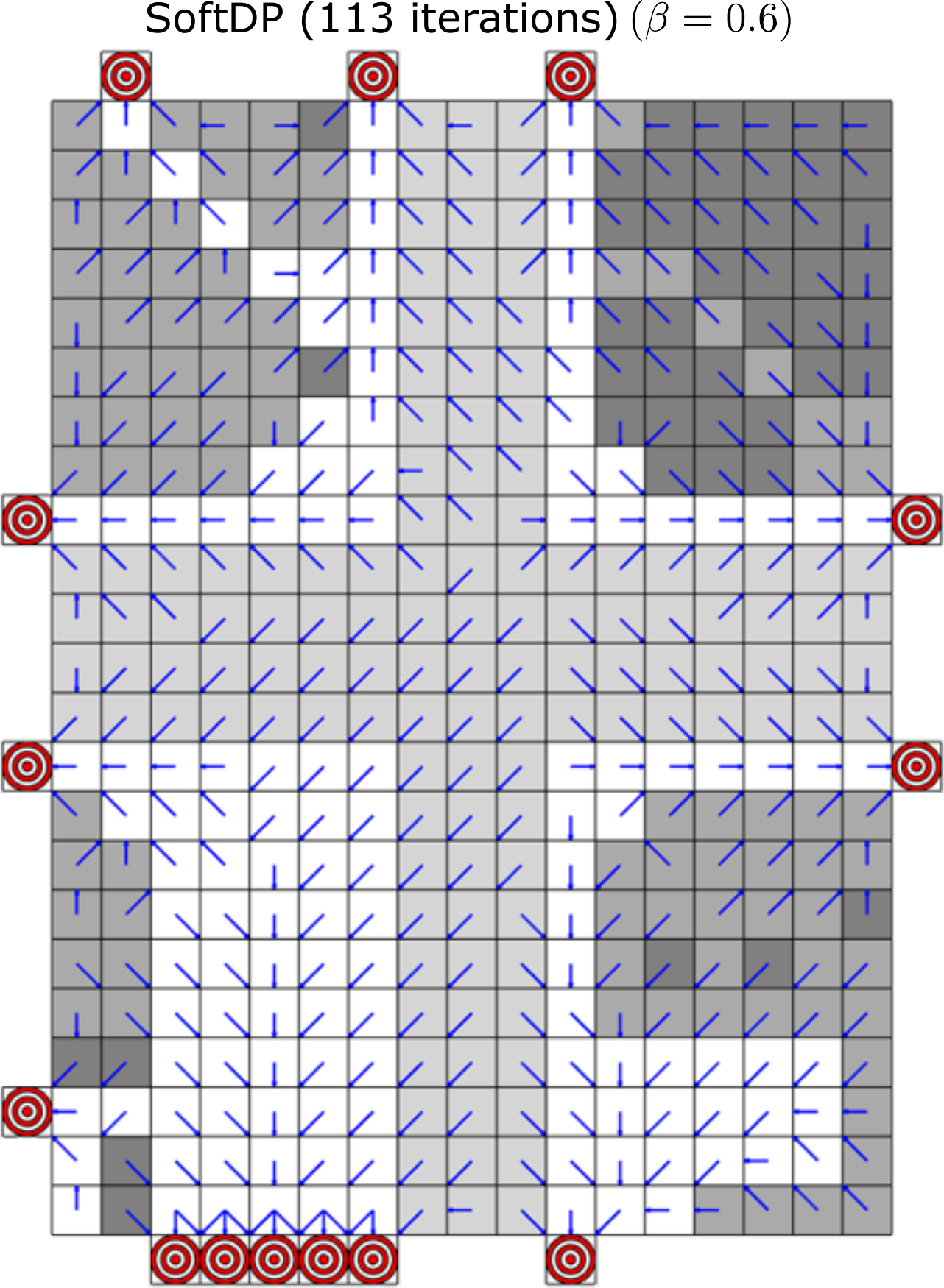}}
  \subfigure{\includegraphics[width=0.28\linewidth]{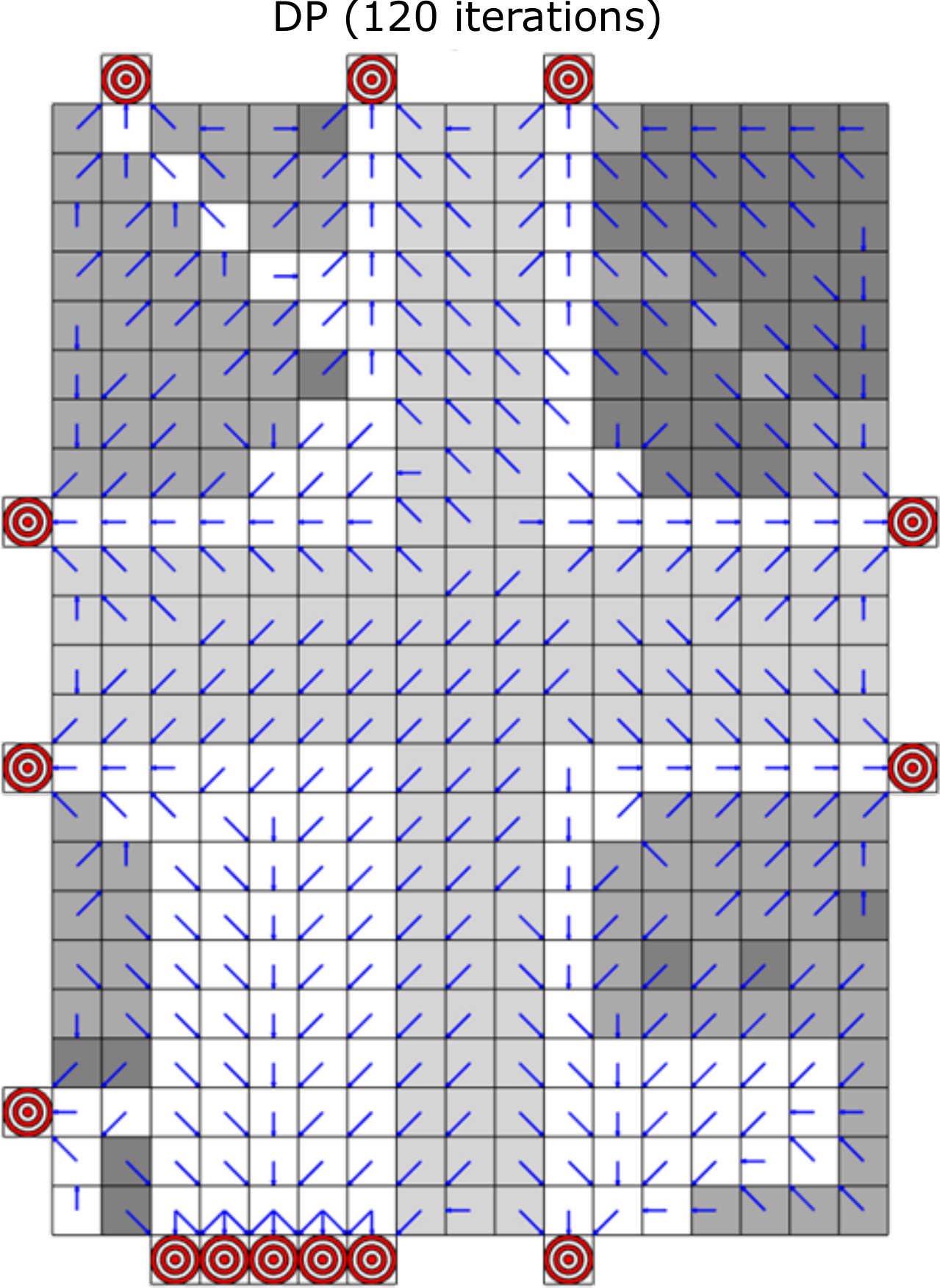}}
  \subfigure{\includegraphics[width=0.28\linewidth]{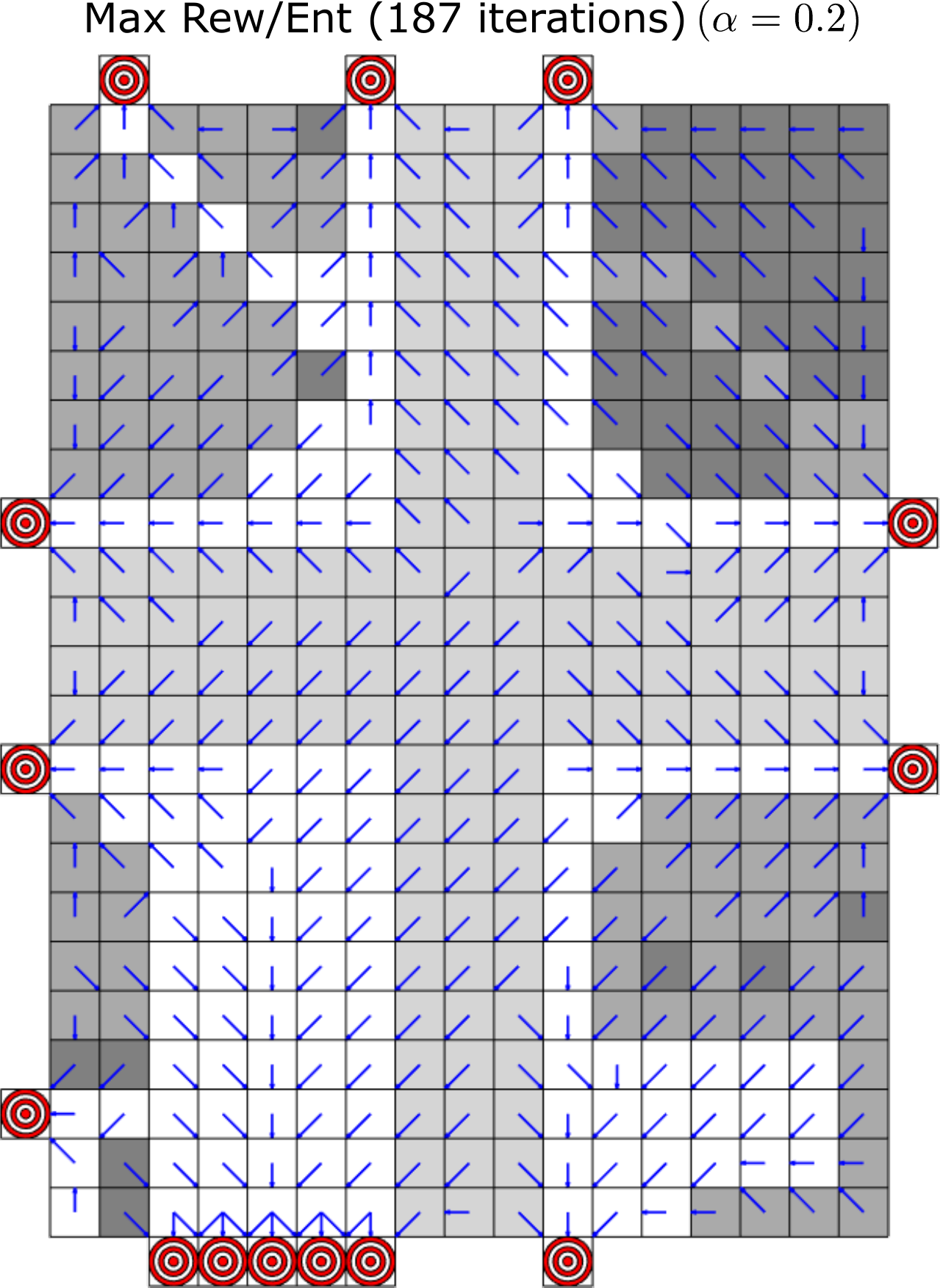}}
  \subfigure{\includegraphics[width=0.28\linewidth]{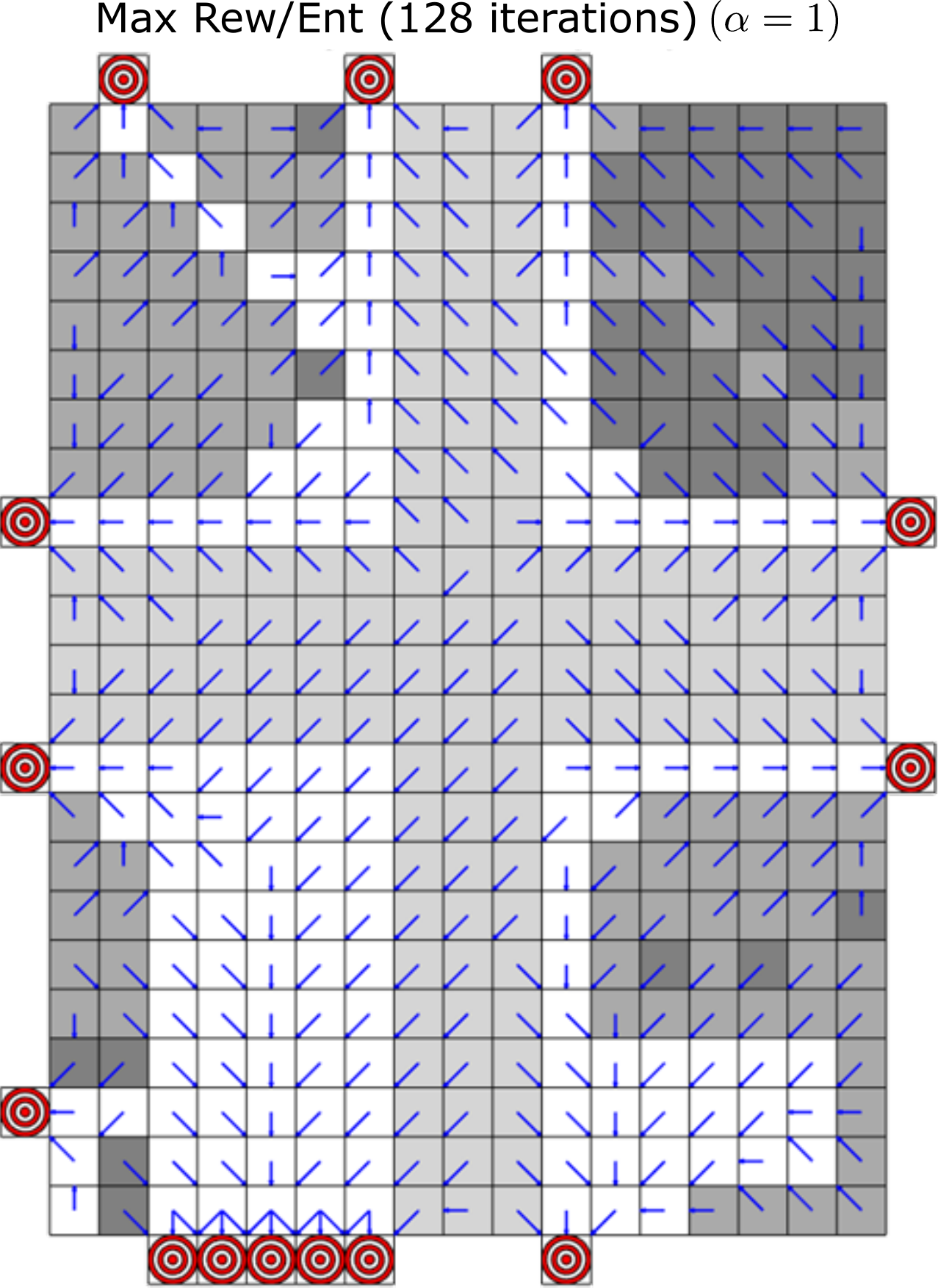}}
  \subfigure{\includegraphics[width=0.28\linewidth]{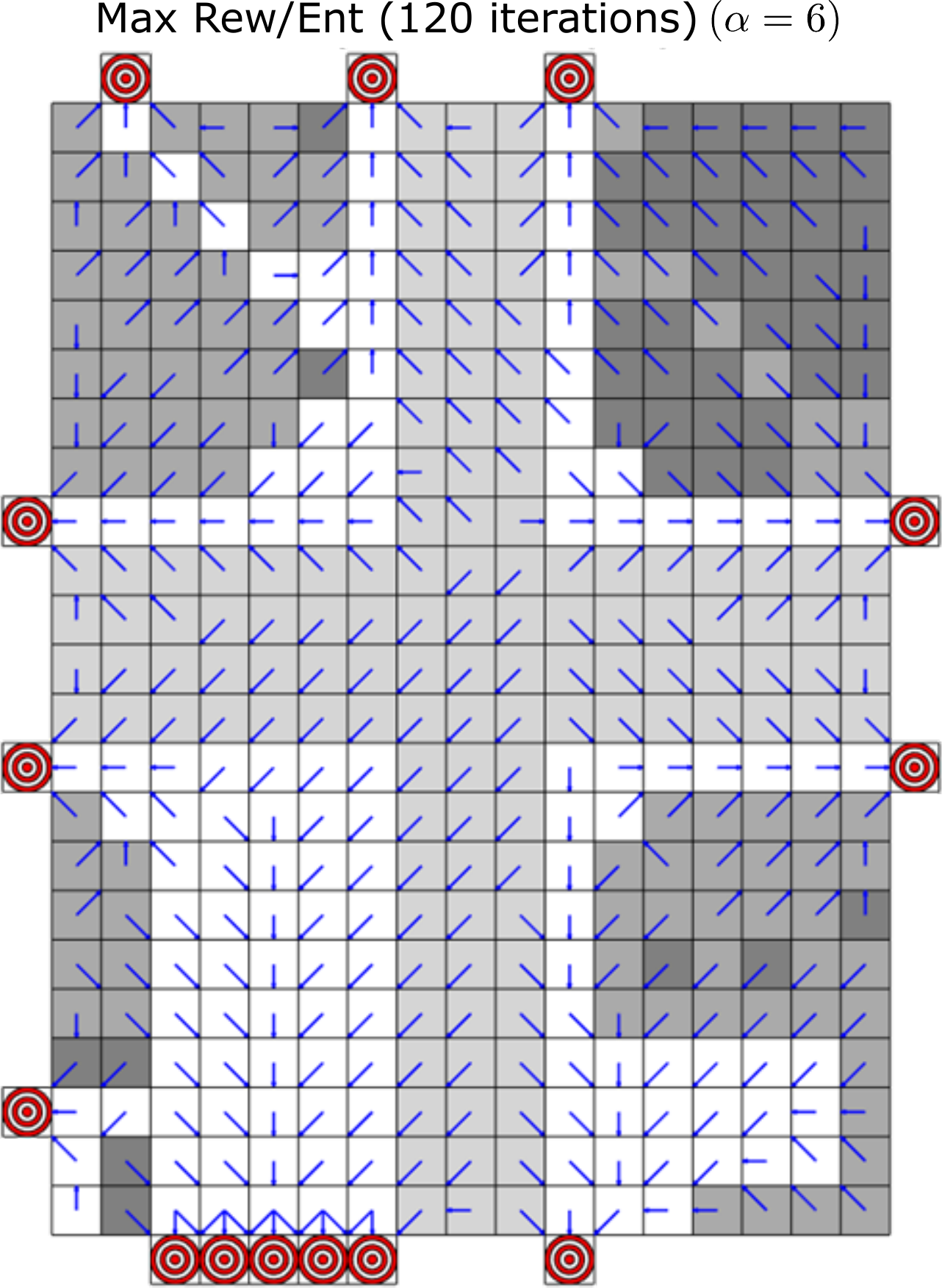}}
  \caption{Visualization of the max policy direction for the various algorithms. At the top each figure are reported also the number of iterations necessary for the value function to reach its steady-state configuration.}
 \label{fig:Policy}
\end{figure}

\begin{figure}[ht]
  \centering
  \subfigure{\includegraphics[width=0.28\linewidth]{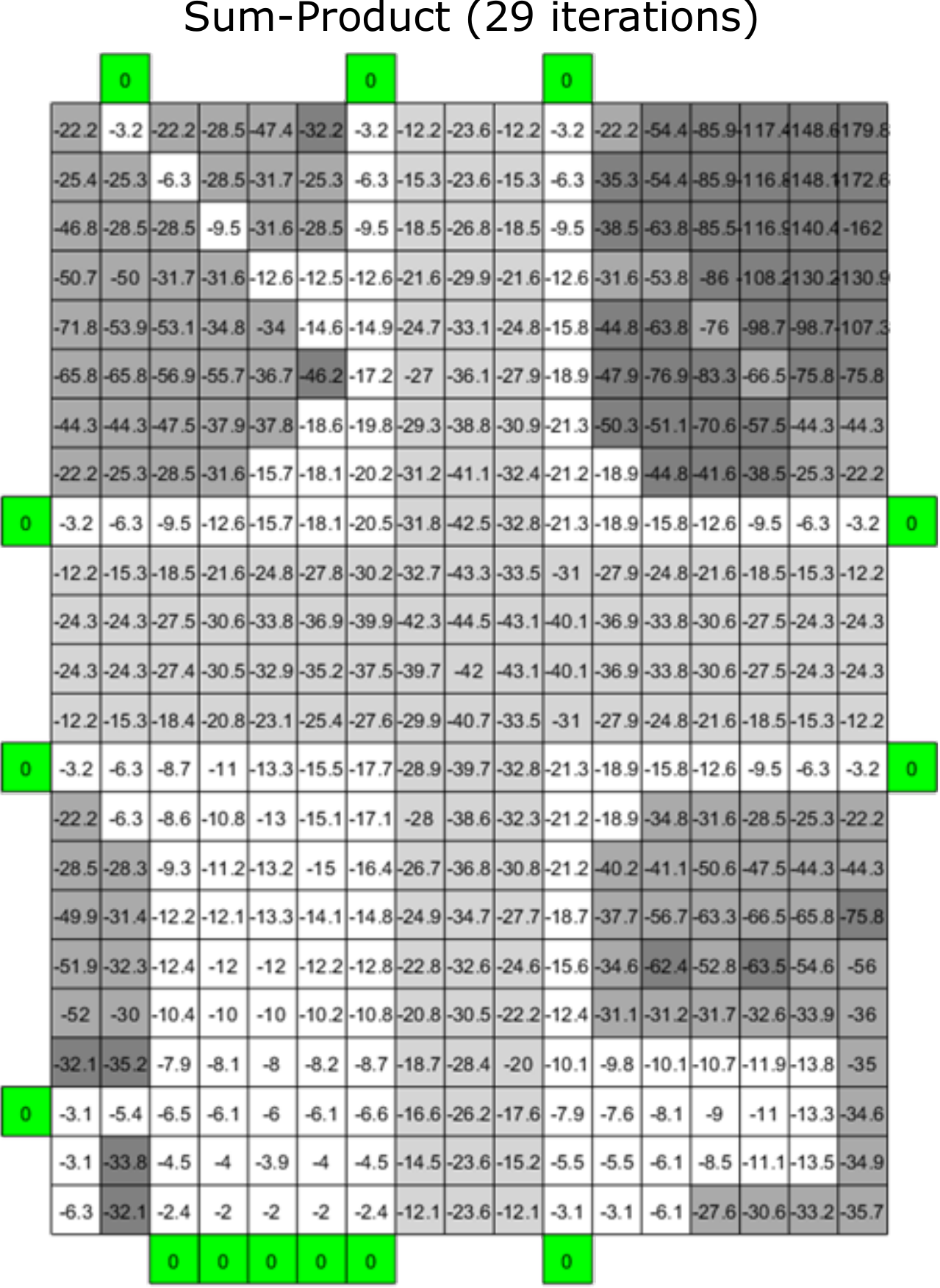}}
  \subfigure{\includegraphics[width=0.28\linewidth]{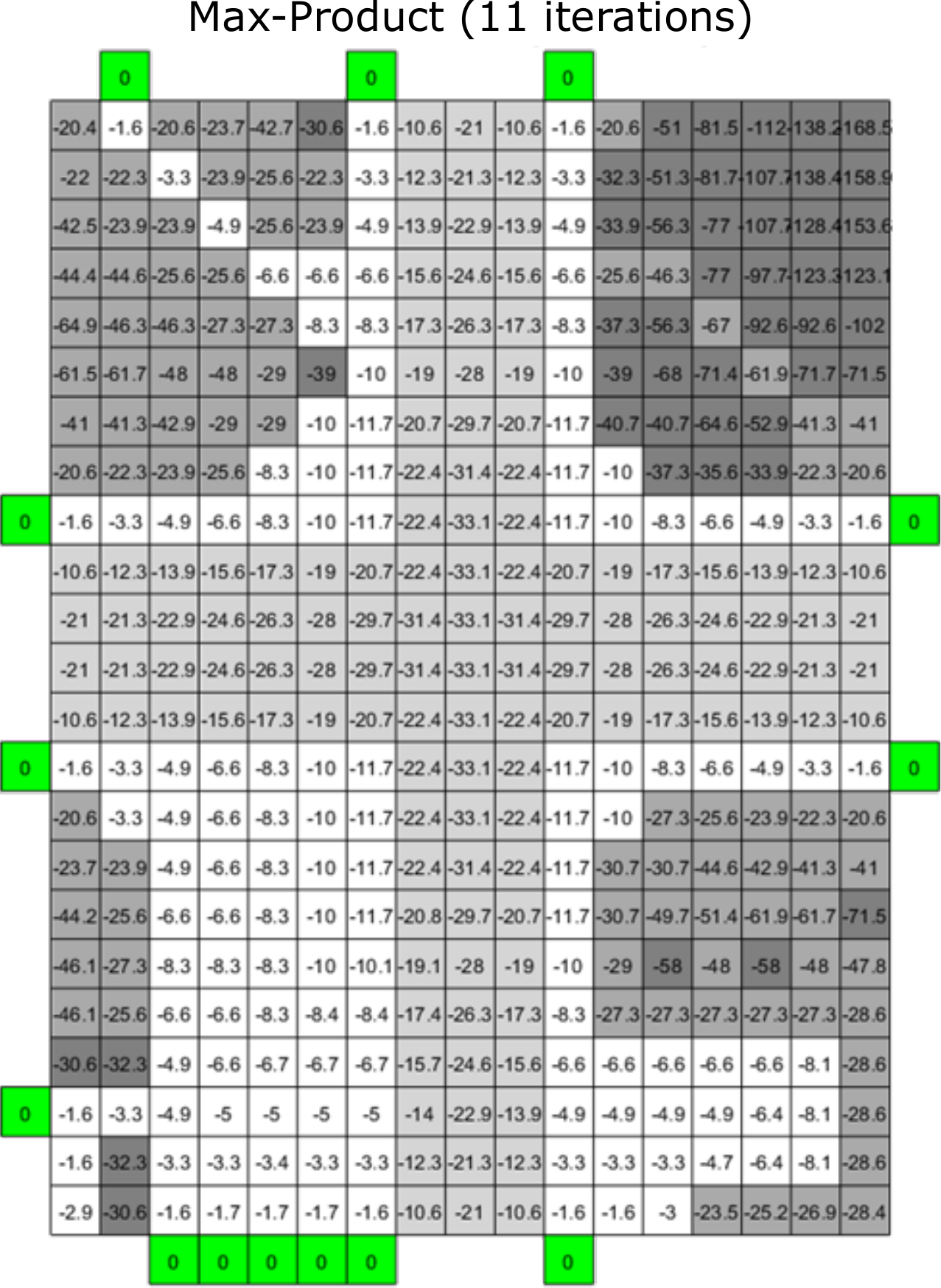}}
  \subfigure{\includegraphics[width=0.28\linewidth]{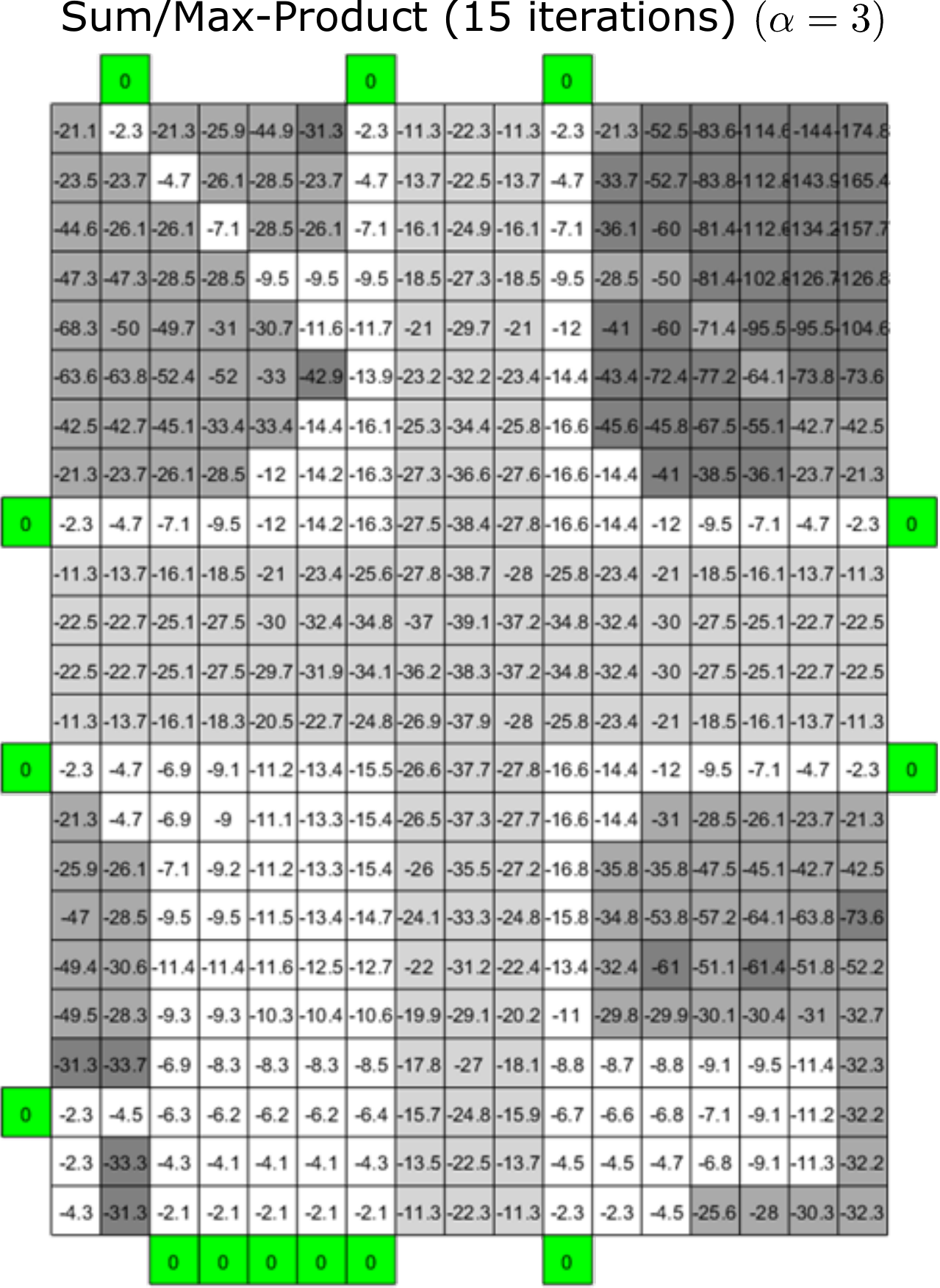}}
  \subfigure{\includegraphics[width=0.28\linewidth]{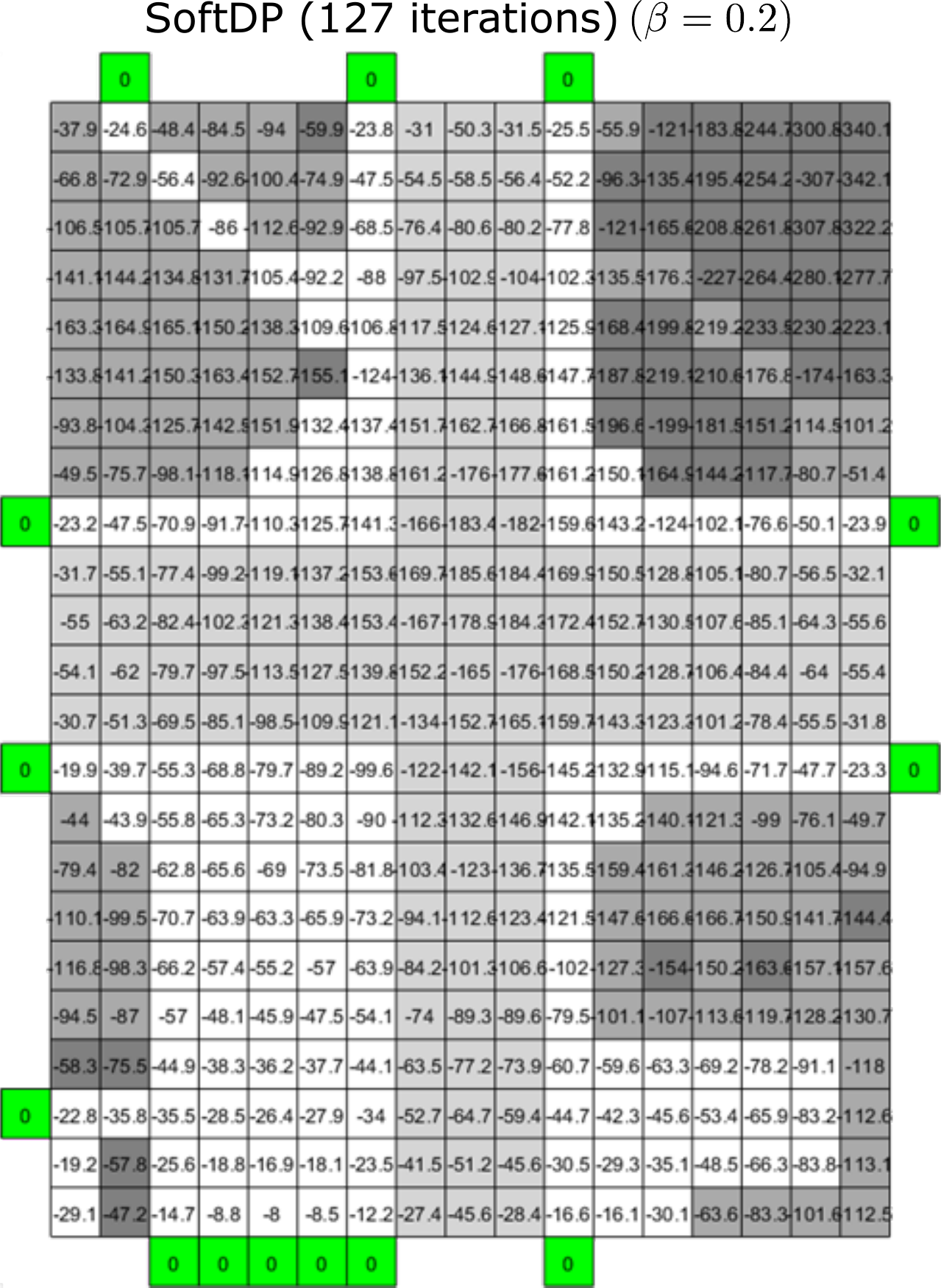}}
  \subfigure{\includegraphics[width=0.28\linewidth]{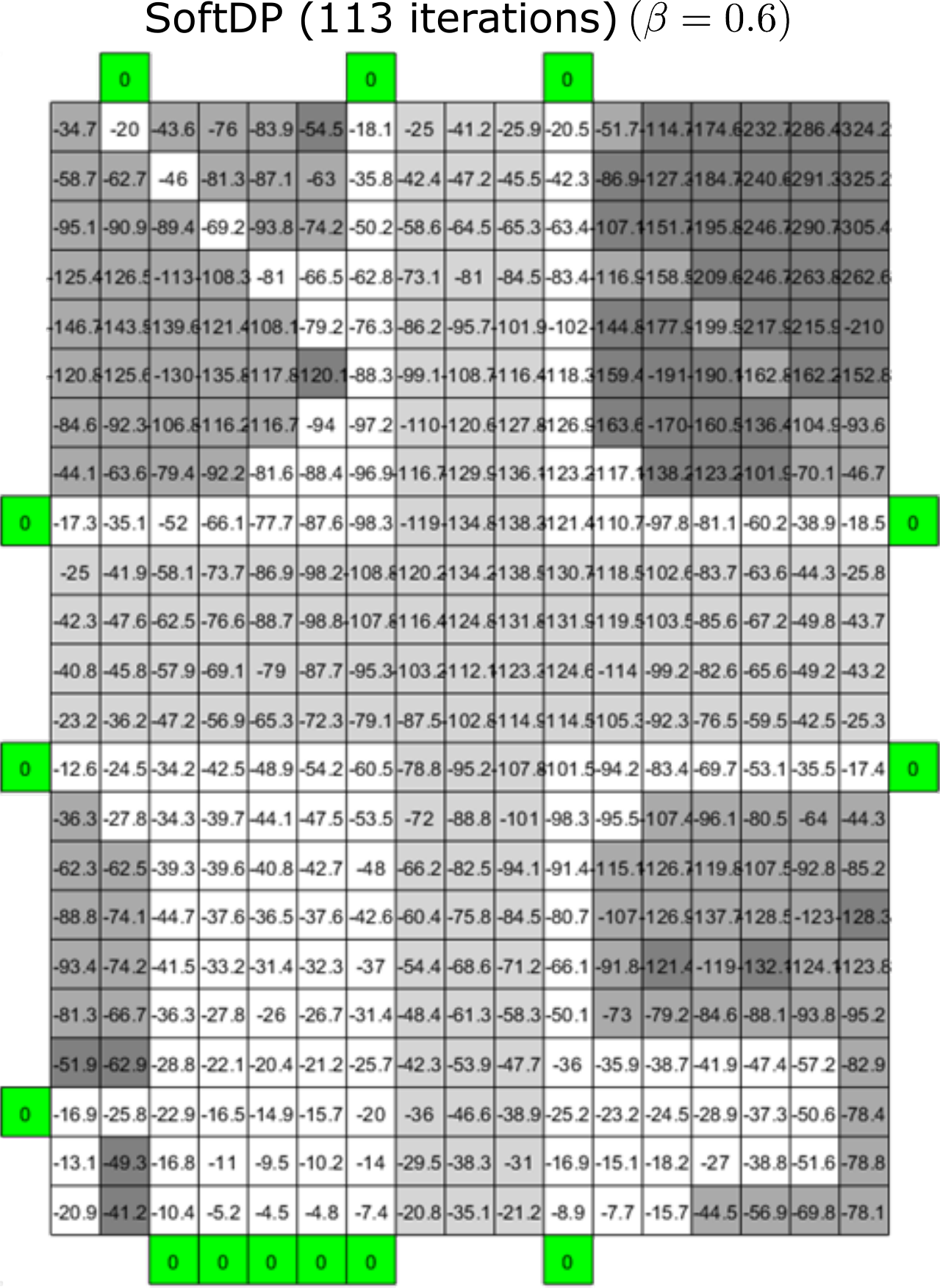}}
  \subfigure{\includegraphics[width=0.28\linewidth]{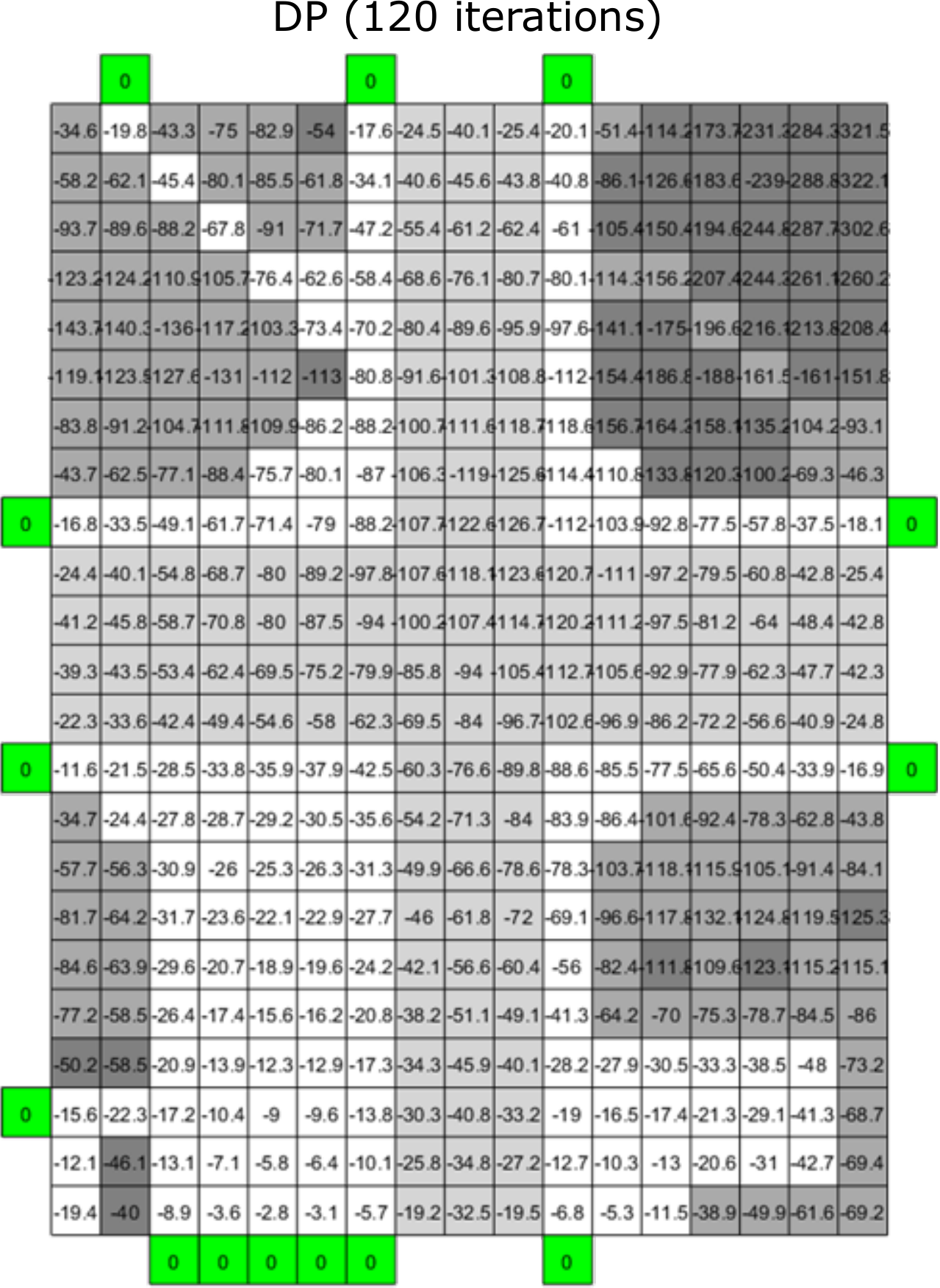}}
  \subfigure{\includegraphics[width=0.28\linewidth]{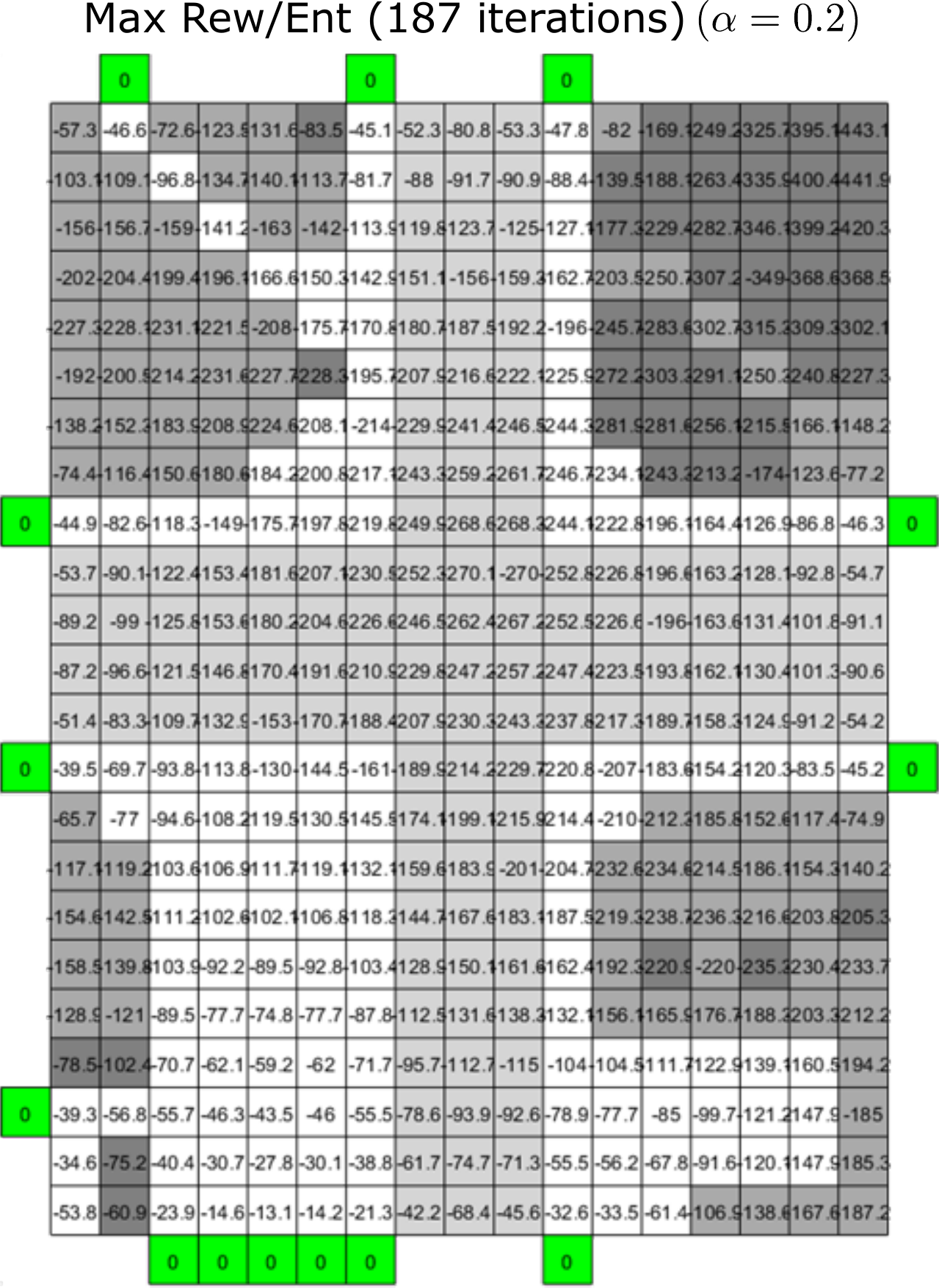}}
  \subfigure{\includegraphics[width=0.28\linewidth]{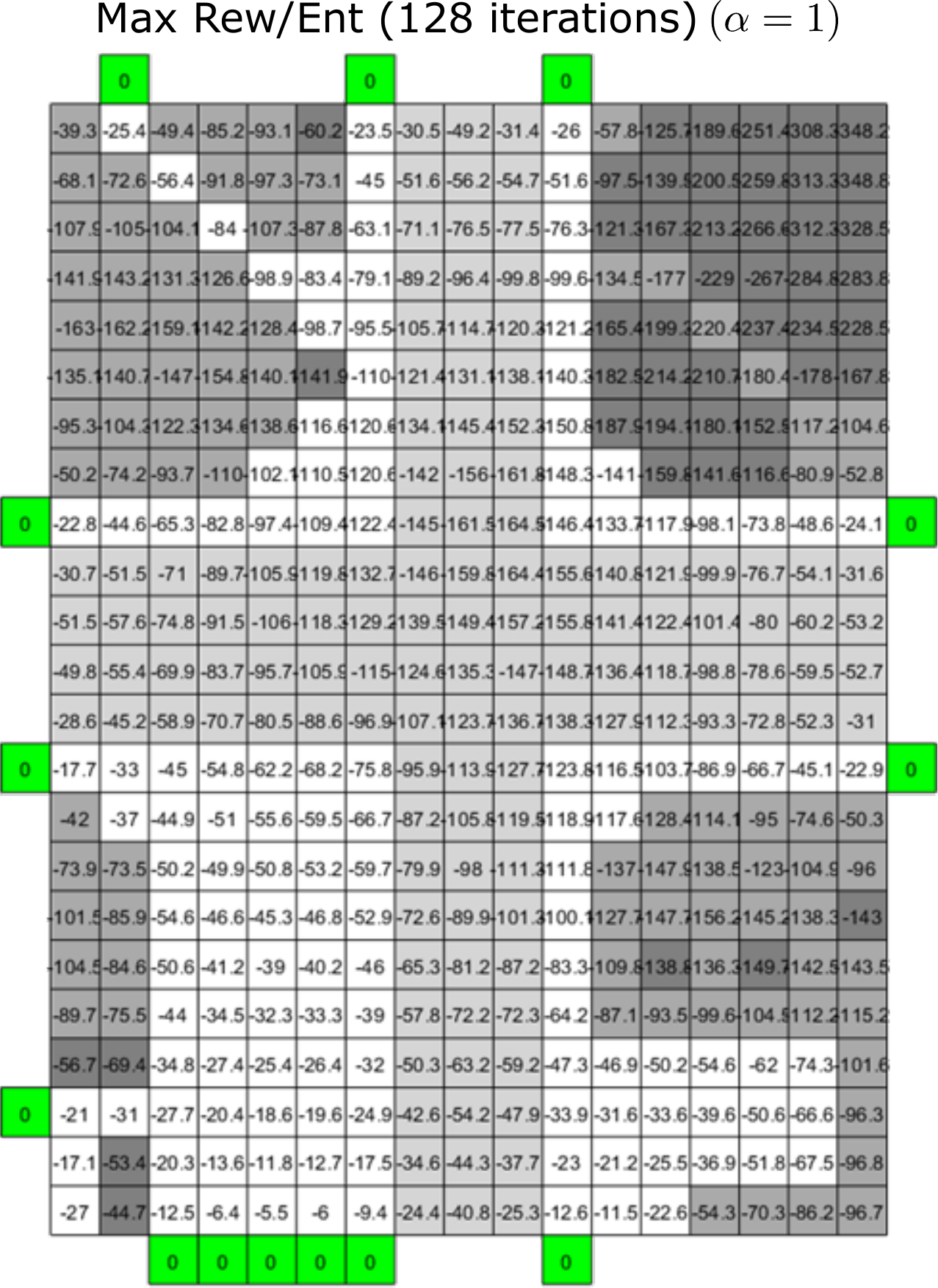}}
  \subfigure{\includegraphics[width=0.28\linewidth]{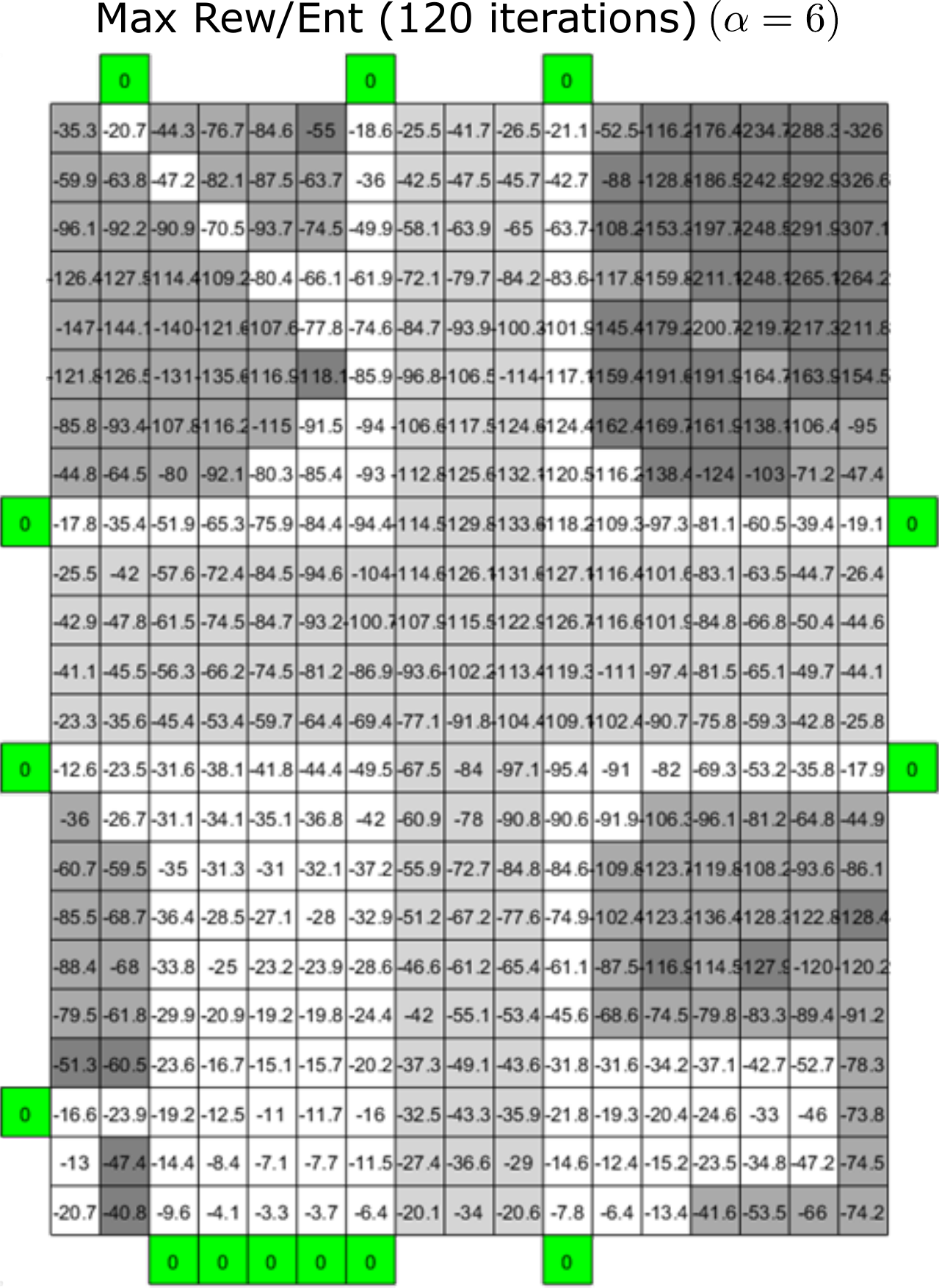}}
  \caption{Numerical visualization of the value function for the various algorithms.}
 \label{fig:Value}
\end{figure}

\begin{figure}[ht]
  \centering
  \subfigure{\includegraphics[width=0.32\linewidth]{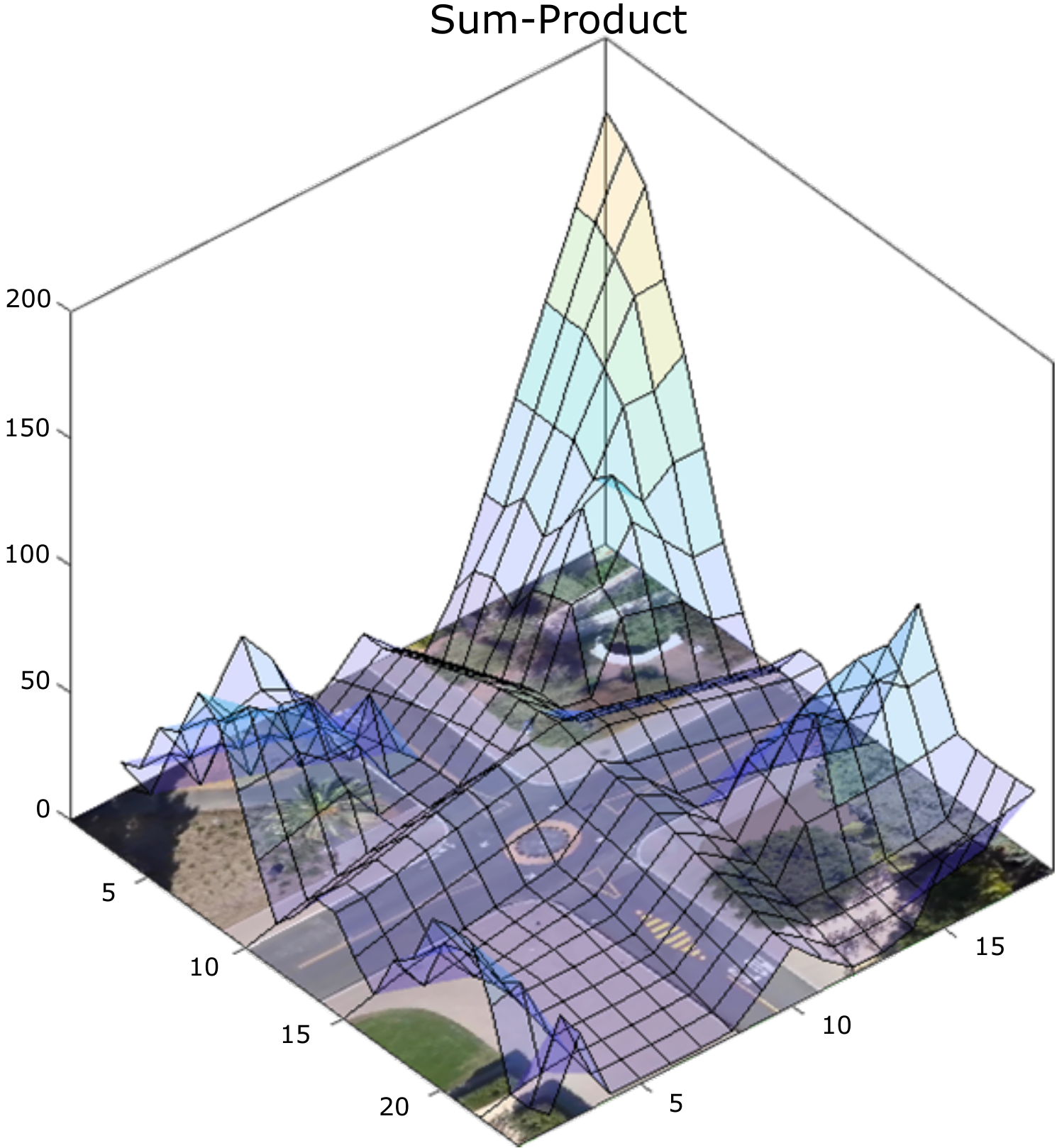}}
  \subfigure{\includegraphics[width=0.32\linewidth]{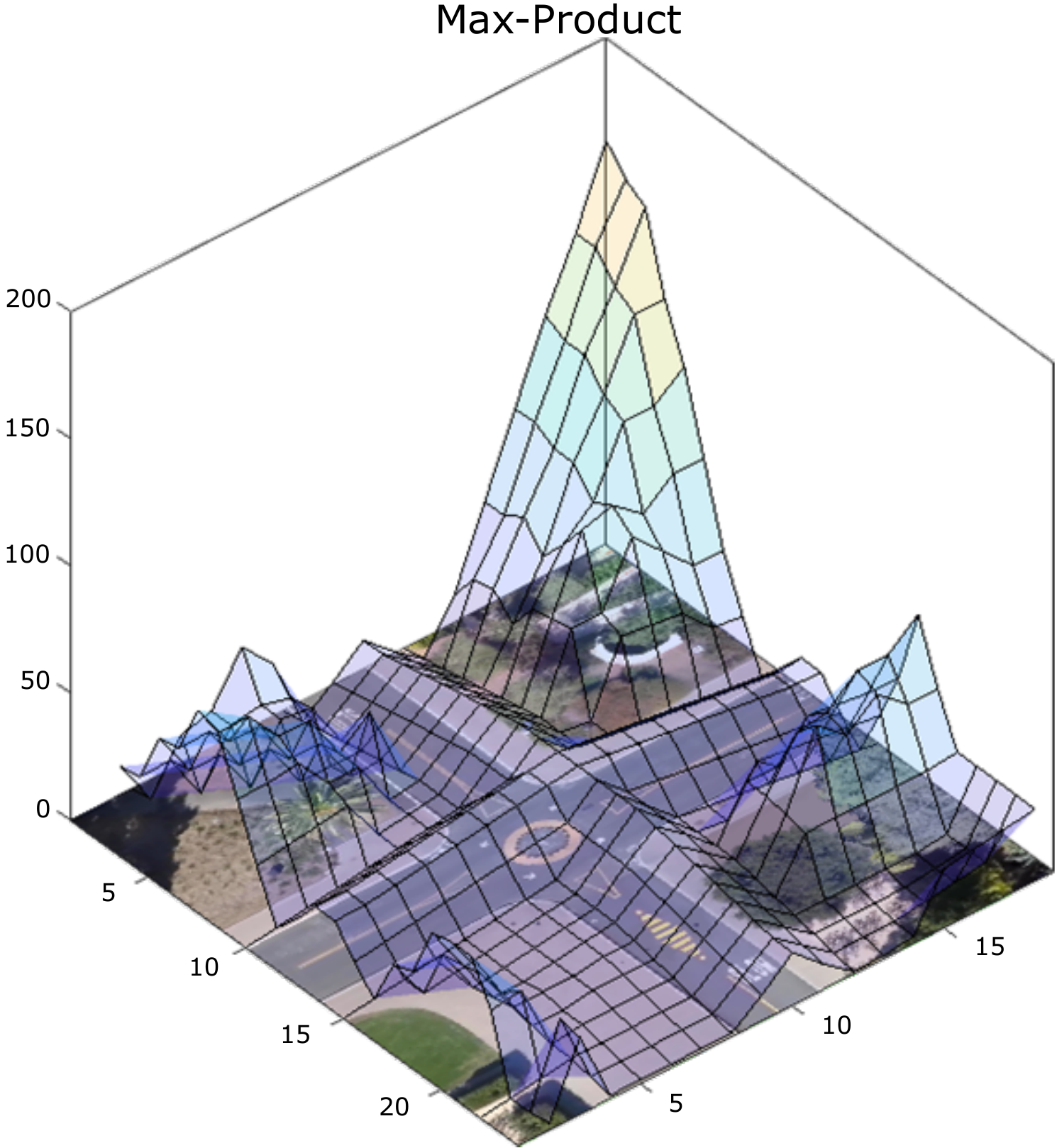}}
  \subfigure{\includegraphics[width=0.32\linewidth]{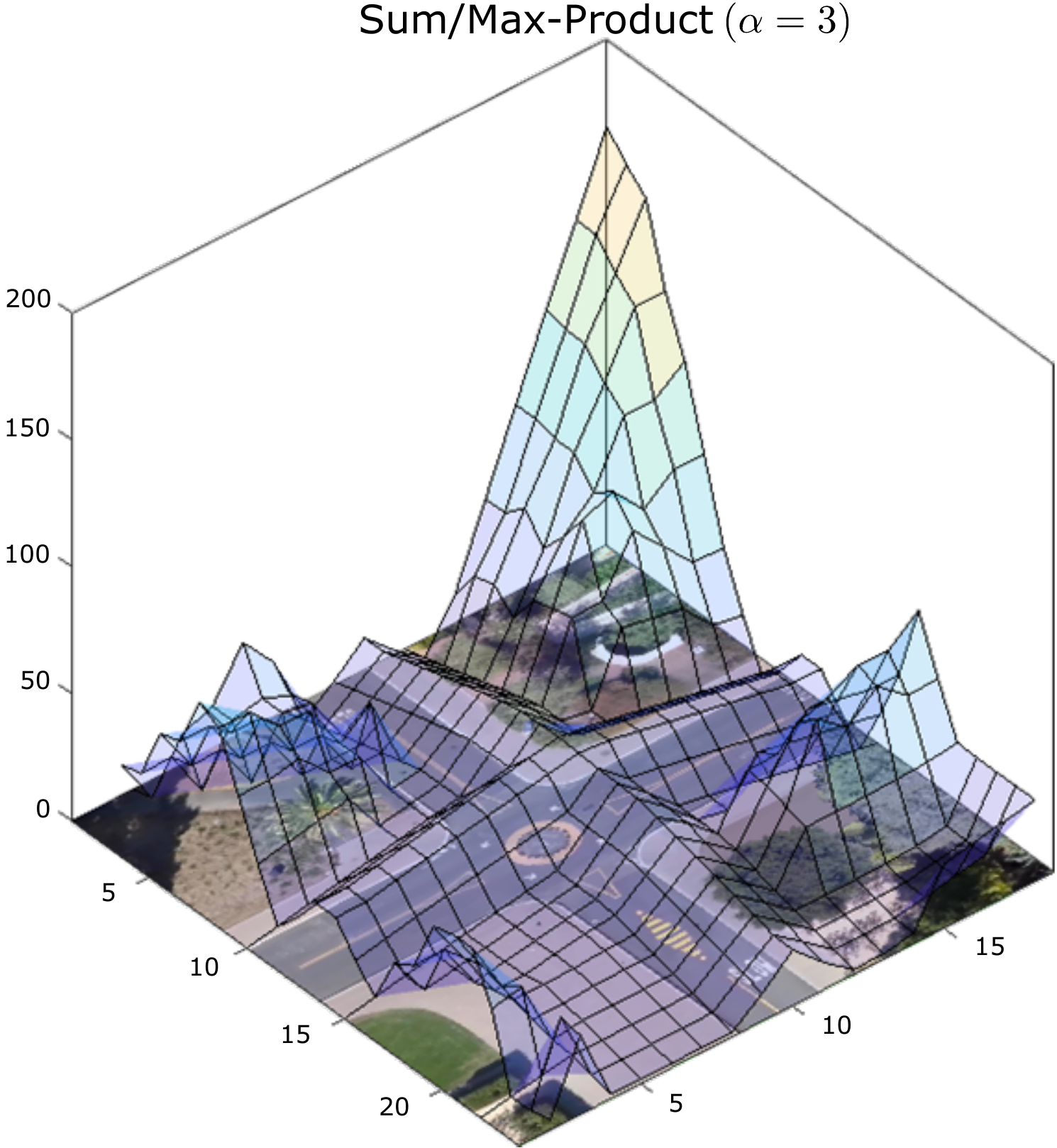}}
  \subfigure{\includegraphics[width=0.32\linewidth]{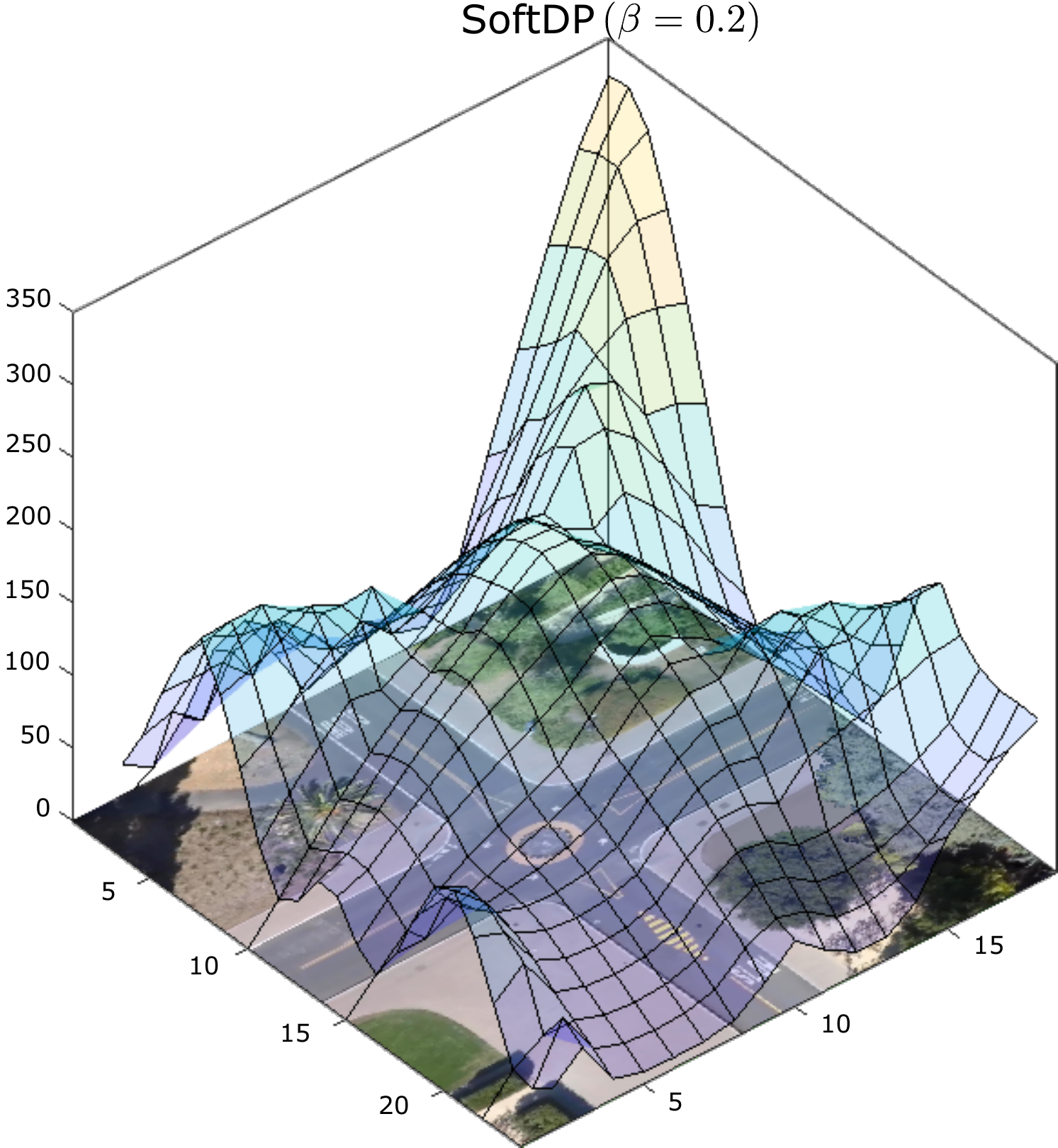}}
  \subfigure{\includegraphics[width=0.32\linewidth]{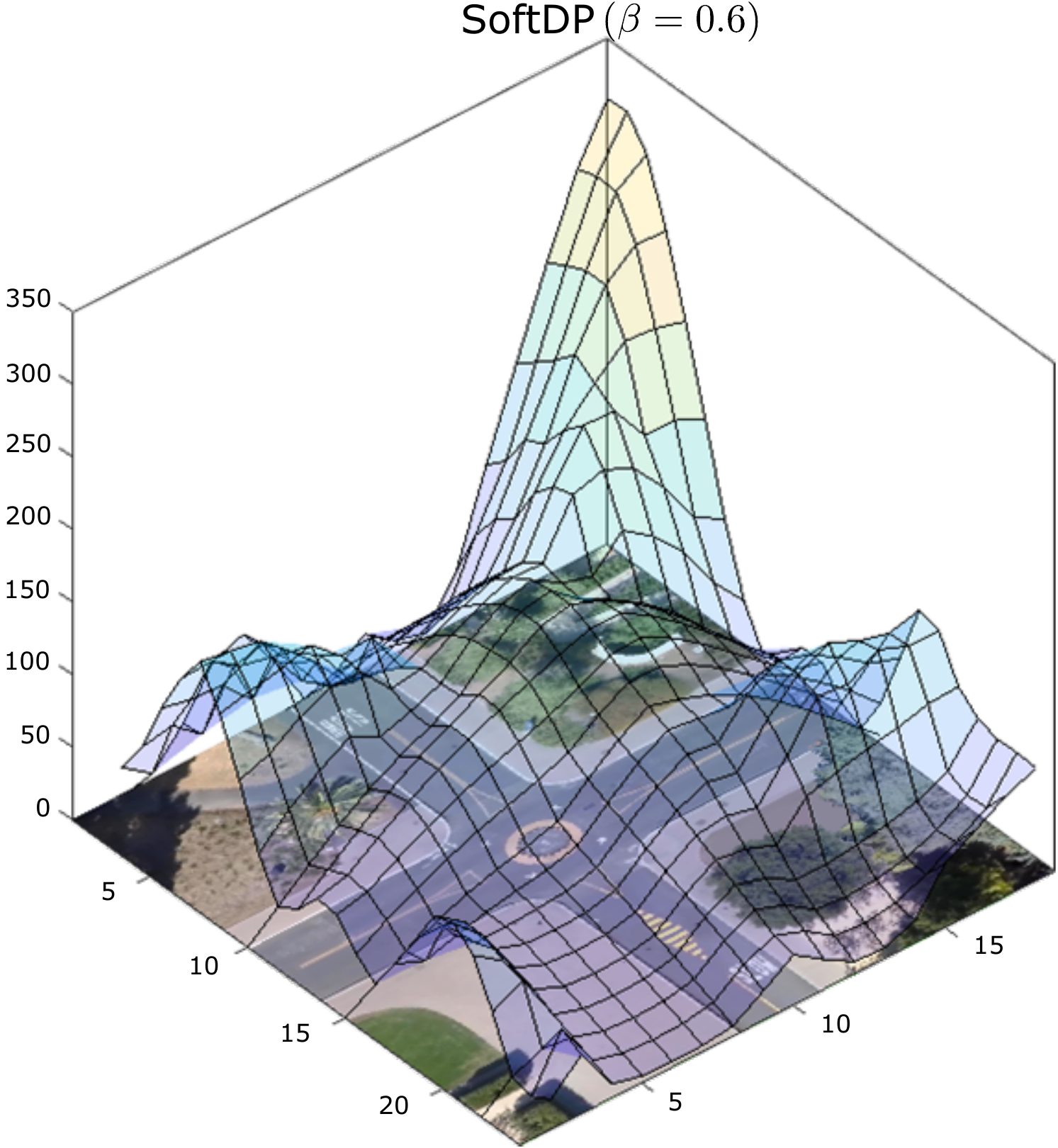}}
  \subfigure{\includegraphics[width=0.32\linewidth]{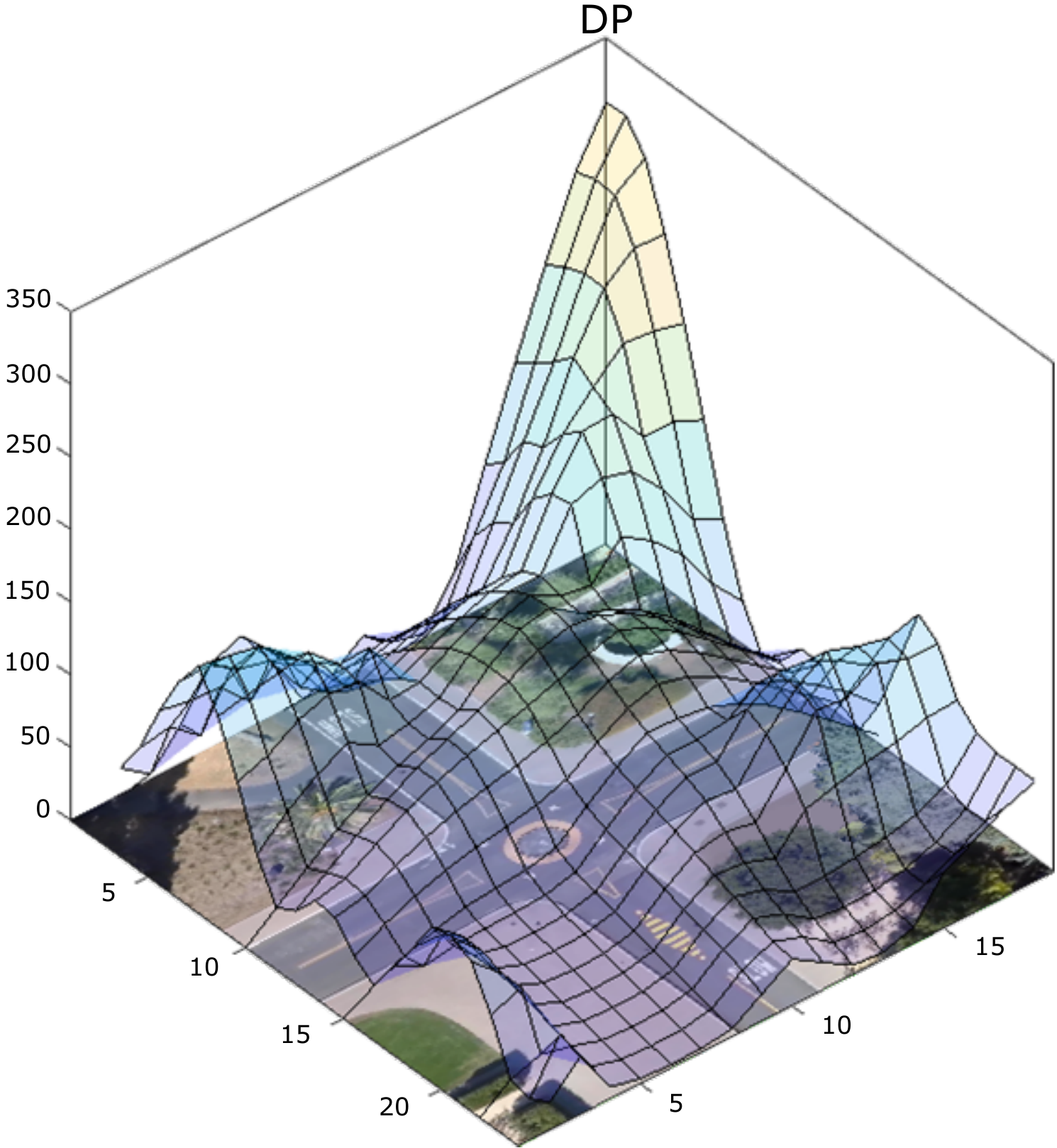}}
  \subfigure{\includegraphics[width=0.32\linewidth]{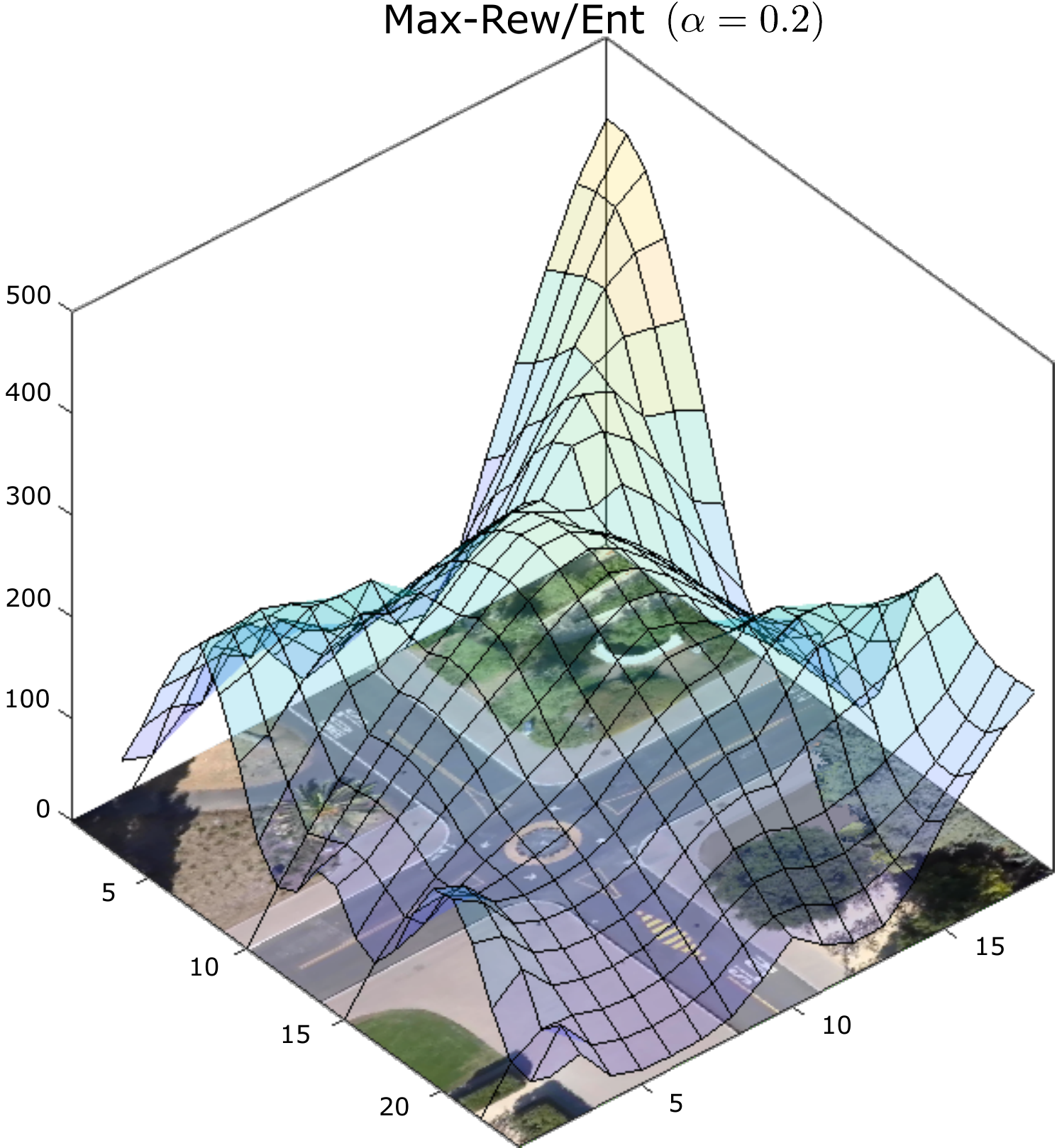}}
  \subfigure{\includegraphics[width=0.32\linewidth]{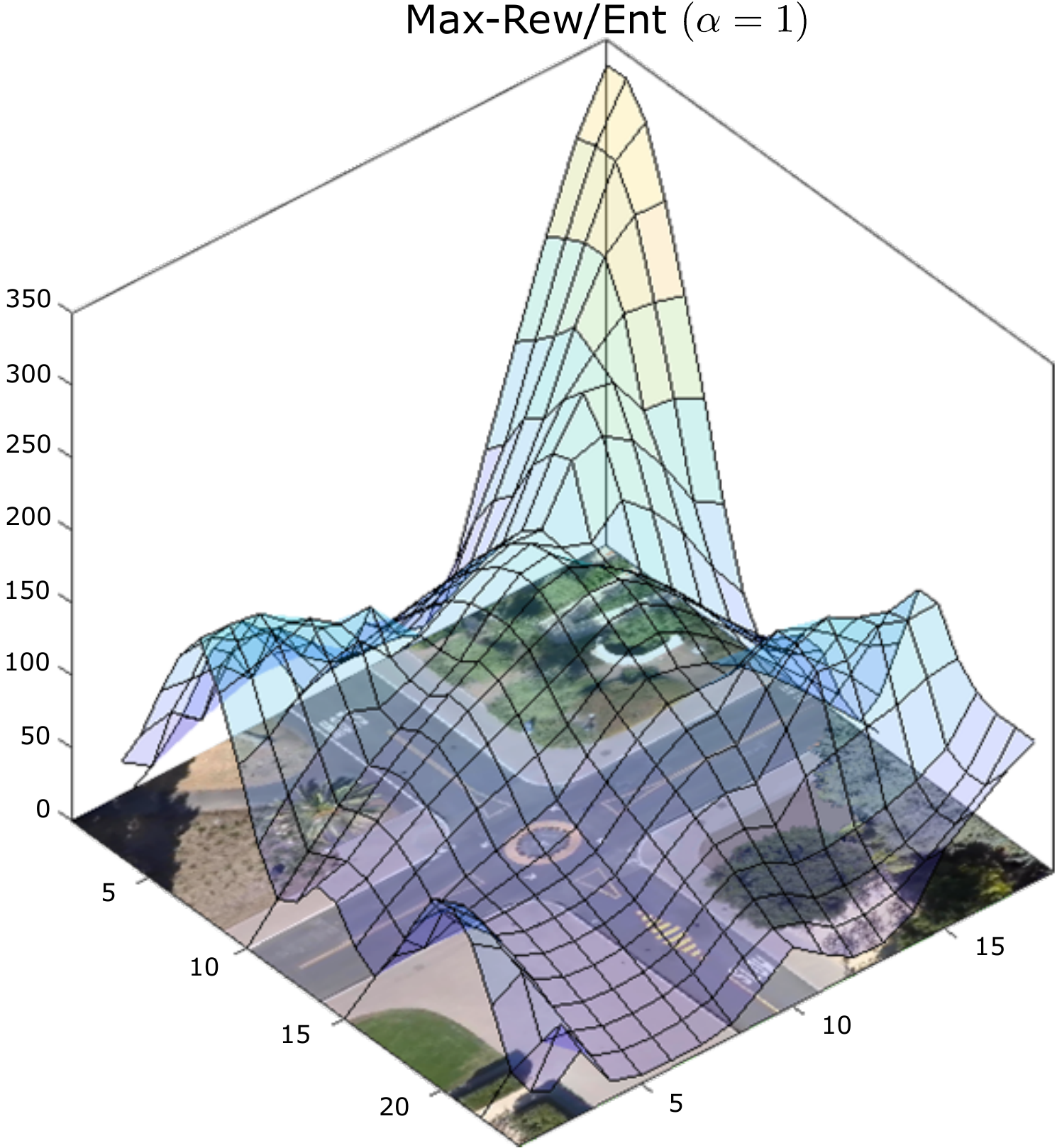}}
  \subfigure{\includegraphics[width=0.32\linewidth]{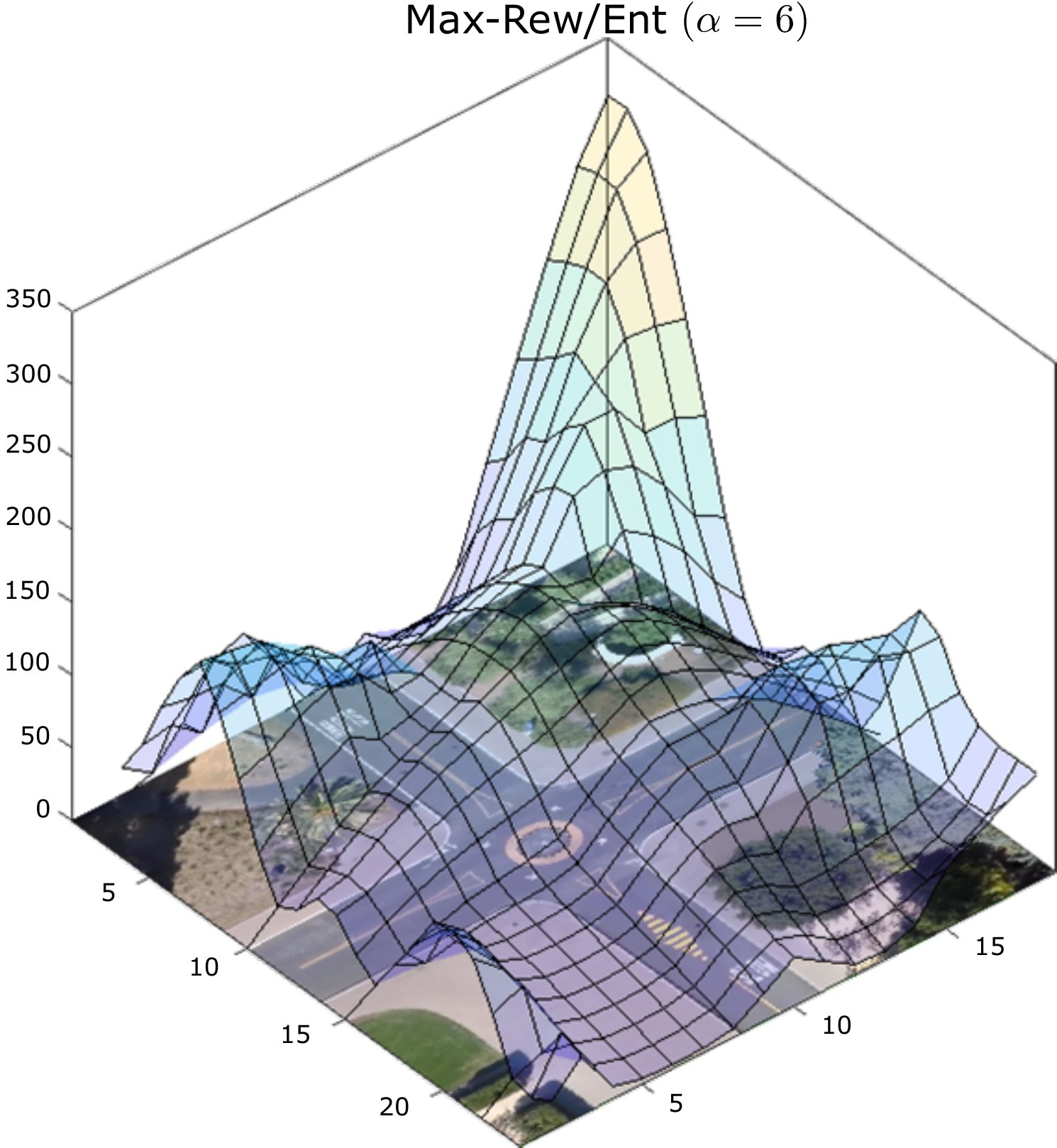}}
  \caption{3D Plots of $ -V_{S_t}(s_t)$ for the various algorithms. }
  \label{fig:3Dplots}
\end{figure}

\begin{figure}[ht]
  \centering
  \includegraphics[width=\linewidth]{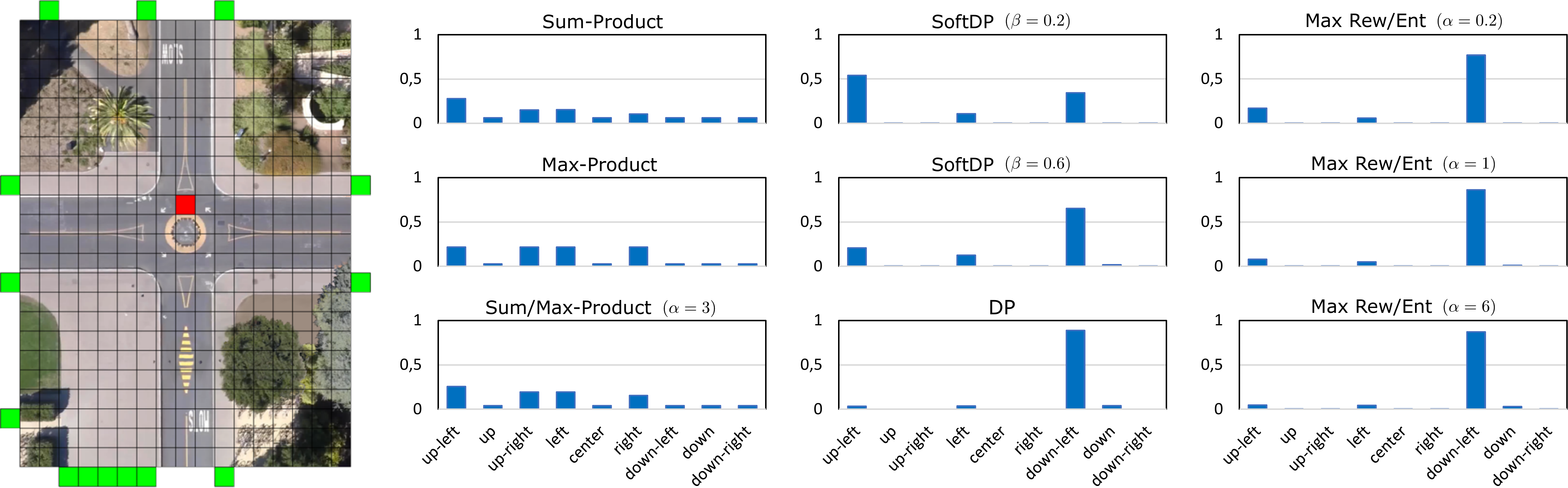} 
  \caption{Policy distributions at a generic point (red on the map) for the various methods for different parameter choices. The goals are depicted in green.}
  \label{fig:PointPolicy}
\end{figure}

\begin{figure}[ht]
  \centering
  \includegraphics[width=\linewidth]{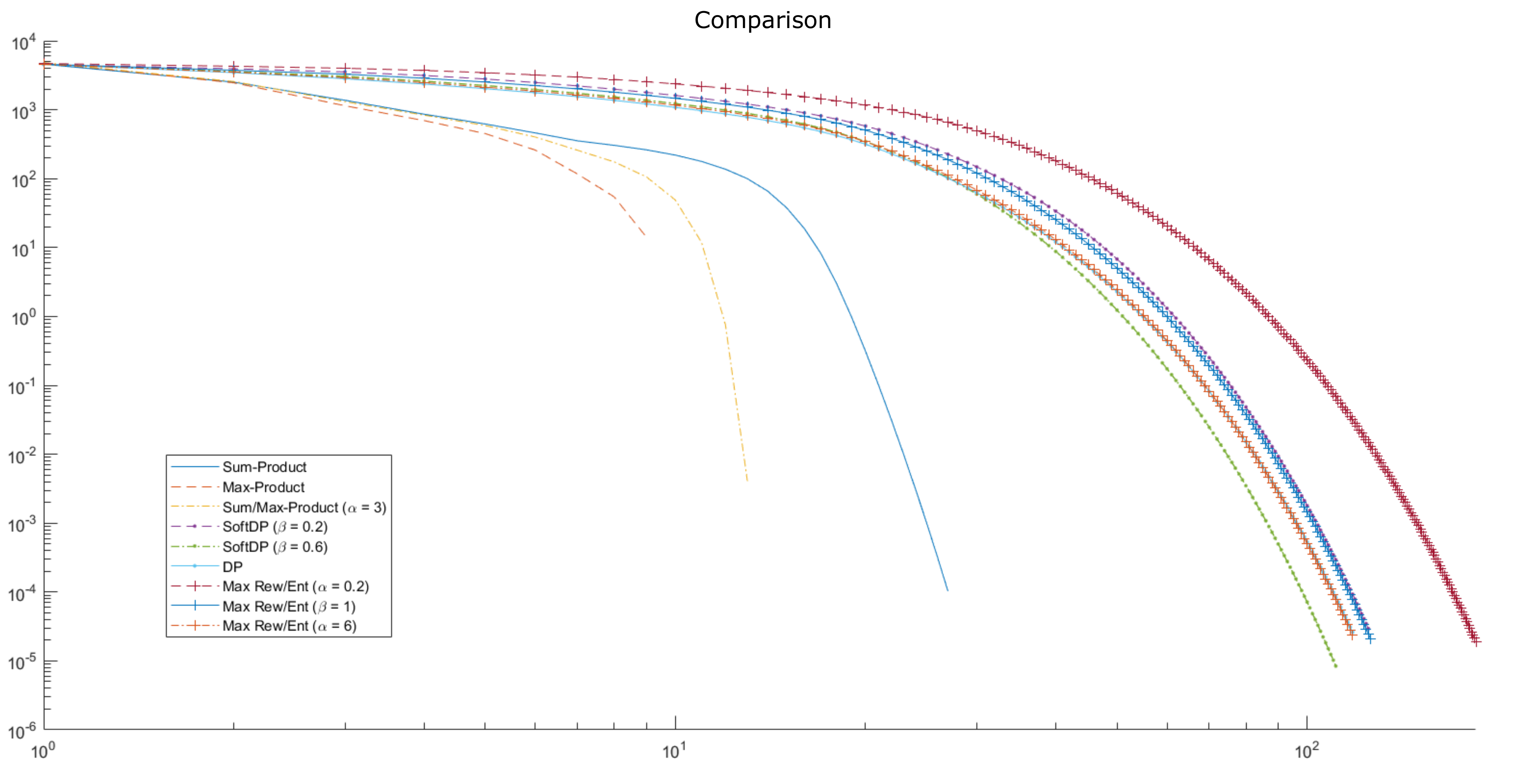} 
  \caption{Comparison of the value function increments for the various algorithm for the example of Figure \ref{fig:Policy}. }
  \label{fig:comp}
\end{figure}

\section{Simulations}
\label{sec:simulations}

We have simulated the various recursions for path planning  on two discrete grids.

The first set of simulations is performed on a  small 6x6 square grid shown in Figure \ref{fig:Maze},  where we have one goal (bull's eye and green) and obstacles (dark gray). The states are the positions on the grid and the actions correspond to one of the nine possible one-pixel motions $\lbrace$up-left, up, up-right, left, center (still), right, down-left, down, down-right$\rbrace$. The reward function has the values $0$ on the goal, $-10$ on the obstacles and $-1$ on other pixel positions.  The motion is stochastic with a transition function $p(s_{t+1}|s_ta_t)$ that has probability $1/2$ for the intended direction and the rest spread equally on the other eight directions. Built in the transition function are also re-normalizations when the transition is close to the boundaries: when some of the new projected states are outside the grid, their probabilities are set to zero, and the remaining probability is spread equally on the other  pixels. No initial or final conditions are set on the model. The recursions are run until convergence to a steady state value function.  All the algorithms lead to policies that would allow an agent, starting from any position on the grid, to reach the goal in a finite number of steps. The values reported in the squares and the max policy arrows in Figure \ref{fig:Maze} reveal how the different  solutions direct our potential agent in slightly different paths to avoid the obstacles. In the lower right corner of the figure, we also report the increments in reaching the steady-state solution for the various algorithms in a log-log graph (also the parameters are reported in the legend). The algorithm is stopped after all the increments in the value function are below $10^{-5}$.  It is noteworthy to see how  the Max-product algorithm reaches the steady-state solution in a very limited number of steps (fastest convergence) and how the Sum-product and the Sum/Max-product algorithms converge at a much faster rate in comparison to the others.   

The results of another set of simulations are reported in Figures \ref{fig:Policy}, \ref{fig:Value}, \ref{fig:3Dplots},  \ref{fig:PointPolicy} and \ref{fig:comp}. Here, we have a grid extracted from a real dataset acquired at an intersection on the Stanford campus with pedestrians and bikes. The scene, with no agents, is simplified to  $17 \times 23$ pixels,  with goals (exits) and  rewards assigned to various areas (semantic map) as shown in Figure \ref{fig:Policy}. We assume that our agent is a pedestrian and the actions are the same nine actions we have used above for the small grid. The rewards are: $R(s_t)=$ 0 (goals: bull's eye and green); -1 (pedestrian walkways: white); -10 (streets: light gray); -20 (grass: dark gray); -30 (obstacles: dark).   The convergence behavior to a steady-state value function is similar to the one shown for the smaller grid. The number of iterations to reach a precision of $10^{-5}$ on all the states are: [Sum-product: 29; Max-product: 11; Sum/Max-product ($\alpha=3$): 15; Soft DP ($\beta=0.2$): 127; Soft DP ($\beta=0.6$): 113; DP: 120; Max Rew/Ent ($\alpha=0.2$): 187; Max Rew/Ent ($\alpha=1$): 128; Max Rew/Ent ($\alpha=6$): 120]. Note how quickly the Max-product and the probabilistic methods converge when compared  to the others. The graph of the actual increments for the various algorithm is shown in log scale in Figure \ref{fig:comp}. 
Figure \ref{fig:Policy} shows, for each state, the maximum policy directions for all the algorithms. The arrows point towards the preferences imposed by the semantic information: pedestrians prefer walkways to streets; grass and obstacles are avoided. We observe a marked effect on the results of the Max-product algorithm that maintains the multiple maxima directions corresponding to the equivalent solutions. These multiple options appear smoothed out in the other methods. The Max-product requires just the minimum number of steps to stabilize its configurations. Figure \ref{fig:3Dplots} shows also some of the value functions $- V(s_t)$ (they can be thought as potential functions) superimposed on the original scene for the various methods and for some parameter choices. The comparison clearly shows that the various algorithm lead to intriguing comparable smooth solutions, except for the Max-product that produces a very sharp value function with very well defined valleys. Just as in the simulations on the small grid, we have included no paths on the map,  because in all methods an agent that starts anywhere, will reach one of the goals in all cases. This can be easily verified by following the arrows in Figure  \ref{fig:Policy}. A much more revealing visualization of the differences among the various methods is displayed in Figure  \ref{fig:PointPolicy}, where at a generic point on the map, we plot the policy distributions. In the first column, the policies for the probabilistic methods are shown with the Max-product clearly producing a  rather sharp behavior with all  the multiple equivalent options. Recall that the map has multiple goals and the agent in that position has more than one option to achieve optimality (see also Figure \ref{fig:Policy} in that position).  In the second column, we report the results of the DP approach in its standard form (bottom graph) and in its soft parametrized versions.  Note how, for the two values $\beta=0.3$ and $\beta=0.6$, an agent may be led  to consider more options with respect to DP and if we look also at the maximum policy on the map of Figure \ref{fig:Policy}, it may even be taken on a different path. In the third column, we report the results of the Max Rew/Ent algorithm for various values of the parameter $\alpha$. We notice,  as expected from the theory, that when $\alpha <1$, the policy distribution is more entropic and that when $\alpha$ increases, the distribution tends to the DP policy.

\section{Conclusions and future directions}
\label{sec:conclusions}

In this paper, we have provided a unified view on the optimal solutions to path planning using a probability formulation on a factor graph. The various methods correspond to different combination rules through two of the blocks on a Factor Graph in Reduced Normal Form and correspond to different cost functions that go from maximum a posteriori to maximum combination of reward and entropy.
We have generalized some of the algorithms previously presented in the literature by using parametrized soft-max functions. The resulting set of choices, presented here in a compact fashion, both in the probability and in the log space,  may enhance the algorithmic design options for  reinforcement learning  that needs to obtain the V-functions or the Q-functions, perhaps on the basis of current model knowledge.  We have included typical results on  discrete grids that reveal marked differences among some of the methods. Our computational results suggest that the Max-product algorithm,  optimal maximum a posteriori solution,  together with the other probabilistic methods, such as the Sum-product and its Sum/Max-product generalizations, shows fast converge to the steady-state configuration in comparison to the other reward-based methods, that are typically derived in the log space. This general approach to the topic may allow agile use of the various methods during exploration in RL. Further work on the topic will be devoted to extensions to continuous space and relative approximations, and  applications involving interacting agents.

\appendix

\section{Soft-max functions}  
\label{app:smax}

We review here some of the soft-max functions that are used in the recursions discussed in the paper. For all the functions, we start from a set or real numbers $x_1, \dots ,x_N$ and consider the 
ranked set $x_{(1)},x_{(2)}, \dots ,x_{(N)}$, with $x_{(1)}\le x_{(2)}\le \dots \le x_{(N)}$. 

{\bf A.} Consider the following expression  
\begin{equation*}
  s(x_1, \dots ,x_N)= \log \sum_{j=1}^Ne^{x_j}.
\end{equation*} 
This function has the property that when $x_1, \dots ,x_N \rightarrow +\infty$, $s(x_1, \dots ,x_N) \rightarrow \max (x_1, \dots ,x_N)$. {\em Proof:} The function can rewritten as  
\begin{equation*}
  \begin{array}{rL}
     s(x_1, \dots ,x_N) & = \log \left(e^{x_{(1)}}+e^{x_{(2)}} + \cdots +e^{x_{(N)}} \right) \\
    & = \log e^{x_{(N)}}\left(e^{x_{(1)}-x_{(N)}}+e^{x_{(2)}-x_{(N)}} + \cdots + 1 \right). 
  \end{array}
\end{equation*} 
When the $x_i$s become large, also their difference to the max $x_{(N)}$ becomes a large negative number. Therefore  the first $N-1$ terms inside the parenthesis tend to zero and $s(x_1, \dots ,x_n) \rightarrow x_{(N)}$.  

{\bf B.} A {\em parametrized soft-max function} can be defined as the expression
\begin{equation*}
  g(x_1, \dots ,x_N;\alpha) = \frac{1}{\alpha} \log \sum_{j=1}^Ne^{\alpha x_j}
\end{equation*} 
where $\alpha \ge 1$.  This function has the property that  
\begin{equation}
  \lim_{\alpha \to \infty} g(x_1, \dots ,x_N;\alpha) = \max (x_1, \dots, x_N).
\end{equation} 
{\em Proof:} Using  the ranked set, the function can be bounded as 
\begin{equation}
  \frac{1}{\alpha} \log e^{\alpha x_{(N)}} \le g(x_1, \dots, x_N;\alpha) \le \frac{1}{\alpha} \log N e^{\alpha x_{(N)}},  
\end{equation} 
\begin{equation*}
  x_{(N)} \le g(x_1, \dots, x_N;\alpha) \le \frac{\log N}{\alpha} + x_{(N)},  
\end{equation*} 
that for $\alpha \to \infty$ achieves the maximum $x_{(N)}$. Note that for $\alpha > 1$, from the bound,  the soft-max always exceeds the maximum value, i.e.,  tends to $x_{(N)}$ from the right.  

It is useful to look at the expression when $0< \alpha<1$. When $\alpha$ is a very small positive number 
\begin{equation*}
  g(x_1, \dots,x_N;\alpha) \simeq \frac{1}{\alpha} \log N + \mu, 
\end{equation*}
where $\mu=(1/N) \sum_{i=1}^N x_i$ is the arithmetic mean. The function diverges for $\alpha \to 0$,  but when  considered for small values of $\alpha$, the function  tends to become  independent on any specific $x_i$.    

{\em Proof:} The function can be written as
\begin{equation*}
  g(x_1, \dots,x_N;\alpha)= \frac{1}{\alpha} \log \left( \sum_{i=1}^N e^{\alpha (x_i - \mu)} e^{\alpha \mu} \right) = 
  \frac{1}{\alpha} \log \sum_{i=1}^N e^{\alpha (x_i - \mu)} + \mu. 
\end{equation*}
When $\alpha$ approaches zero, the exponents become $\simeq 1$ and we have the result. 

{\bf C.} Another {\em parametric soft-max function} is 
\begin{equation*}
  h(x_1,\dots,x_N;\alpha) = \left( \sum_{j=1}^N x_j^\alpha \right)^\frac{1}{\alpha},
\end{equation*}
where here $x_i \ge 0, i=1:N$. 
Here too, for  $\alpha \to \infty$, $h(x_1,\dots,x_N;\alpha) \rightarrow x_{(N)}$. 

\noindent
{\em Proof:} From the bounds
\begin{equation*}
  \begin{array}{rLL}
    (x_{(N)}^{\alpha})^\frac{1}{\alpha} & \le h(x_1,\dots,x_N;\alpha) & \le \left( N x_{(N)}^\alpha \right)^\frac{1}{\alpha}, \\
    x_{(N)} & \le h(x_1,\dots,x_N;\alpha) & \le  N^\frac{1}{\alpha} x_{(N)}, 
  \end{array}
\end{equation*}
when $\alpha$ grows $N^\frac{1}{\alpha} \to 1$ and the function tends to $x_{(N)}$. In all soft-max functions, the larger values in the set $x_1,\dots,x_N$, tend to dominate over the others when they become large. 

The function $h(x_1,\dots,x_N;\alpha)$, just as  $g(x_1,\dots,x_N;\alpha)$, diverges for $\alpha \to 0$, but for small $\alpha$
\begin{equation*}
  h(x_1,\dots,x_N;\alpha) \simeq  N^\frac{1}{\alpha},   
\end{equation*} 
which, again as in $g$,  deas not depend on any of the $x_i$. 

\noindent
{\em Proof:} easily seen as $x_i^\alpha \simeq 1$ for small $\alpha$. 

{\bf D.} Another soft-max functions can be defined as  
\begin{equation*}
  r(x_1,\dots,x_N;\alpha) = \frac{\sum_{i=1}^N x_i e^{\alpha x_i}}{\sum_{j=1}^N e^{\alpha x_j}}.  
\end{equation*} 
This function is well-known in the neural network literature, where the vector function 
$e^{\alpha x_i} / \sum_j e^{\alpha x_j}$ tends to a distribution concentrated on the maximum. By taking the expectation with such a distribution we get the soft-max. Therefore,
when $\alpha \rightarrow \infty$, $r(x_1,\dots,x_N;\alpha) \rightarrow x_{(N)}$. 

\noindent
{\em Proof:} The function can re-written using the ranked set as
\begin{equation*}
  r(x_1,\dots,x_N;\alpha) = \frac{\sum_{i=1}^{N-1} x_{(i)}e^{\alpha x_{(j)}} + x_{(N)} e^{\alpha x_{(N)}}}{\sum_{j=1}^{N-1} e^{\alpha x_{(j)}} +   
  e^{\alpha x_{(N)}}} = \frac{ \sum_{i=1}^{N-1} x_{(i)}e^{\alpha (x_{(j)}-x_{(N)})} + x_{(N)}}{\sum_{j=1}^{N-1} e^{\alpha (x_{(j)}-x_{(N)})} + 1}. 
\end{equation*}
For $\alpha \rightarrow \infty$ both summations tend to zero, because the exponents are negative, and we have the result. 

This soft-max function, when $\alpha \rightarrow 0^+ $ does not diverge, but tends to the arithmetic mean $r(x_1,\dots,x_N;\alpha) \rightarrow 1/N \sum_{i=1}^N x_i$. 

\noindent
{\em Proof:} Trivial, because for $\alpha=0$ all the exponentials are equal to one. 

\newpage
\section{Optimizing Reward and Entropy}
\label{app:levgen}
To better understand the nature of the function being optimized in (\ref{eq:costgen}), and how it gives rise to an entropy term, let us write it explicitely for $T=4$, using the compact notation $R'(s_ta_t) = R(s_ta_t) + \log p(a_t)$. The function to optimize is 
\begin{equation}
  \begin{array}{rL}
    \sum_{s_1...s_4} & \sum_{a_1...a_4} p(s_1) \pi_\alpha(a_1|s_1) p(s_2|s_1a_1) \pi_\alpha(a_2|s_2) p(s_3|s_2a_2) \pi_\alpha(a_3|s_3) p(s_4|s_3a_3)   
      \pi_\alpha(a_4|s_4) \\
    & \left[ R'(s_1a_1) - \frac{1}{\alpha} \log \pi_\alpha(a_1|s_1) + R'(s_2a_2) - \frac{1}{\alpha} \log \pi_\alpha(a_2|s_2) + R'(s_3a_3) \right. \\
    & \left.  - \frac{1}{\alpha} \log \pi_\alpha(a_3|s_3) + R'(s_4a_4) - \frac{1}{\alpha} \log \pi_\alpha(a_4|s_4) \right],
  \end{array}
\end{equation}

Starting from the last term, we identify the backward  recursions 
\begin{equation}
  \begin{array}{L}
     \underbrace{ \frac{1}{\alpha}  \sum_{s_4}p(s_4|s_3a_3) \left[ \underbrace{\sum_{a_4} \pi_\alpha (a_4|s_4) \alpha R'(s_4a_4) +  
      \overbrace{\sum_{a_4}\pi_\alpha (a_4|s_4) \log \frac{1}{\pi_\alpha (a_4|s_4)} }^{{\cal H}(\pi_\alpha (a_4|s_4))}}_{V(s_4)} \right]}_{Q(s_3a_3)} \\ 
     \underbrace{ \frac{1}{\alpha} \sum_{s_3}p(s_3|s_2a_2) \left[ \underbrace{\sum_{a_3} \pi_\alpha (a_3|s_3) (\alpha R'(s_3a_3)+\alpha Q(s_3a_3)) +  
      \overbrace{\sum_{a_3} \pi_\alpha(a_3|s_3) \log \frac{1}{\pi_\alpha (a_3|s_3)} }^{{\cal H}(\pi_\alpha (a_3|s_3))}}_{V(s_3)} \right]}_{Q(s_2a_2)} \\ 
     \underbrace{ \frac{1}{\alpha} \sum_{s_2}p(s_2|s_1a_1) \left[ \underbrace{\sum_{a_2} \pi_\alpha (a_2|s_2) (\alpha R'(s_2a_2)+\alpha Q(s_2a_2)) +  
      \overbrace{\sum_{a_2} \pi_\alpha(a_2|s_2) \log \frac{1}{\pi_\alpha(a_2|s_2)} }^{{\cal H}(\pi_\alpha (a_2|s_2))}}_{V(s_2)} \right]}_{Q(s_1a_1)} \\ 
    \frac{1}{\alpha} \sum_{s_1} p(s_1) \left[ \underbrace{\sum_{a_1} \pi_\alpha (a_1|s_1) (\alpha R'(s_1a_1)+\alpha Q(s_1a_1)) +  \overbrace{\sum_{a_1} 
      \pi(a_1|s_1) \log \frac{1}{\pi_\alpha(a_1|s_1)} }^{{\cal H}(\pi_\alpha (a_1|s_1))}}_{V(s_1)} \right]
  \end{array}
\end{equation}

Note how the value function $V(s_t)$ (not optimized here) is written as a recursive superposition of reward and policy entropy. The parameter $\alpha$ controls the balance between the two terms and the power of the distribution. Note that the policy function multiplies also the reward term. Therefore, the optimized policy distribution will shape, in a non trivial way, the effects of the rewards with respect to the entropy.

Following Levine's approach \citep{Levine2018}, using our modified cost function, we search for the best policy distribution starting from re-writing the last term using the KL-divergence
\begin{equation}
  \begin{array}{L}
    \frac{1}{\alpha} \sum_{s_4}p(s_4|s_3a_3) \left[ \sum_{a_4} \pi_\alpha (a_4|s_4) \left( \alpha R'(s_4a_4) - \log \pi_\alpha (a_4|s_4) \right) \right] = \\ 
    \frac{1}{\alpha} \sum_{s_4}p(s_4|s_3a_3) \left[ \sum_{a_4} \pi_\alpha(a_4|s_4) \left( \log \frac{e^{\alpha R'(s_4a_4)}}{\pi_\alpha(a_4|s_4)} 
      \frac{\sum_{a_4'} e^{\alpha R'(s_4a_4')}}{\sum_{a_4'} e^{\alpha R'(s_4a_4')} } \right) \right] = \\
    \frac{1}{\alpha} \sum_{s_4}p(s_4|s_3a_3) \left[ -{\cal D}_{KL} \left( \pi_\alpha(a_4|s_4)\left\| \frac{e^{\alpha R'(s_4a_4)}}{\sum_{a_4'} e^{\alpha 
      R'(s_4a_4')}} \right. \right) + \log \underbrace{\sum_{a_4'} e^{\alpha R'(s_4a_4')}}_{e^{ \alpha V(s_4)}} \right] = \\
    \sum_{s_4}p(s_4|s_3a_3) \left[ -{\cal D}_{KL} \left( \pi_\alpha(a_4|s_4)\left\|\frac{e^{\alpha R'(s_4a_4)}}{{e^{ \alpha V(s_4)}}} \right) + V(s_4) 
      \right. \right]. 
  \end{array}
\end{equation}

The optimum value is obtained when the ${\cal D}_{KL}(.\|.)=0$, i.e., when 
$\pi_\alpha (a_4|s_4)=\frac{e^{\alpha R'(s_4a_4)}}{{e^{ \alpha V(s_4)}}}$, 
and the optimal policy distribution is 
\begin{equation}
  \pi^*(a_4|s_4) \propto \frac{e^{ R'(s_4a_4)}}{{e^{ V(s_4)}}} = \frac{e^{Q(s_4a_4)}}{{e^{ V(s_4)}}},
\end{equation}
where we have defined $Q(s_4a_4)=R'(s_4a_4)$. Now the optimized expression $\sum_{s_4} p(s_4|s_3a_3) V(s_4)$ is carried over  
\begin{equation}
  \underbrace{R'(s_3a_3) + \sum_{s_4} p(s_4|s_3a_3) V(s_4)}_{Q(s_3a_3)} - \frac{1}{\alpha} \log \pi_\alpha(a_3|s_3).   
\end{equation}
Taking the expectation, we have
\begin{equation}
  \begin{array}{L}
    \frac{1}{\alpha} \sum_{s_3}p(s_3|s_2a_2) \left[ \sum_{a_3} \pi_\alpha(a_3|s_3) \left( \alpha Q(s_3a_3) - \log \pi_\alpha (a_3|s_3) \right) \right] = \\ 
    \frac{1}{\alpha} \sum_{s_3}p(s_3|s_2a_2) \left[ \sum_{a_3} \pi_\alpha (a_3|s_3) \left( \log \frac{e^{ \alpha Q(s_3a_3)}}{\pi_\alpha (a_3|s_3)} 
      \frac{\sum_{a_3'} e^{\alpha Q(s_3a_3')}}{\sum_{a_3'} e^{ \alpha Q(s_3a_3')}} \right) \right] = \\
    \frac{1}{\alpha} \sum_{s_3}p(s_3|s_2a_2) \left[ -{\cal D}_{KL} \left( \pi_\alpha (a_3|s_3) \left\| \frac{e^{\alpha Q(s_3a_3)}}{\sum_{a_3'} e^{ \alpha 
      Q(s_3a_3')}} \right. \right) + \log \underbrace{\sum_{a_3'} e^{ \alpha Q(s_3a_3')}}_{e^{ \alpha V(s_3)}} \right] = \\
    \frac{1}{\alpha} \sum_{s_3}p(s_3|s_2a_2) \left[ -{\cal D}_{KL} \left( \pi_\alpha (a_3|s_3) \left\| \frac{e^{ \alpha Q(s_3a_3)}}{e^{ \alpha V(s_3)}} 
      \right. \right) + \alpha V(s_3) \right]. 
  \end{array}
\end{equation}
Similarly to above, ${\cal D}_{KL}=0$ when $\pi_\alpha(a_3|s_3) = \frac{e^{ \alpha Q(s_3a_3)}}{{e^{ \alpha V(s_4)}}}$. The best policy distribution is then $\pi^*(a_3|s_3) \propto \frac{e^{ Q(s_3a_3)}}{{e^{ V(s_4)}}}$. Carrying over $\sum_{s_3}p(s_3|s_2a_2) V(s_3)$, we have 
\begin{equation}
  \underbrace{ R'(s_2a_2) + \sum_{s_3}p(s_3|s_2a_2) V(s_3)}_{Q(s_2a_2)} - \frac{1}{\alpha} \log \pi_\alpha (a_2|s_3).   
\end{equation}
Following similar steps, we have $\pi^*(a_2|s_2) \propto \frac{e^{ Q(s_2a_2)}}{{e^{ V(s_2)}}}$ and 
\begin{equation}
  \underbrace{ R'(s_1a_1) + \sum_{s_2}p(s_2|s_1a_1) V(s_2)}_{Q(s_1a_1)} - \frac{1}{\alpha} \log \pi_\alpha (a_1|s_1).   
\end{equation}
The last step is 
\begin{equation}
  \begin{array}{L}
    \frac{1}{\alpha} \sum_{s_1}p(s_1) \left[ \sum_{a_1} \pi_\alpha (a_1|s_1) \left(\alpha Q(s_1a_1) - \log \pi_\alpha (a_1|s_1) \right) \right] = \\ 
    \frac{1}{\alpha} \sum_{s_1}p(s_1) \left[ \sum_{a_1} \pi_\alpha (a_1|s_1) \left( \log \frac{e^{\alpha Q(s_1a_1)}}{\pi_\alpha (a_1|s_1)} 
      \frac{\sum_{a_1'} e^{ \alpha Q(s_1a_1')}}{\sum_{a_1'} e^{ \alpha Q(s_1a_1')} } \right) \right] = \\
    \frac{1}{\alpha} \sum_{s_1}p(s_1) \left[ -{\cal D}_{KL} \left( \pi_\alpha (a_1|s_1) \left\| \frac{e^{\alpha Q(s_1a_1)}}{\sum_{a_1'} e^{ \alpha 
      Q(s_1a_1')}} \right. \right) + \log \underbrace{\sum_{a_1'} e^{ \alpha Q(s_1a_1')}}_{e^{ \alpha V(s_1)}} \right] = \\
    \frac{1}{\alpha} \sum_{s_1}p(s_1) \left[ -{\cal D}_{KL} \left( \pi_\alpha (a_1|s_1) \left\| \frac{e^{\alpha Q(s_1a_1)}}{{e^{ \alpha V(s_1)}}} \right. 
      \right) + \alpha V(s_1) \right].  
  \end{array}
\end{equation}
This term is minimized when  $\pi_\alpha (a_1|s_1) = \frac{e^{ \alpha Q(s_1a_1)}}{{e^{ \alpha V(s_1)}}}$, with the optimal policy distribution $\pi^* (a_1|s_1) \propto \frac{e^{  Q(s_1a_1)}}{{e^{ V(s_1)}}}$.
Therefore, the recursions at a generic time step $t$ are 
\begin{equation}
  \begin{array}{L}
    Q_{(S_tA_t)^1}(s_ta_t) =  \log p(a_t) + R(s_ta_t) + \sum_{s_{t+1}} p(s_{t+1}|s_ta_t) V_{S_{t+1}}(s_{t+1}), \\
    V_{S_t}(s_t) = \frac{1}{\alpha } \log \sum_{a_t} e^{\alpha Q_{(S_tA_t)^1}(s_ta_t)},
  \end{array}
\end{equation}
with the optimal policy distribution   $\pi^*(a_t|s_t) \propto e^{Q(s_ta_t) - V(s_t)} $.

\bibliographystyle{apalike}
\bibliography{paper_x_arXiv}

\begin{table}[ht]
  \caption{Forward distributions for the source blocks.}
  \label{tab:source}
  \centering
  \begin{tabular}{m{1in} c c c} \toprule
    & \includegraphics[width=2cm]{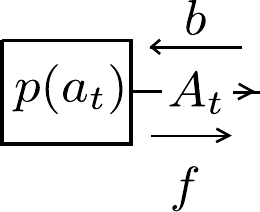} & \includegraphics[width=2cm]{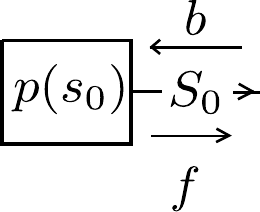} &\includegraphics[width=2cm]{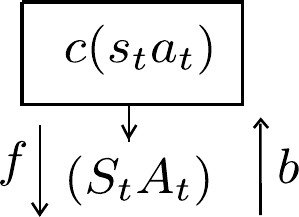} \\
    & $f_{A_t}(a_t)$ & $f_{S_0}(s_0)$ & $f_{(S_tA_t)}(s_ta_t)$ \\ \cmidrule(l){2-4}
    Sum product \newline Max product \newline Sum/Max product \newline DP \newline Max-Rew/Ent \newline SoftDP &
       $p(a_t)$ & $p(s_0)$ & $c(s_ta_t)$ \\ \bottomrule
  \end{tabular}
\end{table}

\begin{table}[ht]
  \caption{Propagation rules for action shaded blocks.}
  \label{tab:shadedA}
  \centering
  \begin{tabular}{m{1.5in} M{2.5in} M{1in}} \toprule
    & \multicolumn{2}{c}{\includegraphics[width=3cm]{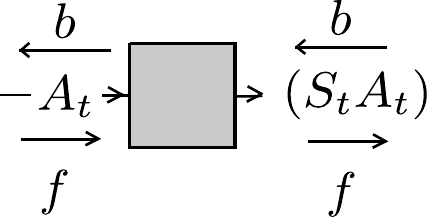}} \\ 
    & b_{A_t}(a_t) & f_{(S_tA_t)}(s_ta_t) \\ \cmidrule(l){2-3}
    Sum product \newline Max-Rew/Ent ($\alpha=1$) & \sum_{s_t} b_{(S_tA_t)}(s_ta_t) & f_{A_t}(a_t) U(s_t) \\ \addlinespace
    Max product \newline DP & \max_{s_t} b_{(S_tA_t)}(s_ta_t) & f_{A_t}(a_t) U(s_t) \\ \addlinespace
    Sum/Max product \newline Max-Rew/Ent ($\alpha \neq 1$) & \sqrt[\alpha]{\sum_{s_t} b_{(S_tA_t)}(s_ta_t)^\alpha} & f_{A_t}(a_t) U(s_t) \\ \addlinespace
    SoftDP & \exp{ \left[ \frac{\sum_{s_t} \log b_{(S_tA_t)}(s_ta_t) b_{(S_tA_t)}(s_ta_t)^\beta}{\sum_{s_t'}  b_{(S_tA_t)}(s_t'a_t)^\beta} \right]} 
      & f_{A_t}(a_t) U(s_t) \\ \bottomrule
  \end{tabular}
\end{table}

\begin{table}[ht]
  \caption{Propagation rules for state shaded blocks.}
  \label{tab:shadedS}
  \centering
  \begin{tabular}{m{1.3in} M{2.5in} M{1in}} \toprule
    & \multicolumn{2}{c}{\includegraphics[width=3cm]{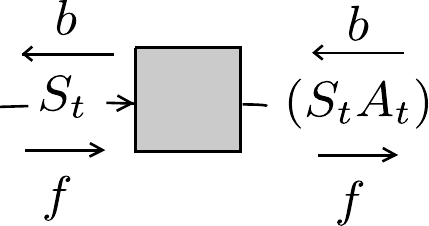}} \\ 
    & b_{S_t}(s_t) & f_{(S_tA_t)}(s_ta_t) \\ \cmidrule(l){2-3}
    Sum product \newline Max-Rew/Ent ($\alpha=1$) & \sum_{a_t} b_{(S_tA_t)}(s_ta_t) & f_{S_t}(s_t) U(a_t) \\ \addlinespace
    Max product \newline DP & \max_{a_t} b_{(S_tA_t)}(s_ta_t) & f_{S_t}(s_t) U(a_t) \\ \addlinespace
    Sum/Max product \newline Max-Rew/Ent ($\alpha \neq 1$)& \sqrt[\alpha]{\sum_{a_t} b_{(S_tA_t)}(s_ta_t)^\alpha} & f_{S_t}(s_t) U(a_t) \\ \addlinespace
    SoftDP & \exp{ \left[  \frac{\sum_{a_t} \log b_{(S_tA_t)}(s_ta_t) b_{(S_tA_t)}(s_ta_t)^\beta}{\sum_{a_t'} b_{(S_tA_t)}(s_ta_t')^\beta} \right]} 
      & f_{S_t}(s_t) U(a_t) \\ \bottomrule
  \end{tabular}
\end{table}

\begin{table}[ht]
  \caption{Propagation rules for the dynamics block.}
  \label{tab:dynamics}
  \centering
  \begin{tabular}{m{1in} M{2in} M{2in}} \toprule
    & \multicolumn{2}{c}{\includegraphics[width=5cm]{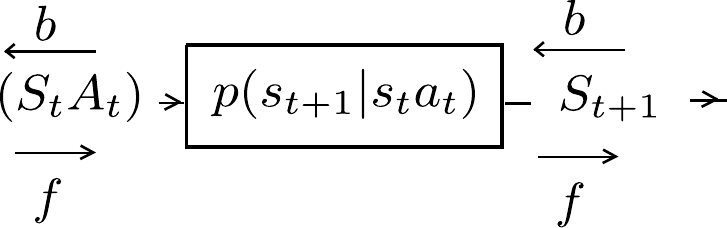}} \\ 
    & b_{(S_tA_t)}(s_ta_t) & f_{S_{t+1}}(s_{t+1}) \\ \cmidrule(l){2-3}
    Sum product & \sum_{s_{t+1}} p(s_{t+1}|s_ta_t) b_{S_{t+1}}(s_{t+1}) & \sum_{s_ta_t} p(s_{t+1}|s_ta_t) f_{(S_tA_t)}(s_ta_t) \\ \addlinespace
    Max product & \max_{s_{t+1}} p(s_{t+1}|s_ta_t) b_{S_{t+1}}(s_{t+1}) & \max_{s_ta_t} p(s_{t+1}|s_ta_t) f_{(S_tA_t)}(s_ta_t) \\ \addlinespace
    Sum/Max product & \sqrt[\alpha]{\sum_{s_{t+1}} p(s_{t+1}|s_ta_t)^\alpha b_{S_{t+1}}(s_{t+1})^\alpha} 
      & \sqrt[\alpha]{\sum_{s_ta_t} p(s_{t+1}|s_ta_t)^\alpha f_{(S_tA_t)}(s_ta_t)^\alpha} \\ \addlinespace
    DP \newline SoftDP \newline Max-Rew/Ent & e^{\sum_{s_{t+1}} p(s_{t+1}|s_ta_t) \log b_{S_{t+1}}(s_{t+1})} 
      & e^{\sum_{s_ta_t} p(s_{t+1}|s_ta_t) \log f_{(S_tA_t)}(s_ta_t)} \\ \bottomrule
  \end{tabular}
\end{table}

\begin{table}[ht]
  \caption{Propagation rules for the diverter.}
  \label{tab:diverter}
  \centering
  \begin{tabular}{m{1in} c} \toprule
    & \includegraphics[width=3cm]{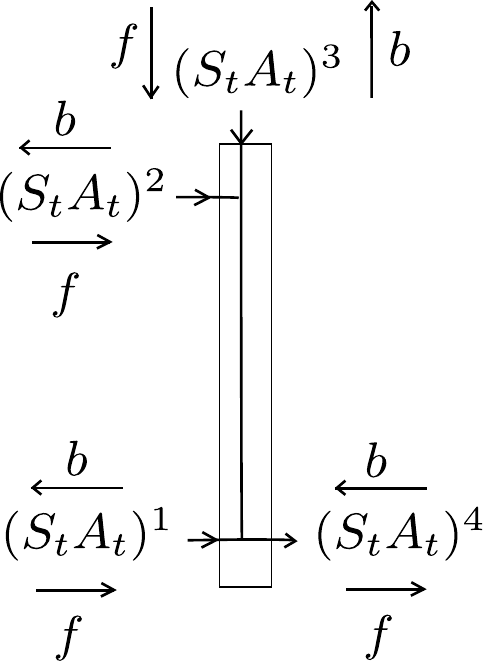} \\ \cmidrule(l){2-2}
    Sum product \newline Max product \newline Sum/Max product \newline DP \newline Max-Rew/Ent \newline SoftDP &
    $\begin{array}{L}  
      b_{(S_tA_t)^1}(s_ta_t) \propto f_{(S_tA_t)^2}(s_ta_t) f_{(S_tA_t)^3}(s_ta_t) b_{(S_tA_t)^4}(s_ta_t) \\
      b_{(S_tA_t)^2}(s_ta_t) \propto f_{(S_tA_t)^1}(s_ta_t) f_{(S_tA_t)^3}(s_ta_t) b_{(S_tA_t)^4}(s_ta_t) \\
      b_{(S_tA_t)^3}(s_ta_t) \propto f_{(S_tA_t)^1}(s_ta_t) f_{(S_tA_t)^2}(s_ta_t) b_{(S_tA_t)^4}(s_ta_t) \\
      f_{(S_tA_t)^4}(s_ta_t) \propto f_{(S_tA_t)^1}(s_ta_t) f_{(S_tA_t)^2}(s_ta_t) b_{(S_tA_t)^3}(s_ta_t)
    \end{array}$ \\ \bottomrule
  \end{tabular}
\end{table}

\end{document}